%% file: main.tex
\documentclass[10pt,twocolumn,letterpaper]{article}

\usepackage{cvpr}              %
\usepackage{epsfig}
\usepackage{graphicx}
\usepackage{amsmath}
\usepackage{amssymb}
\usepackage{mathtools}
\usepackage{makecell}
\usepackage{array}
\usepackage{soul}
\newcolumntype{P}[1]{>{\centering\arraybackslash}p{#1}}
\usepackage{float}
\usepackage{pifont}%
\usepackage{multirow}
\usepackage{nicefrac}       %
\usepackage{microtype}  
\usepackage{tabularx}
\usepackage{array}
\usepackage{enumitem}
\usepackage{afterpage}
\usepackage{scalerel,xparse}
\usepackage{capt-of,etoolbox}

\usepackage{url}            %
\usepackage{booktabs}       %
\usepackage{amsfonts}       %
\usepackage{nicefrac}       %
\usepackage{microtype}      %
\usepackage{graphics}
\usepackage{xfrac}

\input{preamble}

\usepackage{colortbl}

\definecolor{cvprblue}{rgb}{0.21,0.49,0.74}
\usepackage[pagebackref,breaklinks,colorlinks,citecolor=cvprblue]{hyperref}
\usepackage[capitalize]{cleveref}
\crefname{section}{Sec.}{Secs.}
\Crefname{section}{Section}{Sections}
\Crefname{table}{Table}{Tables}
\crefname{table}{Tab.}{Tabs.}
\def\paperID{4607} %
\def\confName{CVPR}
\def\confYear{2024}

\title{\emph{Shadows Don't Lie and Lines Can't Bend!} \\ Generative Models don't know Projective Geometry...for now}
\author{Ayush Sarkar\thanks{equal contribution}$^{*1}$\quad 
Hanlin Mai$^{*1}$ \quad 
Amitabh Mahapatra$^{*1}$ \quad 
Svetlana Lazebnik$^{1}$ \quad 
\\
D.A. Forsyth$^{1}$ \quad  
Anand Bhattad$^{2}$\\
$^{1}$University of Illinois Urbana-Champaign \quad $^{2}$Toyota Technological Institute at Chicago    \\
\texttt{\href{https://projective-geometry.github.io/}{https://projective-geometry.github.io/}}
}

\begin{document}
\makeatletter
\g@addto@macro\@maketitle{
    \begin{figure}[H]
    \scriptsize
        \setlength{\linewidth}{1\textwidth}
        \setlength{\hsize}{\textwidth}
        \vspace{-10mm}
    \centering
  \footnotesize
  \setlength\tabcolsep{0.2pt}
  \renewcommand{\arraystretch}{0.1}
  \begin{tabular}{ccccc}
 
    \includegraphics[width=0.2\linewidth]{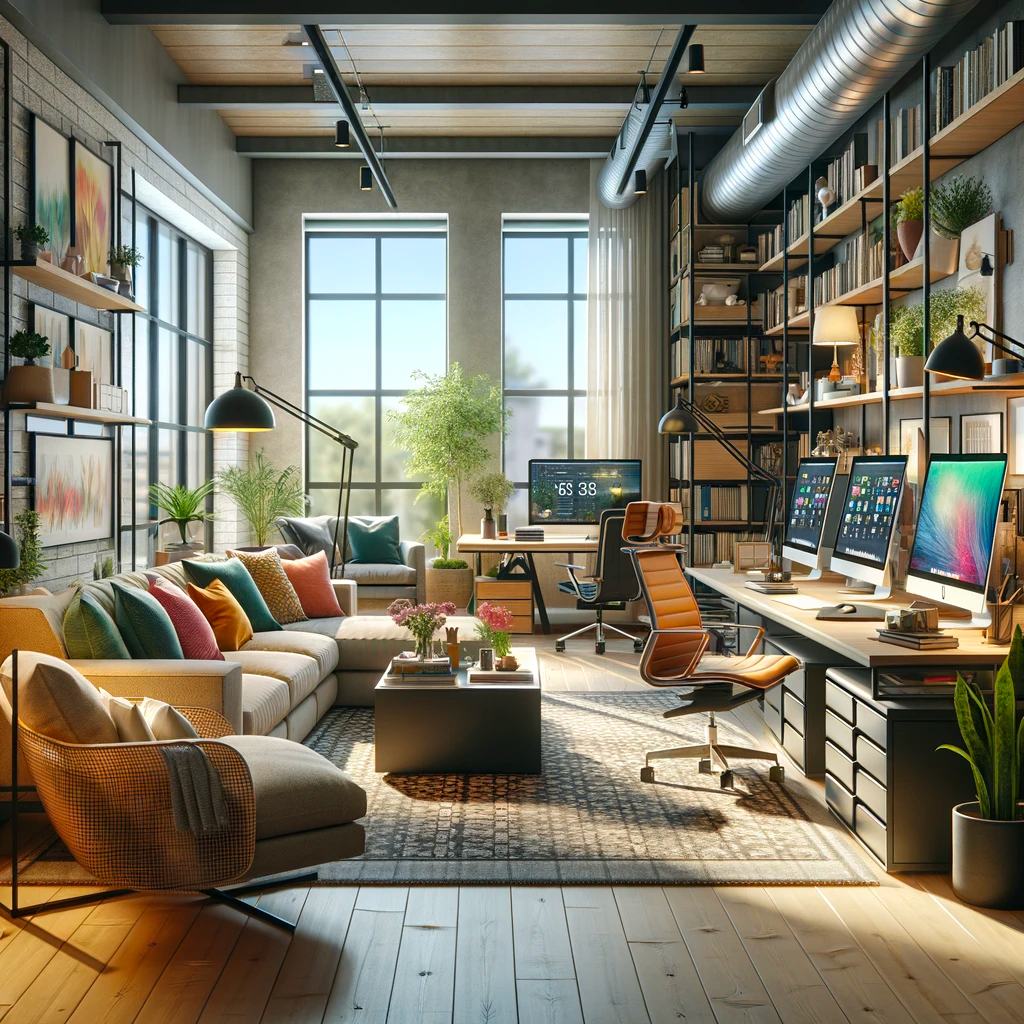} &
    \includegraphics[width=0.2\linewidth]{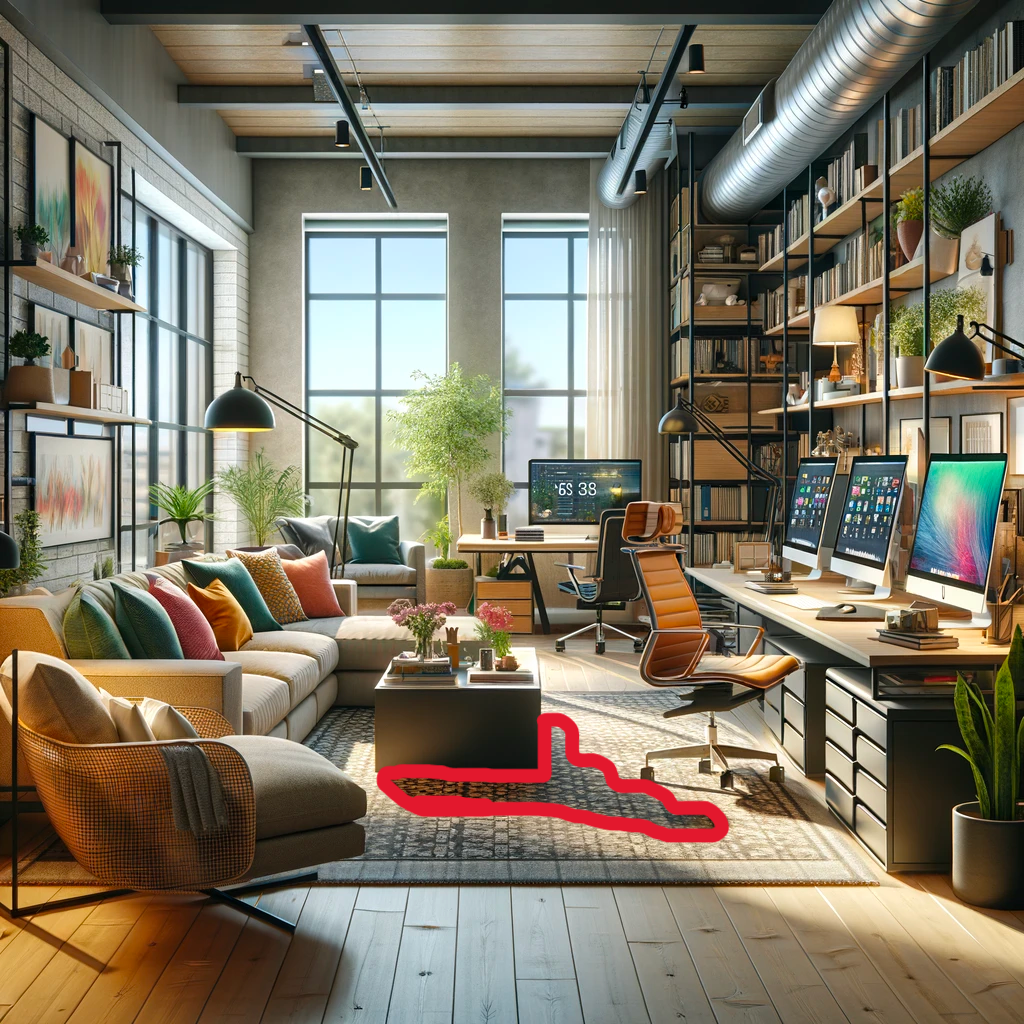} &
    \includegraphics[width=0.2\linewidth]{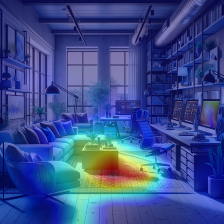} &
    \includegraphics[width=0.2\linewidth]{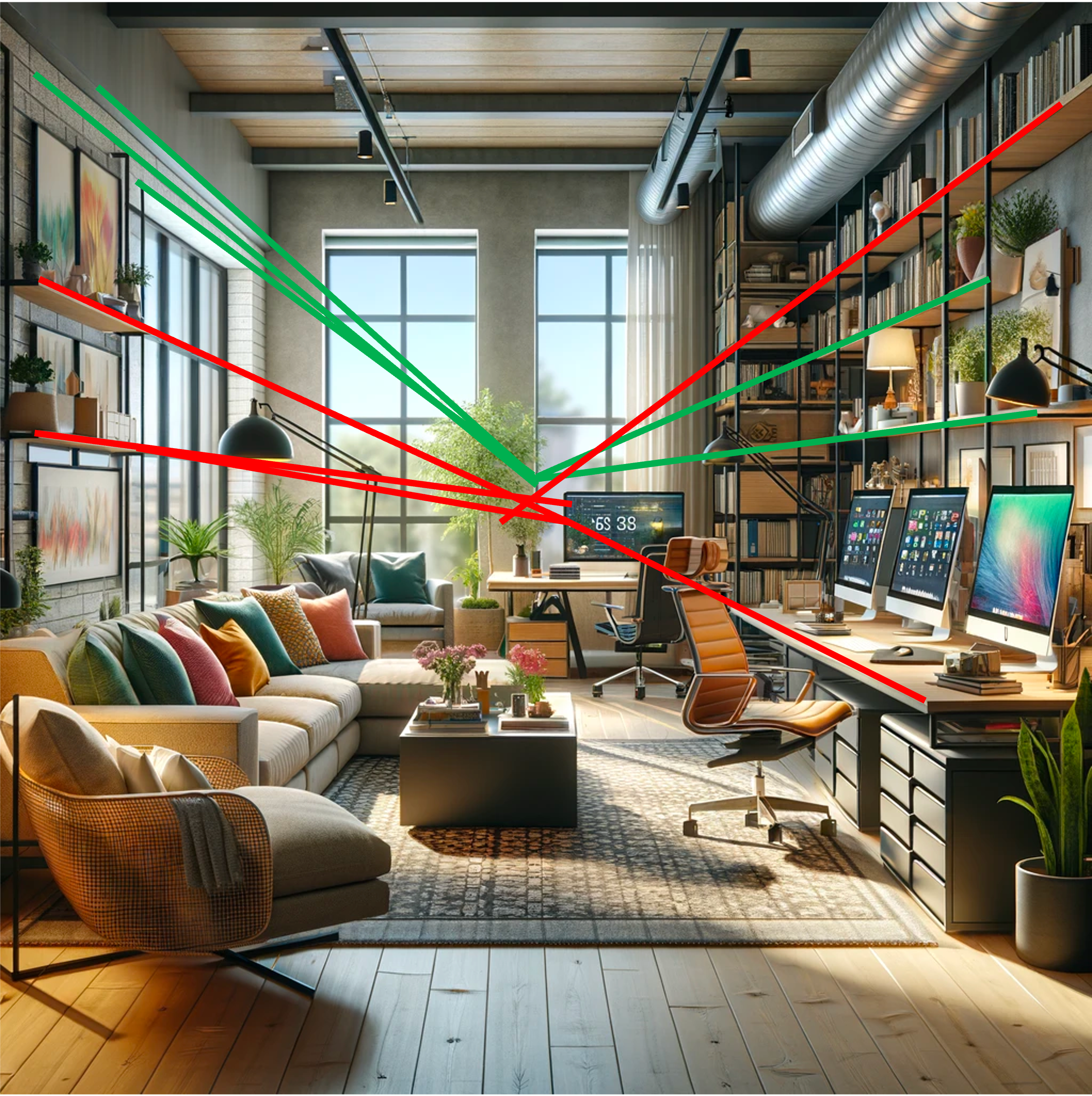} &
    \includegraphics[width=0.2\linewidth]{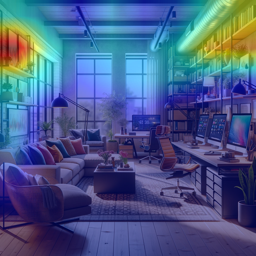}

 \\
    \includegraphics[width=0.2\linewidth]{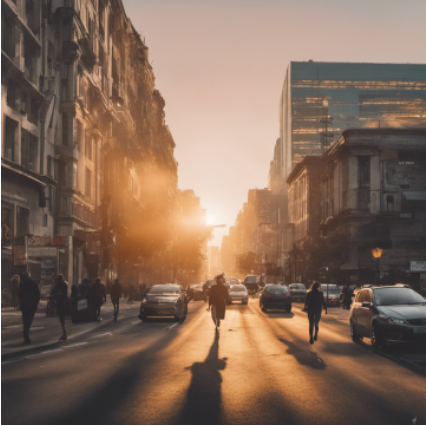} &
    \includegraphics[width=0.2\linewidth]{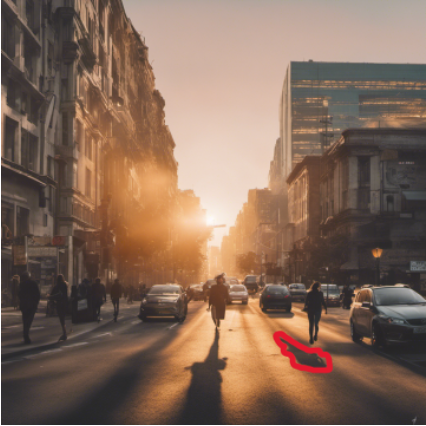} &
    \includegraphics[width=0.2\linewidth]{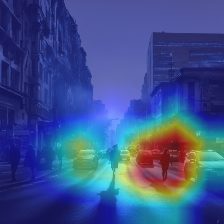} &
    \includegraphics[width=0.2\linewidth]{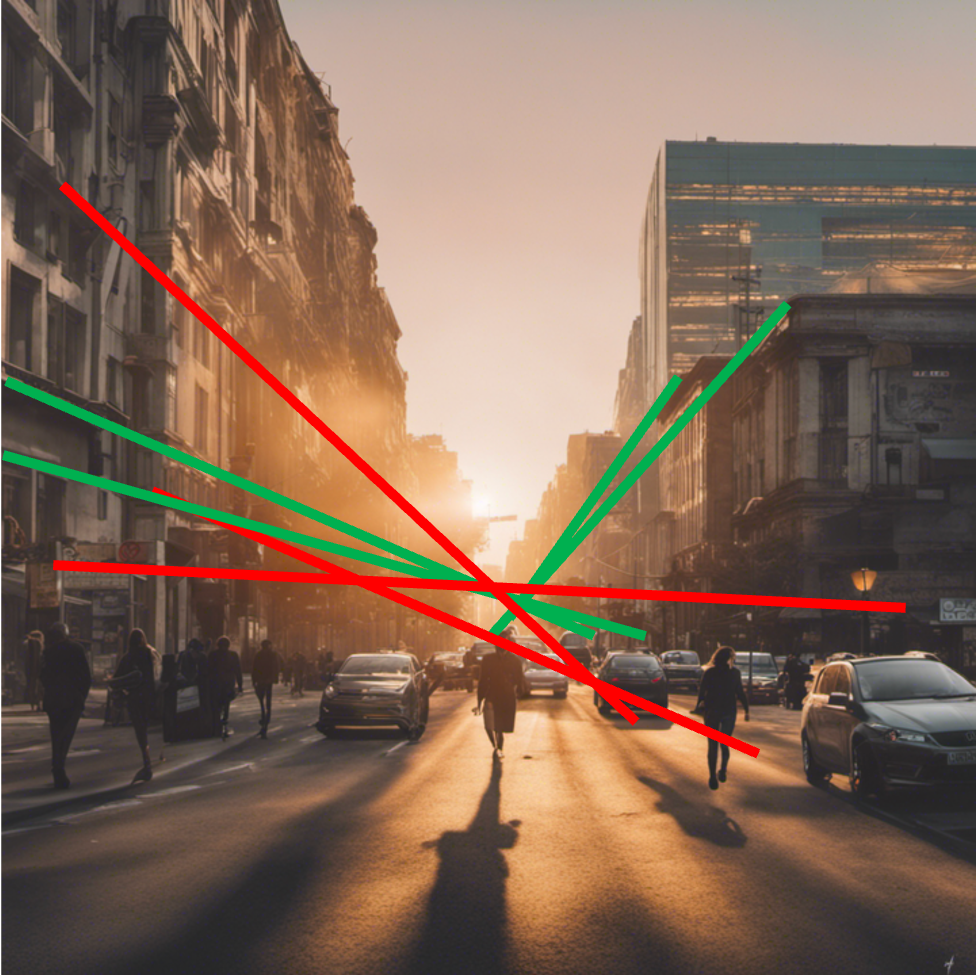} &
    \includegraphics[width=0.2\linewidth]{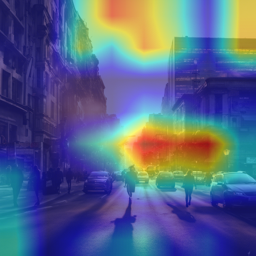}

\\
  \thead{Generated Image }&  \thead{Shadow Errors} &  Detected Shadow Errors & Vanishing Point Errors  &  Detected Perspective Errors
  \vspace{-5pt}
        \end{tabular}
  \caption{ The first column presents visually compelling AI-generated images. However, a closer examination reveals fundamental inconsistencies, such as those in shadow alignment (second column) and vanishing point accuracy (fourth column). Our model's analysis, shown in the third and fifth columns, detects these shadow and perspective geometry errors. We show that these errors are systematic and can be used to identify generated images.}
        \label{fig:teaser}
        \vspace{-2pt}
	\end{figure} 
}
\makeatother

\maketitle
\input{sec/0_abstract}   
\input{sec/1_intro}

\input{sec/2_related}

\input{sec/2.5_background}

\input{sec/3_dataset}

\input{sec/4_method}

\input{sec/5_results}

\input{sec/6_Discussion}

\section*{Acknowledgment}
This paper is based on work supported in part by the National Science Foundation under Grant No. 2106825 and a gift from Boeing Corporation.  
{
    \small
    \bibliographystyle{ieeenat_fullname}
    \bibliography{main}
}

\input{sec/X_suppl}

\end{document}

%% file: preamble.tex
\usepackage[dvipsnames]{xcolor}

%% file: sec/0_abstract.tex
\begin{abstract}
  Generative models can produce impressively realistic images.
  This paper demonstrates that generated images have geometric
  features different from those of real images.  We  build a set of
  collections of generated images, prequalified to fool simple,
  signal-based classifiers into believing they are real.  We then show
  that prequalified generated images can be identified reliably by
  classifiers that only look at geometric properties.  We use three
  such classifiers.  All three
  classifiers are denied access to image pixels, and look only at
  derived
  geometric features. The first classifier looks at the perspective field of the image,
  the second looks at lines detected in the image, and the third looks at
  relations between detected objects and shadows.  Our procedure
  detects generated images more reliably than SOTA local signal based
  detectors, for images from a number of distinct generators. Saliency
  maps suggest that the classifiers can identify geometric problems reliably.
  We conclude that current generators cannot reliably reproduce
  geometric properties of real images.
\end{abstract}

%% file: sec/1_intro.tex
\section{Introduction}
\label{sec:intro}
Both StyleGAN~\cite{karras2019style, karras2020analyzing, karras2021alias} and diffusion
models~\cite{rombach2022high, ramesh2022hierarchical,  saharia2022photorealistic}
are renowned for generating images that are strikingly similar to real-world photos and
consistently fool people. But, as we show, generated images have distinctive geometric
features, likely from a failure to fully capture projective geometry.

Bhattad et al.~\cite{bhattad2023stylegan, bhattad2023StylitGAN}, Chen et al.~\cite{chen2023beyond}, Du et al.~\cite{du2023generative}, Zhan et al.~\cite{zhan2023does}, and have
shown generative models implicitly capture the complex scene
properties, including normals, depth, albedo, and support relations. These works
suggest these models ``understand'' geometry, which would be useful 
for rendering 3D scenes.  Our detailed, population-level analysis of generated images
suggests generative models~\cite{podell2023sdxl, arkhipkin2023kandinsky, deepfloyd2023, betker2023improving, adobefirefly2023} cannot fully translate this ``understanding''
into accurate geometry.  Specifically, we demonstrate that generative models produce
images with lines that differ from those of real images (likely due to problems aligning
vanishing points); that generative models produce images with perspective fields that are
unlike those of real images; and that object-shadow relations in generated images differ
reliably from those in real images.  We use
advanced pretrained models (Line Segment Detection~\cite{Pautrat_2023_DeepLSD};
Perspective Fields~\cite{jin2023perspective};  and PointNet~\cite{qi2017pointnet})
that inspect geometric representations to distinguish between real and generated images.  We rely on derived geometry cues from SOTA methods as manual image analysis and explicit geometry-based rules like drawing lines perspective lines are not scalable or automatable.

To guarantee the accuracy of our findings, we follow a strict process of data curation. This involves using a controlled pixel-level classifier to filter out any biases related to color, texture, and local features within our test set. This precision in data selection is crucial to isolate and accurately assess the inconsistencies in projective geometry and illumination present in generated images. By carefully screening the data, we can ensure that our results are not obscured by common artifacts that are usually found in generated images. This enhances the reliability of our conclusions.  
Our contributions are:
\begin{itemize}
    \item \textbf{Unearthing Geometric Discrepancies:} We present a comprehensive analysis
      that goes beyond existing literature to both demonstrate and quantify geometric discrepancies produced by current generative models.
    \item \textbf{Scalable, Data-Driven Approach:} We introduce a scalable, data-driven approach by utilizing three distinct projective geometry cues to detect geometric inaccuracies. 
    \item \textbf{Robust and Generalizable:} We demonstrate robustness and generalizability by effectively identifying geometric errors across a wide array of images from recent generative models, including those with the latest time stamps.
    \item  \textbf{Broadening the Scope of Model Assessment:} Our approach can be used as an
      alternative method for evaluating models: do they get projective geometry right? 
\end{itemize}

%% file: sec/2_related.tex
\section{Related Work}
\label{sec:related}
\textbf{Generative Models:} The advancement of generative models, particularly in creating visually realistic images, marks a significant milestone in computer vision. Pioneering efforts by Karras et al.~\cite{karras2019style, karras2020analyzing, karras2021alias} with StyleGAN, and the emergence of diffusion models~\cite{rombach2022high, ramesh2022hierarchical, saharia2022photorealistic}, have set new benchmarks in realism. These models, used in diverse fields from art to data augmentation, have yet to fully grasp the nuances of projective geometry, which is the focus of our analysis, primarily using open diffusion models.

\noindent\textbf{Geometric Understanding in Generative Models:} While studies like Bhattad et al.~\cite{bhattad2023stylegan}, Du et al.\cite{du2023generative}, Chen et al. \cite{chen2023beyond} and Zhan et al.~\cite{zhan2023does} demonstrate these models' potential in understanding scene geometry, our work diverges by scrutinizing the generated images themselves, examining their adherence to the principles of projective geometry and illumination, rather than analyzing learned features.

\noindent\textbf{Detecting Generated Images:} The realism of modern generative models has made image forensics increasingly challenging. Traditional methods focused on detecting synthetic images using signals like resampling artifacts~\cite{popescu2005exposing} and JPEG quantization~\cite{agarwal2017photo}. Kee et al.~\cite{kee2013exposing} introduced a geometric technique for detecting shadow inconsistencies, paralleling our pursuit of physical realism. However, our work extends beyond identifying photo manipulation to evaluating the overall perspective geometry and illumination consistency in images from generative models.

Zhang et al.'s work \cite{Epstein_2023_ICCV} focuses on detecting AI-generated images using diverse generative models and online training for future model adaptation. Our research, in contrast, assesses the projective geometry in these images, examining their ability to render scenes with accurate perspective and illumination. Boh\'{a}\v{c}ek et al. \cite{bohavcek2023geometric}, while detecting geometric inconsistencies related to shadows, align with our interest in physical realism. However, we delve deeper, thoroughly evaluating perspective geometry and illumination in generative models for a more comprehensive understanding of their geometric accuracy.

 The rise of generative methods has steered image forensics towards using discriminative methods to detect synthetic content~\cite{zhou2018learning, huh2018fighting, wang2019detecting, yu2019attributing, wang2019cnngenerated, chai2020makes, epstein2023online}. These advancements align with our objective of analyzing the physical and geometrical congruence of generated images. However, our work goes a step further by critically assessing whether generative models fundamentally understand and accurately replicate projective geometry, rather than simply distinguishing between real and synthetic images. This deeper level of analysis aims to unveil the intricacies and limitations of current models in faithfully rendering geometrically coherent images.

\noindent\textbf{Evaluation Metrics:} Traditional metrics like the Inception Score (IS) \cite{NIPS2016_8a3363ab} and Fréchet Inception Distance (FID) \cite{NIPS2017_8a1d6947} focus on pixel-level fidelity. The emergence of CLIP-based scores \cite{hessel-etal-2021-clipscore, radford2021learning} and DIRE \cite{Wang_2023_ICCV} offers a semantic perspective. In contrast, our approach, distinct in its focus on perspective geometry and illumination consistency, seeks to ensure comprehensive realism, bridging the gap between visual and physical authenticity.

Recent studies like Davidsonian Scene Graph~\cite{cho2023davidsonian} and ImagenHub \cite{ku2023imagenhub} address fine-grained evaluation inconsistencies, while the HEIM benchmark~\cite{lee2023holistic} assesses models across multiple aspects. Our work complements these by providing an in-depth evaluation of the physical and geometric realism of images generated by state-of-the-art models.

%% file: sec/2.5_background.tex
\section{Background on Projective Geometry}

Projective geometry is a mathematical framework that enables the accurate representation of three-dimensional spaces in two-dimensional images. It provides the rules for perspective, which are crucial for creating realistic scenes with depth and spatial orientation   ~\cite{forsyth2002computer}. In this section, we will examine the common inconsistencies that may arise during image synthesis according to projective geometry. Our evaluation framework is intended to detect and measure these discrepancies, which are essential for evaluating the realism and physical plausibility of generated images.

\noindent\textbf{Inconsistent Vanishing Points.}
Vanishing points are fundamental to capturing the essence of perspective in images. They should align with the direction of parallel lines converging at a distance. Generated images often exhibit inconsistencies where these lines do not meet at the correct vanishing points, leading to a distorted sense of perspective.

\noindent\textbf{Lighting and Shadow Inconsistencies.}
Accurate shadows are essential for reinforcing the position and shape of objects within a scene. Discrepancies in shadow direction, length, and softness can indicate a misalignment with the scene's light sources, disrupting the image's three-dimensionality.

\noindent\textbf{Scale Discrepancies.}
The principle of size constancy dictates that objects of the same size should appear smaller as their distance from the observer increases. Generated images sometimes fail to maintain this scaling, resulting in a compromised depth perception. 

\noindent\textbf{Distortion of Geometric Figures.}
Geometric figures should maintain their shape when projected onto the image plane, barring intentional perspective distortion. Errors in this projection can result in circles appearing as ellipses or squares as trapezoids, indicating a flawed perspective rendering.

\noindent\textbf{Depth Cues.}
Depth perception in images is conveyed through cues such as overlapping, texture gradients, and relative size. Misrepresentation of these cues can lead to an unnatural spatial arrangement that the human eye can readily detect as artificial.

Our evaluation framework, detailed in the subsequent sections, is designed to rigorously test generated images against these projective geometry principles. While a comprehensive evaluation of projective geometry would consider all the aforementioned inconsistencies, our framework prioritizes the detection of inconsistent vanishing points and lighting and shadow inconsistencies. These elements are particularly telling indicators of an image's projective geometry realism and are often the most challenging for generative models to replicate accurately.

%% file: sec/3_dataset.tex
\section{Dataset Curation}

\begin{table*}[ht]
\centering
\caption{Statistical overview of the Data Curation and Filtering Process: We present the distribution of real and generated images for indoor and outdoor datasets. The ResNet50 Prequalifier helps us curate datasets, creating an `Unconfident Set' for images with low classifier certainty and a `Misclassified Subset' for images incorrectly labeled by the classifier. These rigorously curated sets are instrumental for our subsequent analysis, concentrating on projective geometry while mitigating the influence of signal cues. The datasets for each prequalifier—indoor, outdoor, and combined—are prepared separately to tailor the models to their specific contexts.}
\label{tab:data_stats}
\vspace{-5pt}
\resizebox{0.95\linewidth}{!}{%
\begin{tabular}{lcccccc}
\toprule
 & \textbf{Indoor Real} & \textbf{Indoor Generated} & \textbf{Outdoor Real} & \textbf{Outdoor Generated} & \textbf{Combined Real} & \textbf{Combined Generated}\\
\midrule
Total Images & 400,000 & 400,000 & 125,000 & 125,000 & 525,000 & 525,000\\
\midrule
\multicolumn{7}{c}{\textbf{Training and Test Sets}} \\
\midrule
Training Set Size & 75,000 & 75,000 & 25,000 & 25,000 & 100,000 & 100,000\\
Validation Set Size &  10,000 & 10,000 & 5,000 & 5,000 & 15,000 & 15,000\\ 
Test Set Size & 315,000 & 315,000 & 95,000 & 95,000 & 410,000 & 410,000\\
\midrule
\multicolumn{7}{c}{\textbf{Post ResNet50 Prequalifier}} \\
\midrule
Unconfident Set & 23213 & 13840 & 1444 & 1018 & 44392 & 2540\\
Misclassified Subset & 10399  & 5756   & 609 & 443  & 23800 & 825\\

\bottomrule
\end{tabular}
}
\vspace{-5pt}
\end{table*}
Our data curation process is carefully designed to distinguish between real images and generated images from several generative models. This process includes models that the classifier has not seen during its training. Another important goal is to ensure our prequalifier effectively identifies images with recent timestamps. This helps prevent the classifier from relying on cues specific to the dataset, such as whether the image is from the training distribution or not, rather than determining whether the image is real or generated.

\subsection{Real Images}
We experiment with a diverse set of images, including:

\noindent\textbf{(a) Indoor Scenes:} A collection of 400,000 interior images featuring a variety of furniture arrangements and lighting conditions, sourced from LSUN \cite{yu2015lsun}. Specifically, we used 100,000 images each from Bedroom, Dining Room, Kitchen, and Living Room categories.

\noindent\textbf{(b) Outdoor Scenes:} A dataset of 125,000 outdoor scenes with varying landscapes and urban settings, sourced from Berkeley Deep Drive 100K \cite{yu2020bdd100k} and Mapillary Vistas \cite{8237796}. We selected images that represent a wide range of weather conditions and times of day.

\noindent\textbf{(c) Combination of Indoor and Outdoor scenes:} We also analyzed a combination of above indoor and outdoor scenes to assess the performance on a more diverse dataset.

\noindent\textbf{(d) Recent Timestamp Images:} A curated test set of 500 indoor and 500 outdoor images with timestamps ranging from May 2023 to March 2024. These images were collected from various social media platforms and online sources to ensure our classifier's ability to handle recent real-world data and that the models are not obscured because of any data-source-specific biases. 

\subsection{Image Captions}
We use the ViT-bigG-14/laion2b\_s39b\_b160k model \cite{ilharco_gabriel_2021_5143773} and the BLIP model \cite{pmlr-v162-li22n} in succession, to generate refined captions for real images. These models were chosen for their state-of-the-art performance in image captioning tasks. The ViT model uses a Vision Transformer architecture, while BLIP employs a multimodal pre-training approach. By using common captions, we ensure a fair evaluation of the projective geometry in generated images.

\subsection{Generated Images}
We generate images from Stable Diffusion XL v1.0 \cite{podell2023sdxl}, Kandinsky-v3 \cite{arkhipkin2023kandinsky}, DeepFloyd IF v1.0 \cite{deepfloyd2023}, and PixArt-$\alpha$ v1.0 \cite{chen2023pixartalpha}. We use the same caption from the real images to generate these images with the default settings of each generative model.

\subsection{Robust Prequalifier}
Our goal is to identify challenging images that a signal-based classifier may struggle to differentiate. Therefore, we need to develop a robust prequalifier. This is to ensure that our results are not affected by any false data signals. Thus, we aim to eliminate all potential factors that may influence the geometry cues we obtain because of signal weirdness in the generated images. To accomplish this, we begin by training signal-based classifiers using a vanilla ResNet-50 \cite{he2016deep} and primarily concentrate on identifying instances that prove challenging for these signal-based classifiers.

We trained prequalifiers on three distinct settings - indoor scenes, outdoor scenes and a combination of the two. Each prequalifier was trained on a dataset consisting of real images and images generated by one of the four models: Stable Diffusion XL, Kandinsky-v3, DeepFloyd IF, and PixArt-$\alpha$. While all four types of prequalifiers performed well on their respective test sets, we found that Kandinsky exhibited superior generalization capabilities when evaluated on images generated by other models or on recent timestamp images. The fundamental experiment conducted to assess the generalizability of different generators is covered in \cref{fig:roc_curves_other_generators_suppl} in the Supplementary Material.

For indoor scenes, our Kandinsky-based prequalifier achieves an Area Under the Curve (AUC) of 0.99 on its test set with an accuracy of 97.43 and maintained high performance on images generated by other models such as Stable Diffusion XL and PixArt-$\alpha$, achieving AUCs of 0.97 and 0.98 respectively. Furthermore, our Kandinsky-trained prequalifier meant for both indoor and outdoor settings combined showed robust performance on recent timestamp generated indoor and outdoor images, achieving AUC scores of 0.90, 0.95, and 0.72 on images generated by Stable Diffusion XL, PixArt-$\alpha$, and DeepFloyd IF, respectively.

Given their robustness and strong generalizability, we selected the Kandinsky-trained prequalifiers as the basis for training our derived geometry classifiers on Kandinsky-generated images. This choice ensures that our classifiers effectively look at geometric discrepancies and can handle various generated images. This enhances the reliability of our approach and supports our conclusion that the classifiers are not affected by spurious signal artifacts.

\begin{figure*}[t!]
    \centering
    \includegraphics[width=1\textwidth]{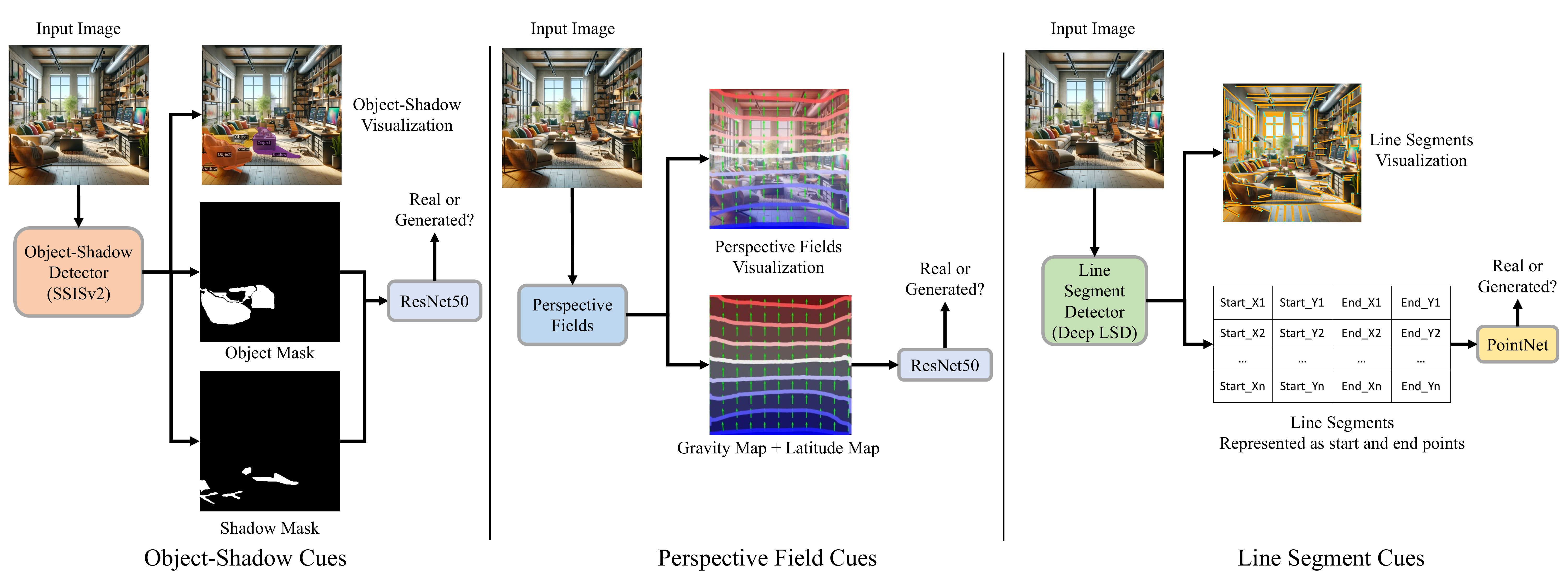}
    \vspace{-25pt}
    \caption{We train three classifiers to identify discrepancies in projective geometry. These classifiers are trained using derived geometry cues such as object-shadow associations (left), perspective fields (middle), and line segments (right) without looking at image intensity. We use the ResNet architecture for Object-Shadow and Perspective Fields, and PointNet for Line Segments to process unordered data sets. 
    }
    \label{fig:horizontal_classifier}
    \vspace{-10pt}
\end{figure*}

\subsection{Final Test Sets}
We categorize our test sets into three groups based on the accuracy of the prequalifier: easy, unconfident, and misclassified. The {\bf easy test set} includes images with 100\% accuracy, indicating that the prequalifier can reliably distinguish between real and generated images. The {\bf unconfident test set} includes images where the prequalifier performs at chance level, suggesting that it struggles to make confident predictions. Finally, the {\bf misclassified test set} includes images where the prequalifier makes completely wrong predictions, either classifying real images as generated or vice versa. A summary of this split for 
Kandinsky-v3 is provided in \cref{tab:data_stats}.

Our approach primarily evaluates the ``hard set'' (unconfident and misclassified) for geometric and shadow inconsistencies, assessing adherence to projective geometry principles. This ensures rigorous testing of generative models' ability to reproduce geometric correctness and photometric accuracy, beyond surface-level or signal details.

It is important to note that projective geometry inconsistencies are prevalent in generated images but often go undetected by conventional methods. Our models, trained on geometric abstractions and projective cues, can identify subtle but critical inaccuracies that texture artifacts cannot explain. This distinction is critical, as it allows us to rigorously test our models, which, unlike the prequalifier, do not have direct access to the images.

We employ a suite of models to capture different facets of projective geometry, trained on datasets emphasizing geometric consistency and photometric accuracy. By combining their strengths, we comprehensively assess the quality of generated images and provide insights into the limitations and potential improvements of generative models from a projective geometry perspective.

%% file: sec/4_method.tex
\section{Analyzing Projective Geometry}
For our analysis, we rely on three geometry cues -- object shadow, line segments, and perspective fields. We train three separate classifiers solely based on these derived geometry cues, without utilizing any pixel information. The training process for each classifier is described below, and a schematic pipeline of these classifiers can be found in \cref{fig:horizontal_classifier}.

\subsection{Object-Shadow Cues}
Our first model examines object-shadow relationships to address illumination. Shadows follow projective geometry principles, and inconsistencies can reveal generated images.

We use a pretrained object-shadow instance detection model~\cite{wang2022instance} to identify shadows and geometric heuristics to evaluate their plausibility given the objects and their orientation. A ResNet50 classifier is trained on binary masks of object and shadow instances to score the consistency of shadows with objects. Images with implausible shadows are marked as likely generated.

\subsection{Perspective Field Cues}
Our framework's second model uses Perspective Fields~\cite{jin2023perspective}, vector fields encoding the spatial orientation of pixels relative to vanishing points and the horizon, to assess projective geometry. We generate these fields from single images using a pretrained model.

We train a ResNet50 classifier on the Perspective Fields to differentiate real from generated images by focusing on projective geometry anomalies. The classifier evaluates the consistency of these fields with projective geometry principles, scoring images on their geometric plausibility. This method enables precise evaluation of projective geometry, enhancing the detection of subtle inconsistencies.

\subsection{Line Segment Cues}
Our method also assesses projective geometry in generated images by identifying key structural lines using Deep LSD~\cite{Pautrat_2023_DeepLSD}. These lines indicate adherence to perspective rules. We then train a PointNet-like architecture~\cite{qi2017pointnet} to classify images based on line segment patterns, differentiating real from generated images.

PointNet's flexibility in handling unordered data makes it suitable for analyzing line segments without pre-sorting. The model assigns scores representing the likelihood of an image being real based on the spatial arrangement of its lines. Analyzing these scores reveals the model's ability to detect subtle discrepancies in line arrangements, which often indicate a generated image.

%% file: sec/5_results.tex
\begin{figure*}[ht!]
    \centering
    \begin{subfigure}[b]{0.24\textwidth}
        \includegraphics[width=\textwidth]{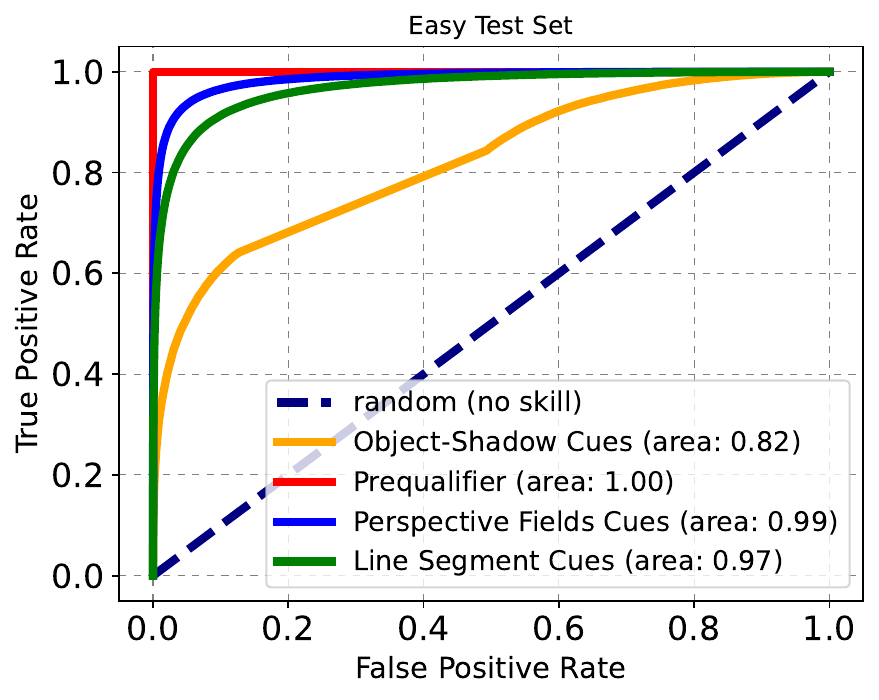}
        \caption{Easy Test Set (Indoor)}
        \label{fig:full_test_set}
    \end{subfigure}
    \hfill
    \begin{subfigure}[b]{0.24\textwidth}
        \includegraphics[width=\textwidth]{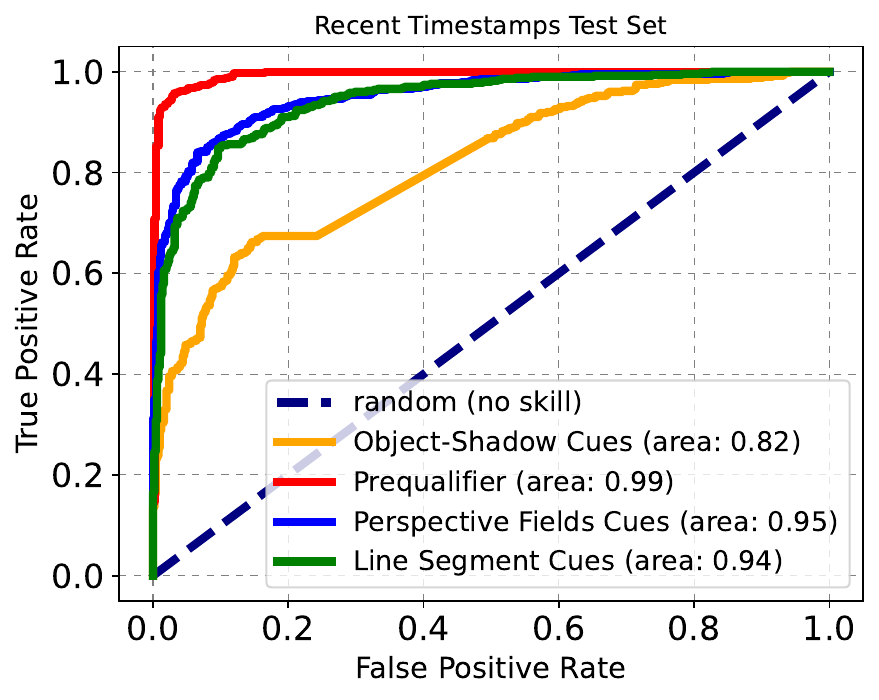}
        \caption{Recent Timestamp Set (Indoor)}
        \label{fig:recent_timestamp_test_set}
    \end{subfigure}
    \begin{subfigure}[b]{0.24\textwidth}
        \includegraphics[width=\textwidth]{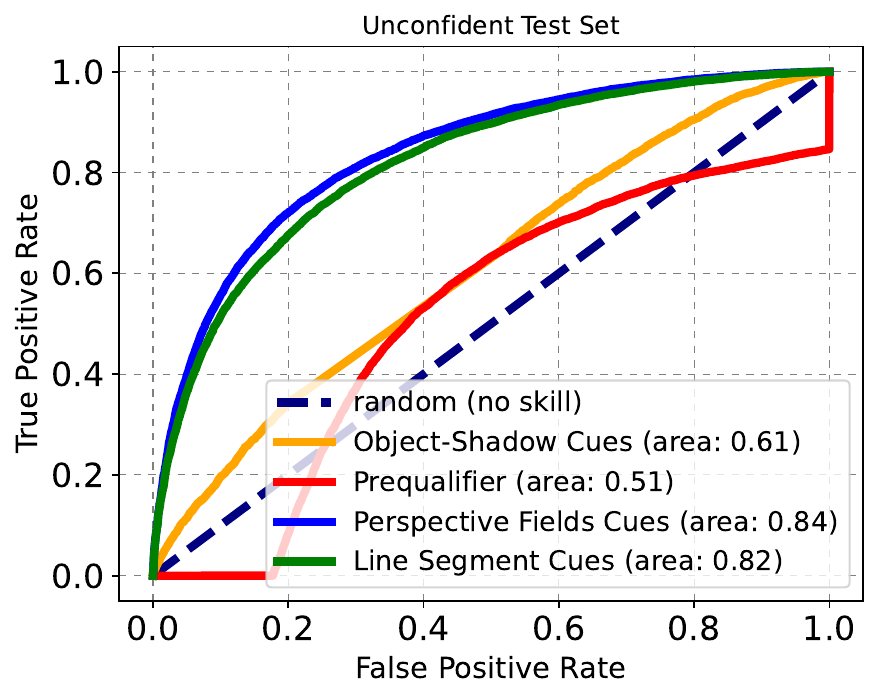}
        \caption{Unconfident Test Set (Indoor)}
        \label{fig:unconfident_test_set}
    \end{subfigure}
    \hfill
    \begin{subfigure}[b]{0.24\textwidth}
        \includegraphics[width=\textwidth]{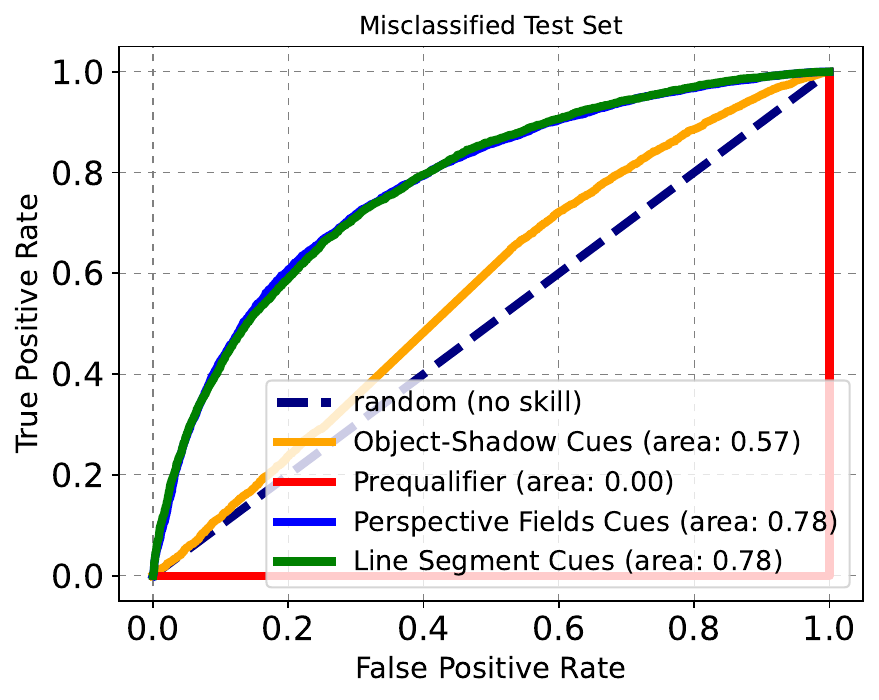}
        \caption{Misclassified Test Set (Indoor)}
        \label{fig:misclassified_test_set}
    \end{subfigure}
    \begin{subfigure}[b]{0.24\textwidth}
        \includegraphics[width=\textwidth]{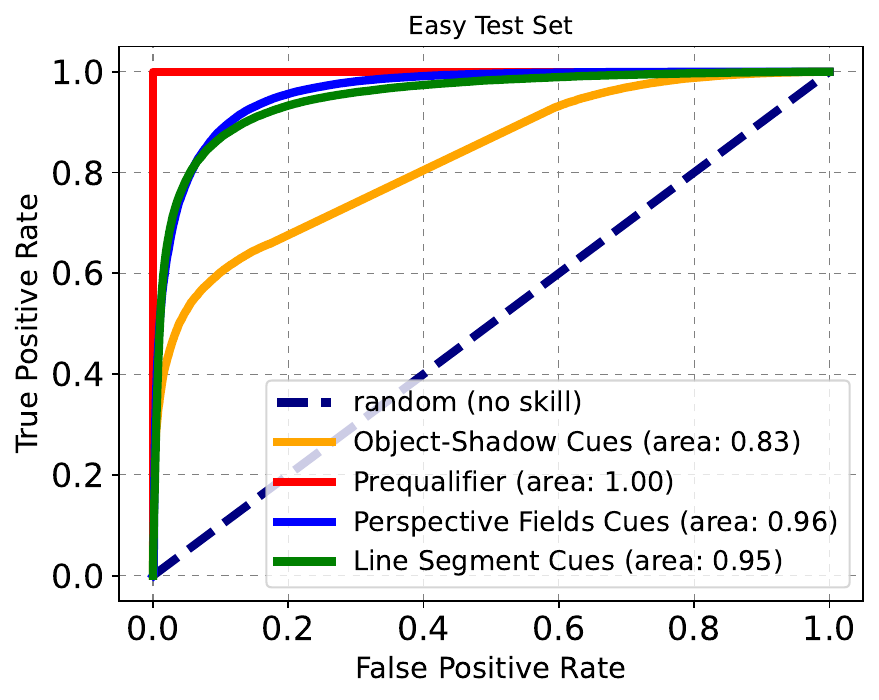}
        \caption{Easy Test Set (Outdoor)}
        \label{fig:outdoor_full_test_set}
    \end{subfigure}
    \hfill
    \begin{subfigure}[b]{0.24\textwidth}
        \includegraphics[width=\textwidth]{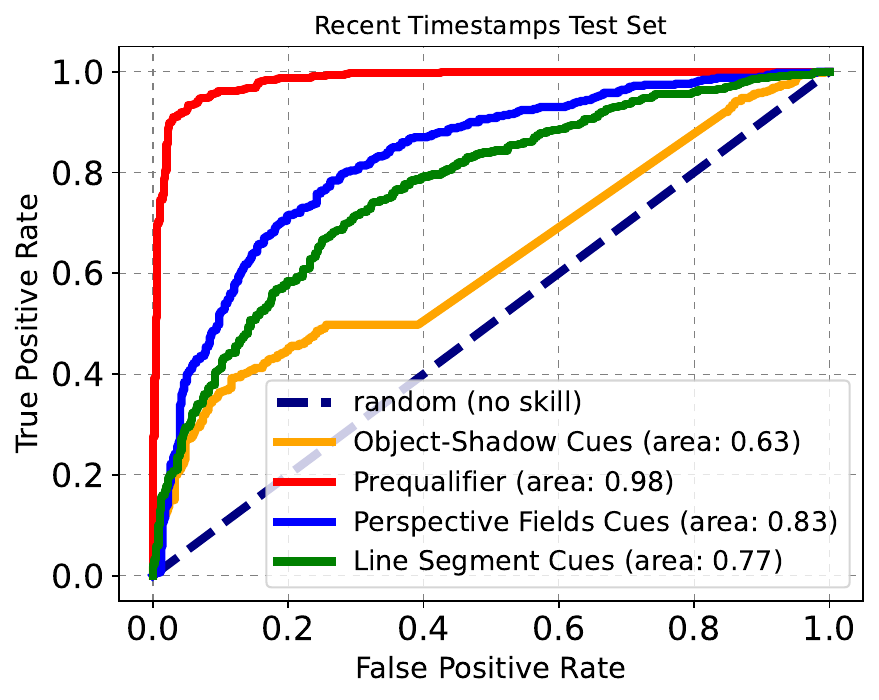}
        \caption{Recent Timestamp Set (Outdoor)}
        \label{fig:outdoor_recent_timestamp_test_set}
    \end{subfigure}
    \begin{subfigure}[b]{0.24\textwidth}
        \includegraphics[width=\textwidth]{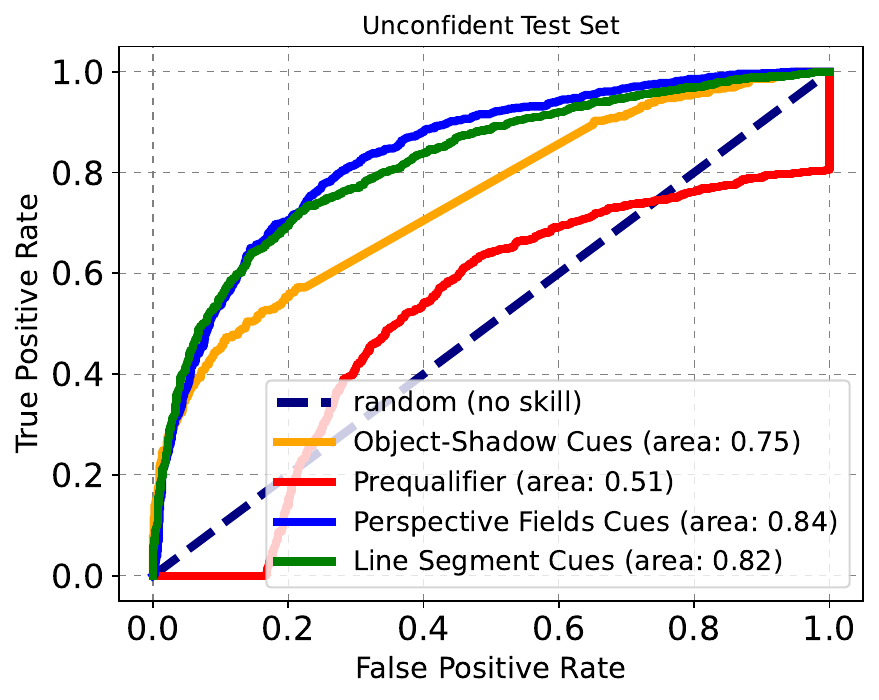}
        \caption{Unconfident Test Set (Outdoor)}
        \label{fig:outdoor_unconfident_test_set}
    \end{subfigure}
    \hfill
    \begin{subfigure}[b]{0.24\textwidth}
        \includegraphics[width=\textwidth]{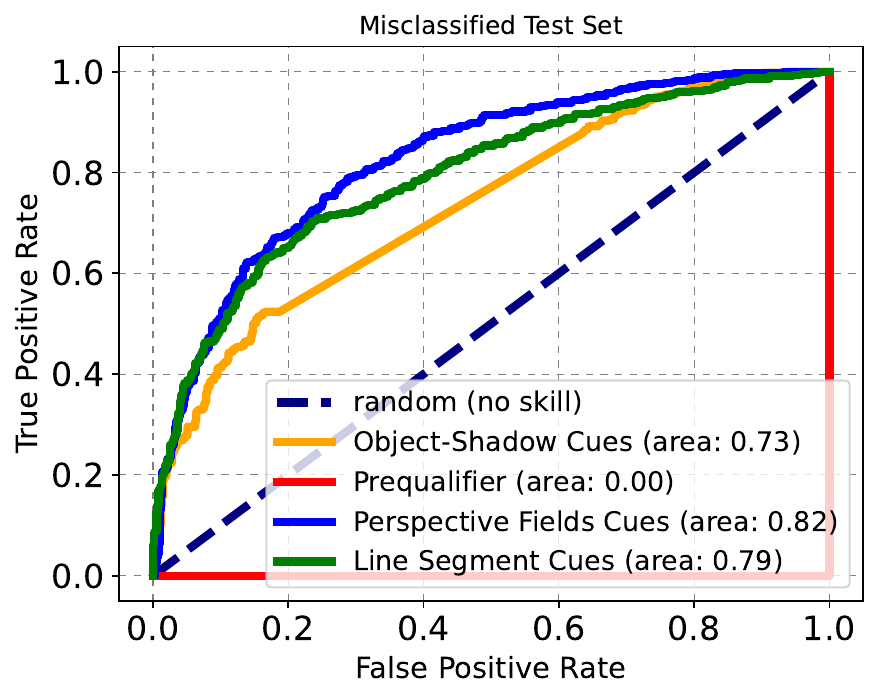}
        \caption{Misclassified Test Set ((Outdoor))}
        \label{fig:outdoor_misclassified_test_set}
    \end{subfigure}
    \begin{subfigure}[b]{0.24\textwidth}
        \includegraphics[width=\textwidth]{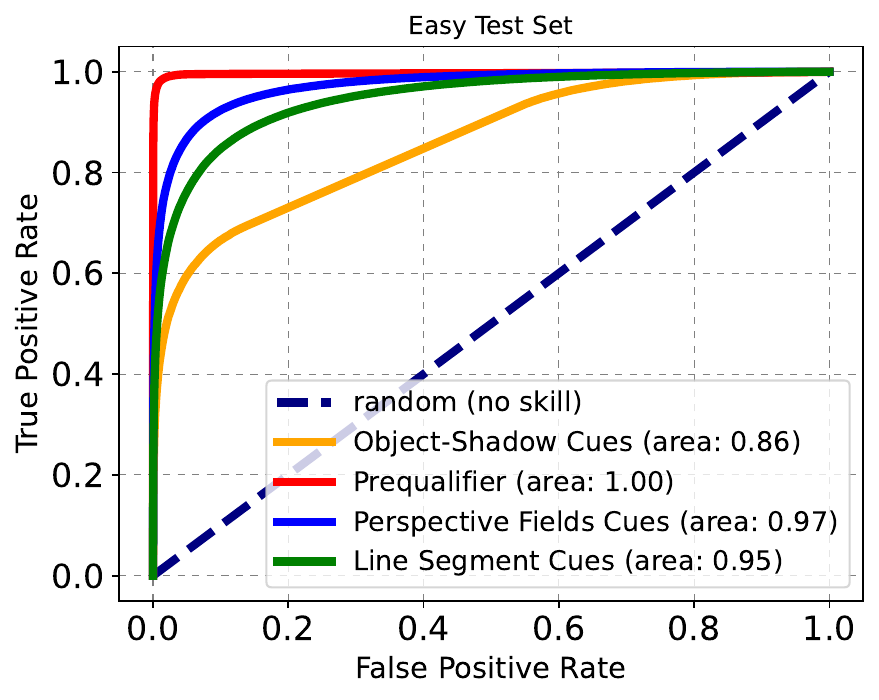}
        \caption{Easy Test Set (Combined)}
        \label{fig:combined_full_test_set}
    \end{subfigure}
    \hfill
    \begin{subfigure}[b]{0.24\textwidth}
        \includegraphics[width=\textwidth]{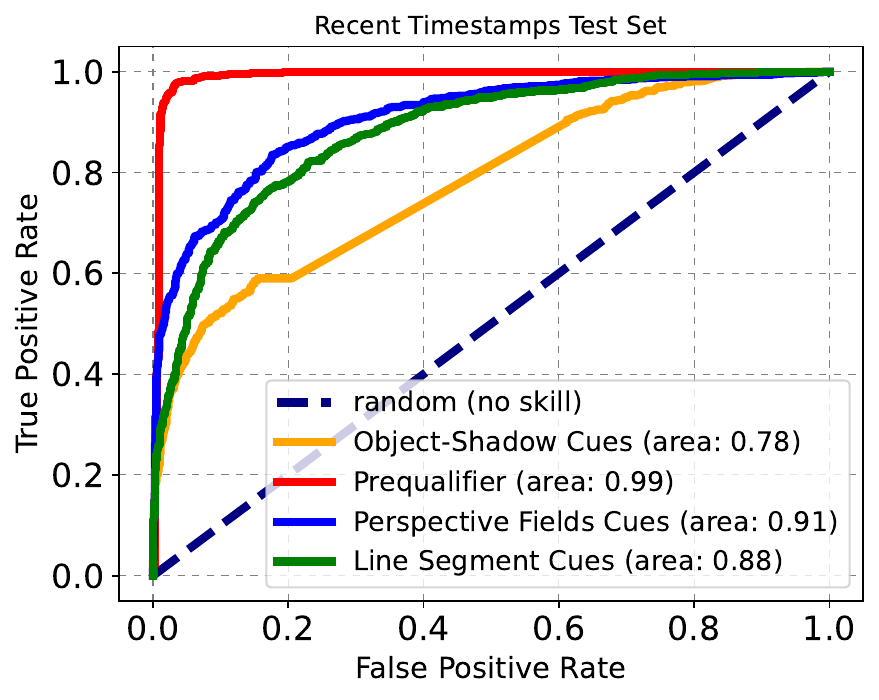}
        \caption{Recent Timestamp Set (Combined)}
        \label{fig:combined_recent_timestamp_test_set}
    \end{subfigure}
    \begin{subfigure}[b]{0.24\textwidth}
        \includegraphics[width=\textwidth]{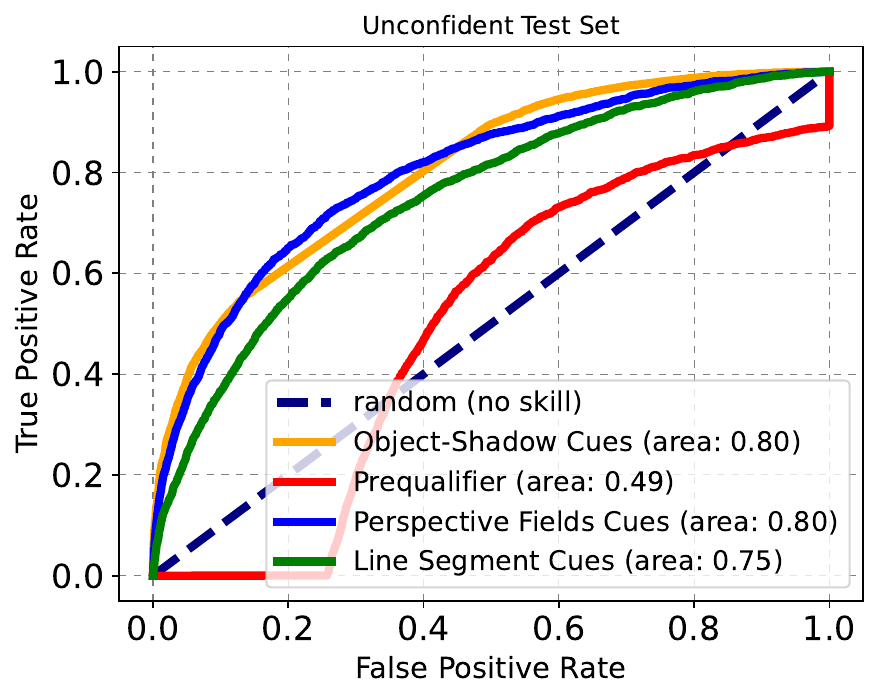}
        \caption{Unconfident Test Set (Combined)}
        \label{fig:combined_unconfident_test_set}
    \end{subfigure}
    \hfill
    \begin{subfigure}[b]{0.24\textwidth}
        \includegraphics[width=\textwidth]{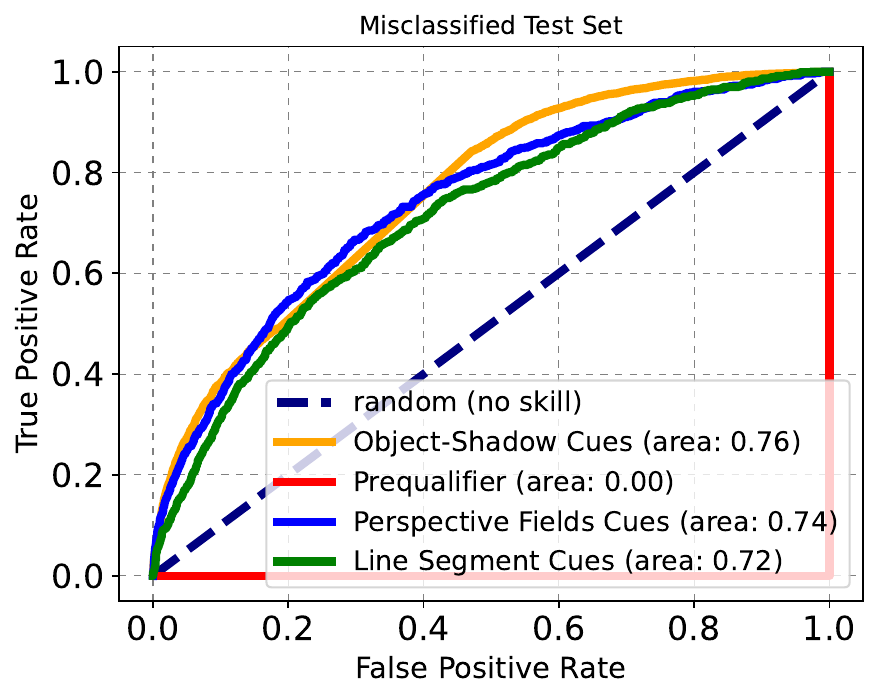}
        \caption{Misclassified Test Set  (Combined)}
        \label{fig:combined_misclassified_test_set}
    \end{subfigure}
    \vspace{-8pt}
    \caption{ROC Curves Assessing Projective Geometry Cues in Generated Images trained on Kandinsky-v3. We trained separate models for indoor scenes (top row), outdoor scenes (middle row), and a combination of indoor and outdoor scenes (last row). All our derived geometry cues classifiers are trained without looking at image intensity information and can reliably detect projective geometry errors. The recent timestamp test set (second column) confirms that these models are robust. We find hard examples using a prequalifier trained on image pixels. Our derived geometry cues consistently show high AUC for finding projective geometry errors on hard test sets -- the last two columns -- unconfident and misclassified test sets. For the unconfident test set, where the prequalifier has an AUC of 0.51 (c), 0.51 (g), and 0.49 (h) for indoor, outdoor, and combined partition, our classifiers can still accurately identify the generated images with high AUCs -- 0.82 from line segments in the indoor set, 0.84 from perspective fields in the outdoor set, and 0.80 from perspective field cues and object shadows in the combined set. Similarly, for the misclassified test set, where the prequalifier has an AUC of 0.00, as it should, our classifiers remain reliable with AUC up to 0.82. We conclude that generated images contain geometric structures not seen in real images, and these structures very reliably identify generated images by only looking at derived geometry cues.
}
    \label{fig:roc_curves}
    \vspace{-10pt}
\end{figure*}

\begin{figure*}[t!]
  \centering
  \footnotesize
  \setlength\tabcolsep{0.2pt}
  \renewcommand{\arraystretch}{0.1}
  \begin{tabular}{ccccccc}
 
    \includegraphics[width=0.141\linewidth]{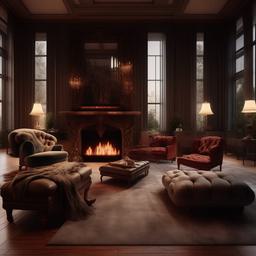} &
    \includegraphics[width=0.141\linewidth]{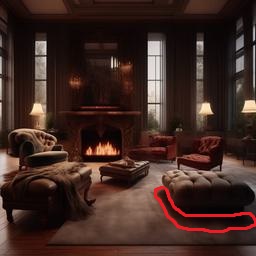} &
    \includegraphics[width=0.141\linewidth]{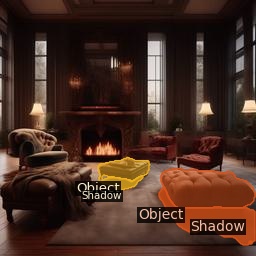} &
    \includegraphics[width=0.141\linewidth]{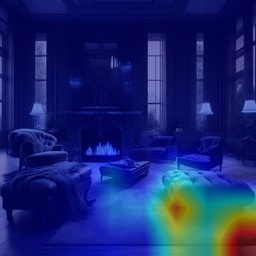} &
    \includegraphics[width=0.141\linewidth]{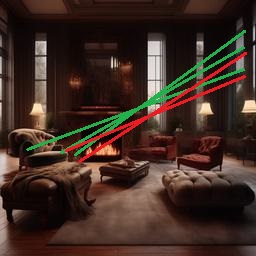} &
    \includegraphics[width=0.141\linewidth]{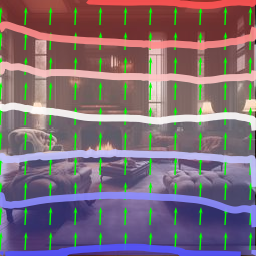} &
    \includegraphics[width=0.141\linewidth]{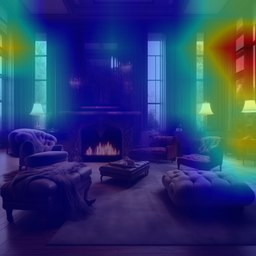} 
\\   
    \includegraphics[width=0.141\linewidth]{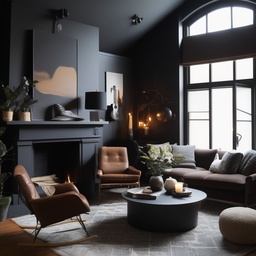} &
    \includegraphics[width=0.141\linewidth]{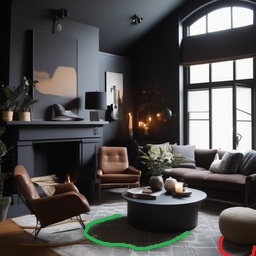} &
    \includegraphics[width=0.141\linewidth]{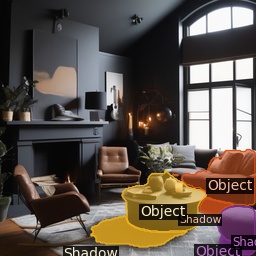} &
    \includegraphics[width=0.141\linewidth]{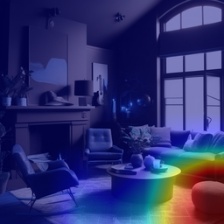} &
    \includegraphics[width=0.141\linewidth]{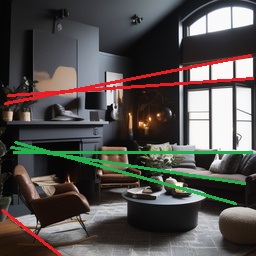} &
    \includegraphics[width=0.141\linewidth]{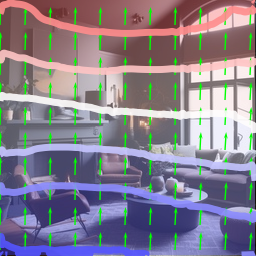} &
    \includegraphics[width=0.141\linewidth]{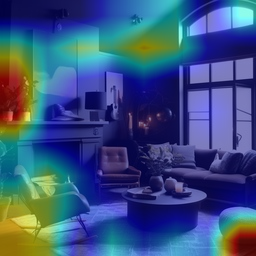} 
\\   
    \includegraphics[width=0.141\linewidth]{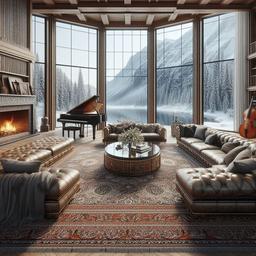} &
    \includegraphics[width=0.141\linewidth]{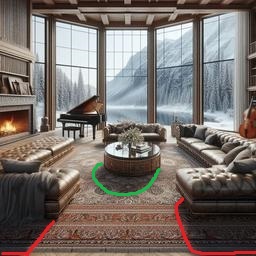} &
    \includegraphics[width=0.141\linewidth]{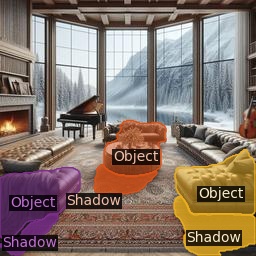} &
    \includegraphics[width=0.141\linewidth]{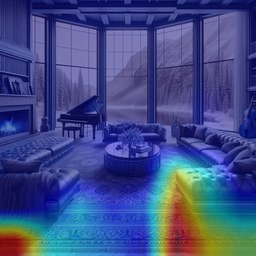} &
    \includegraphics[width=0.141\linewidth]{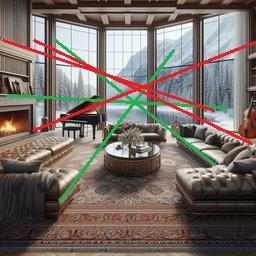} &
    \includegraphics[width=0.141\linewidth]{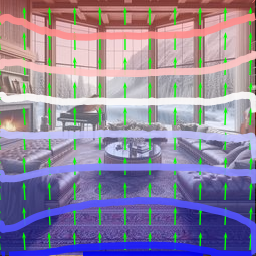} &
    \includegraphics[width=0.141\linewidth]{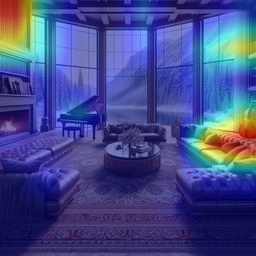} 
\\   
    \vspace{2pt}
  Generated Image & Shadow Errors & Object-Shadow (OS) &  OS GradCam & VP Errors & {\scriptsize Perspective Fields (PF)} & PF GradCam\\
        \end{tabular}
        \vspace{-5pt}
  \caption{Grad-CAM applied to our Object-Shadow and Perspective Field classifiers reveals that the high AUCs in \cref{fig:roc_curves} are based on real geometric errors in indoor scenes generated by Kandinsky, Stable Diffusion XL and Dalle-3 shown in each row respectively. The second and fifth columns highlight shadow and vanishing point errors, respectively. The third column overlays detected object-shadow pairs~\cite{wang2022instance}. Grad-CAM applied to our Object-Shadow classifier (fourth column) identifies diagnostic areas for synthetic generation, such as inconsistent shadow directions (in all three rows), mismatched shadow lengths (second row). The sixth column shows Perspective Fields~\cite{jin2023perspective}, and Grad-CAM applied to our Perspective Fields classifier (last column) reveals geometric errors in all three rows, particularly at ceilings and side walls, with noticeable errors also present in window grills in the first and second rows.
  }
        \label{fig:various_generator_indoor}
	\end{figure*}

\begin{figure*}[t!]
  \centering
  \footnotesize
  \setlength\tabcolsep{0.2pt}
  \renewcommand{\arraystretch}{0.1}
  \begin{tabular}{ccccccc}

    \includegraphics[width=0.141\linewidth]{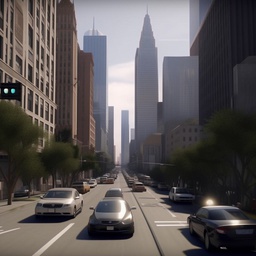} &
    \includegraphics[width=0.141\linewidth]{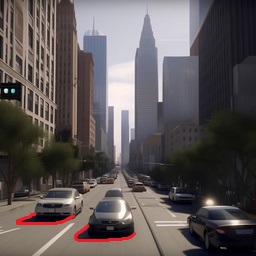} &
    \includegraphics[width=0.141\linewidth]{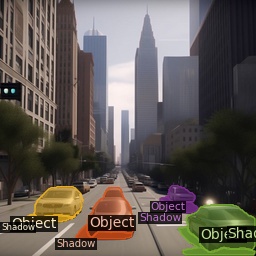} &
    \includegraphics[width=0.141\linewidth]{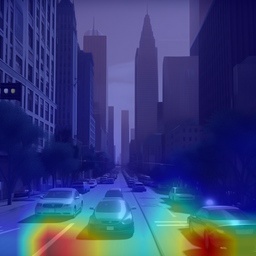} &
    \includegraphics[width=0.141\linewidth]{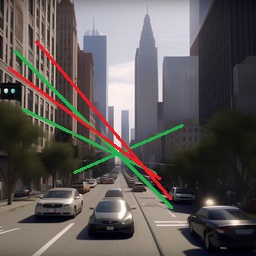} &
    \includegraphics[width=0.141\linewidth]{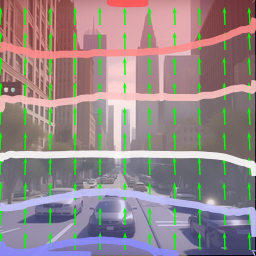} &
    \includegraphics[width=0.141\linewidth]{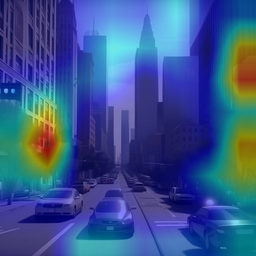} 
\\   
    \includegraphics[width=0.141\linewidth]{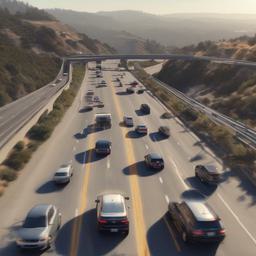} &
    \includegraphics[width=0.141\linewidth]{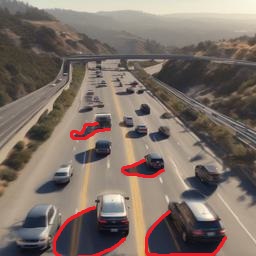} &
    \includegraphics[width=0.141\linewidth]{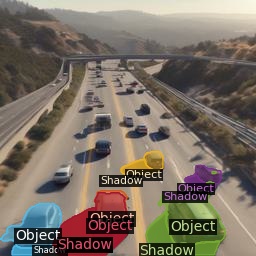} &
    \includegraphics[width=0.141\linewidth]{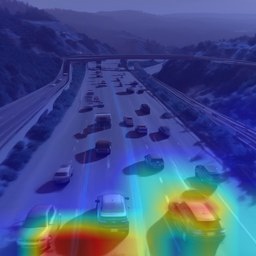} &
    \includegraphics[width=0.141\linewidth]{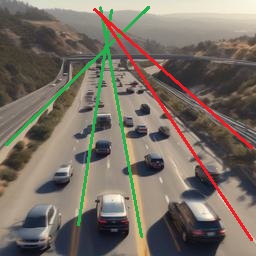} &
    \includegraphics[width=0.141\linewidth]{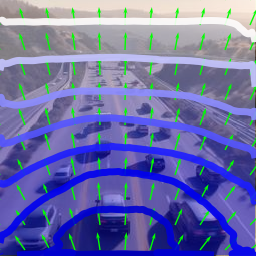} &
    \includegraphics[width=0.141\linewidth]{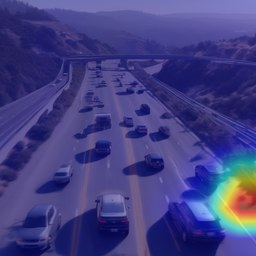} 
\\   
    \includegraphics[width=0.141\linewidth]{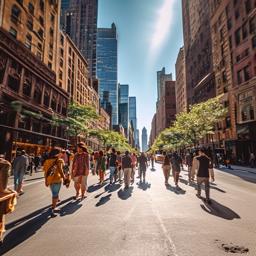} &
    \includegraphics[width=0.141\linewidth]{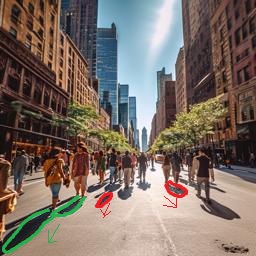} &
    \includegraphics[width=0.141\linewidth]{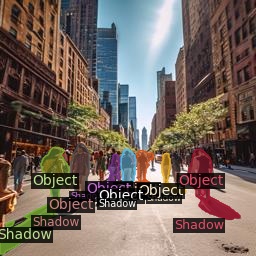} &
    \includegraphics[width=0.141\linewidth]{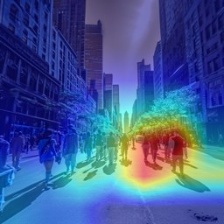} &
    \includegraphics[width=0.141\linewidth]{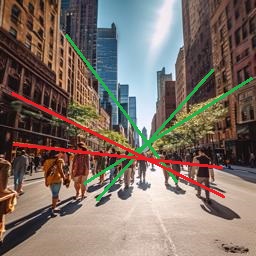} &
    \includegraphics[width=0.141\linewidth]{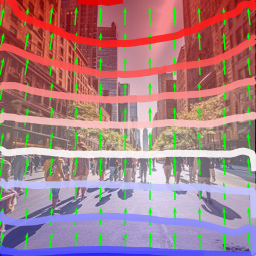} &
    \includegraphics[width=0.141\linewidth]{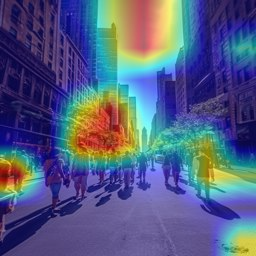} 
\\       \vspace{2pt}
\\
  Generated Image & Shadow Errors & Object-Shadow (OS) &  OS GradCam & VP Errors & Perspective Fields & PF GradCam\\
        \end{tabular}
        \vspace{-5pt}
  \caption{Grad-CAM results for outdoor scenes generated by Kandinsky, Stable Diffusion XL, and Adobe Firefly, shown in each row respectively. The second column highlights shadow errors, while the third column overlays detected object-shadow pairs~\cite{wang2022instance}. Grad-CAM applied to our Object-Shadow classifier (fourth column) reveals incorrect shadow shapes in the first and second rows, with shadows on the right-side pedestrians pointing in a different direction than those on the left. The fifth column shows vanishing point errors, and the sixth column presents Perspective Fields~\cite{jin2023perspective}. Grad-CAM applied to our Perspective Fields classifier (last column) confirms large perspective distortions on building facades and road markings, corroborating the vanishing point errors in the fifth column.
  }
        \label{fig:various_generator_outdoor}
        \vspace{-3pt}
	\end{figure*}

 \begin{figure*}[t!]
  \centering
  \footnotesize
  \setlength\tabcolsep{0.2pt}
  \renewcommand{\arraystretch}{0.1}
  \begin{tabular}{ccccccc}

\\
    \includegraphics[width=0.141\linewidth]{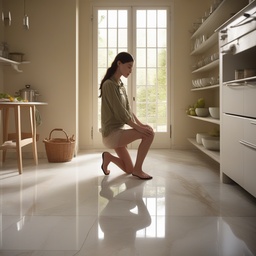} &
    \includegraphics[width=0.141\linewidth]{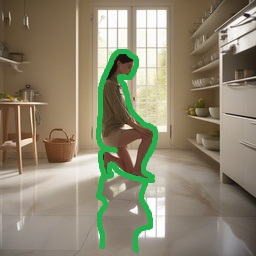} &
    \includegraphics[width=0.141\linewidth]{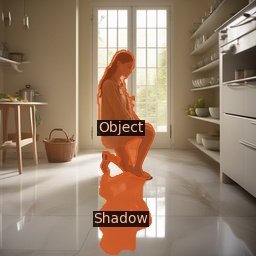} &
    \includegraphics[width=0.141\linewidth]{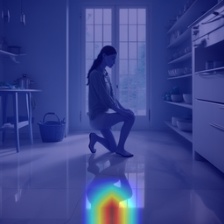} &
    \includegraphics[width=0.141\linewidth]{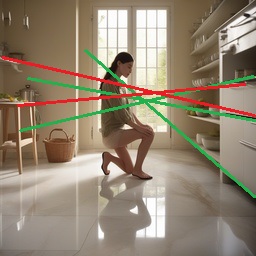} &
    \includegraphics[width=0.141\linewidth]{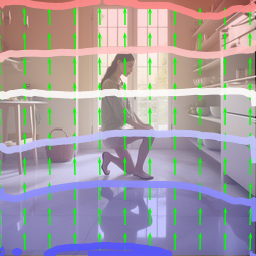} &
    \includegraphics[width=0.141\linewidth]{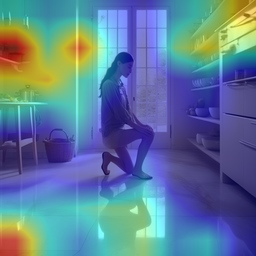}     \vspace{1pt}

\\
    \includegraphics[width=0.141\linewidth]{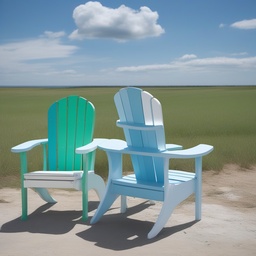} &
    \includegraphics[width=0.141\linewidth]{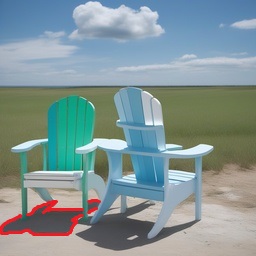} &
    \includegraphics[width=0.141\linewidth]{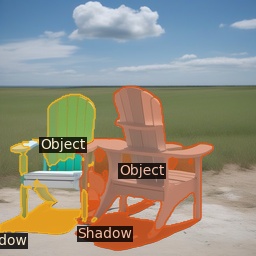} &
    \includegraphics[width=0.141\linewidth]{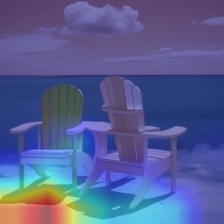} &
    \includegraphics[width=0.141\linewidth]{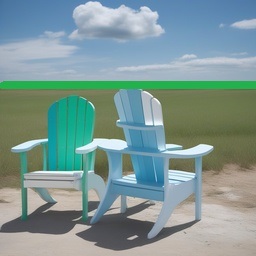} &
    \includegraphics[width=0.141\linewidth]{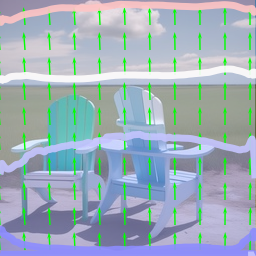} &
    \includegraphics[width=0.141\linewidth]{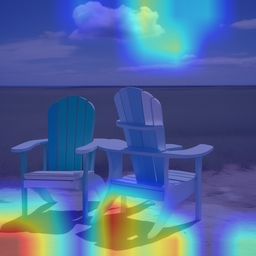}      \vspace{2pt}
    \\

  Generated Image & Shadow Errors & Object-Shadow (OS) &  OS GradCam & VP Errors & {\scriptsize Perspective Fields (PF)} & PF GradCam\\
        \end{tabular}
                \vspace{-5pt}
  \caption{Our projective geometry classifiers identify distinct types of problems in generated images. The top row presents an example that was classified as real by the Object-Shadow classifier but correctly identified as generated by the Perspective Fields classifier. While the shadow cast by the person appears realistic, the Perspective Fields Grad-CAM highlights the problematic geometry of the shelf on the top left. In contrast, the bottom row shows an example that was correctly identified as generated by the Object-Shadow classifier but misclassified as real by the Perspective Fields classifier. Although the perspective effects in the image appear plausible, the Grad-CAM weights correctly reveal that the two chairs are casting shadows from different light sources, indicating inconsistency in scene's illumination.
  }
        \label{fig:grad_cam_complement}
        \vspace{-2pt}
	\end{figure*}

\begin{figure*}[t!]
\scriptsize
    \centering
       \begin{subfigure}[b]{0.19\textwidth}
        \includegraphics[width=\textwidth]{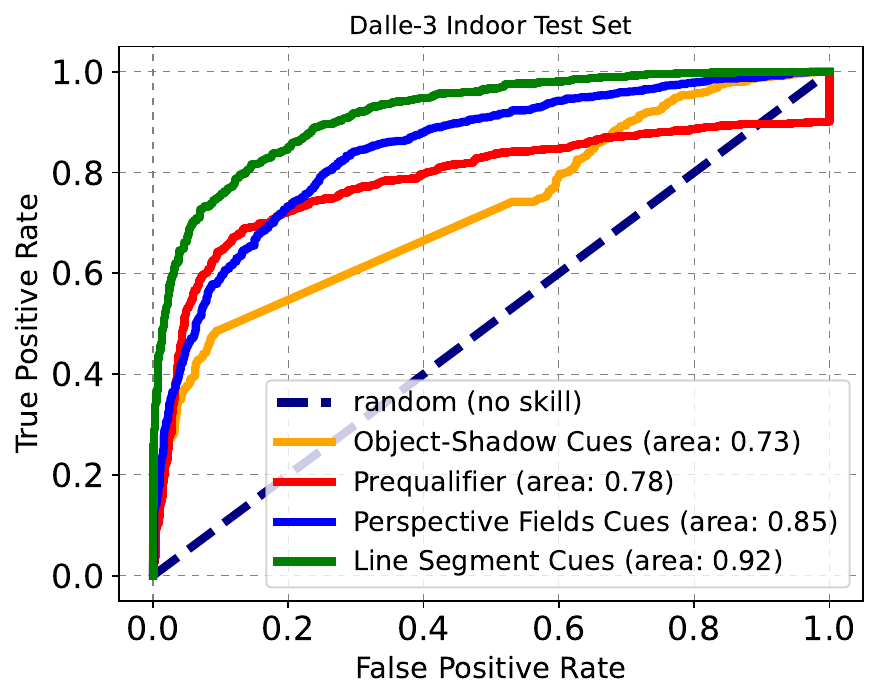}
        \caption{Dall-E 3 (Indoor)}
        \label{fig:dalle-indoor}
    \end{subfigure}
    \hfill
    \begin{subfigure}[b]{0.19\textwidth}
        \includegraphics[width=\textwidth]{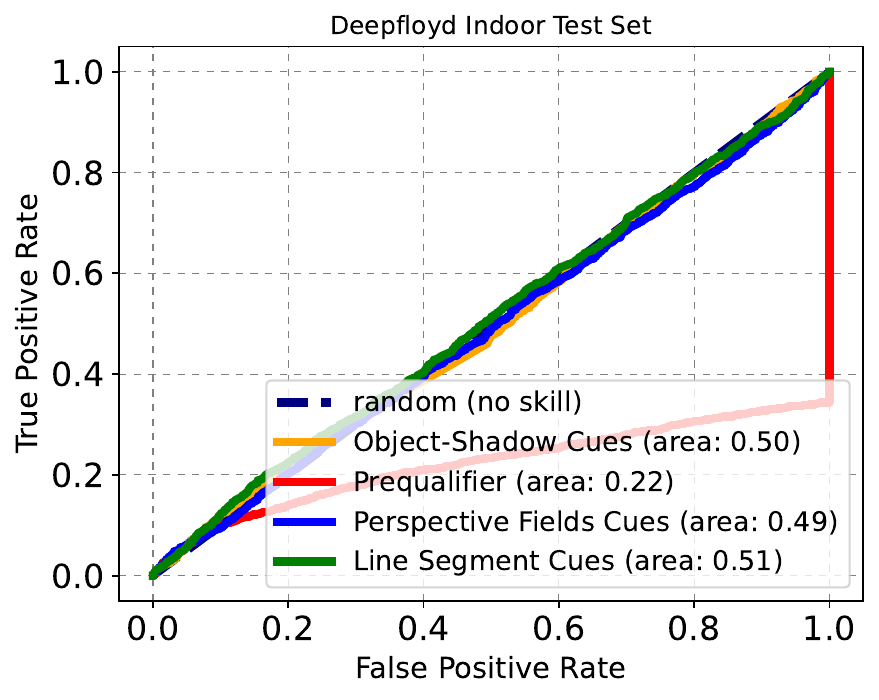}
        \caption{DeepFloyd (Indoor)}
        \label{fig:deepfloyd-indoor}
    \end{subfigure}
    \hfill
     \begin{subfigure}[b]{0.19\textwidth}
        \includegraphics[width=\textwidth]{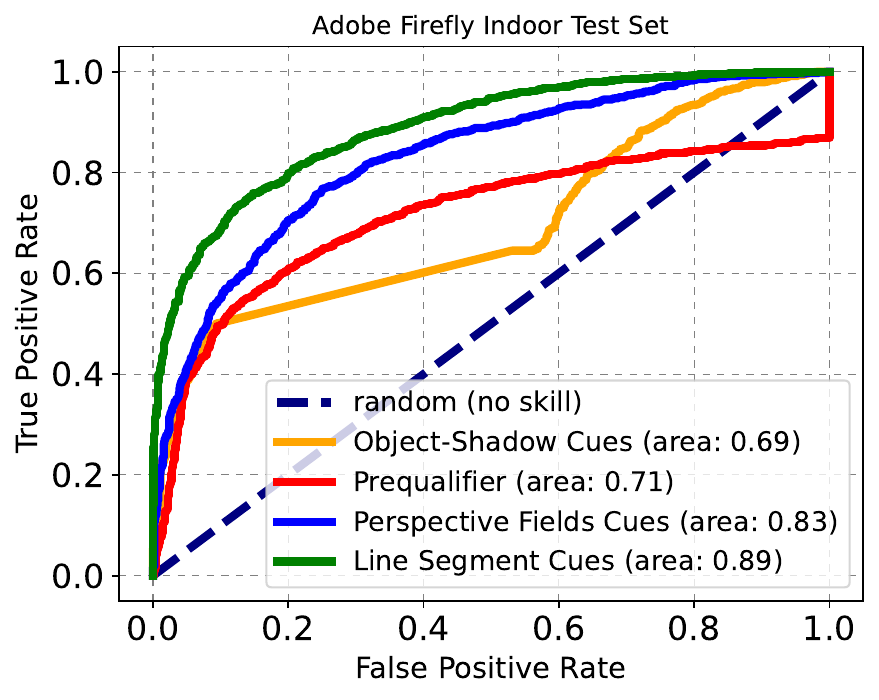}
        \caption{Adobe Firefly (Indoor)}
        \label{fig:firefly-indoor}
    \end{subfigure}
    \begin{subfigure}[b]{0.19\textwidth}
        \includegraphics[width=\textwidth]{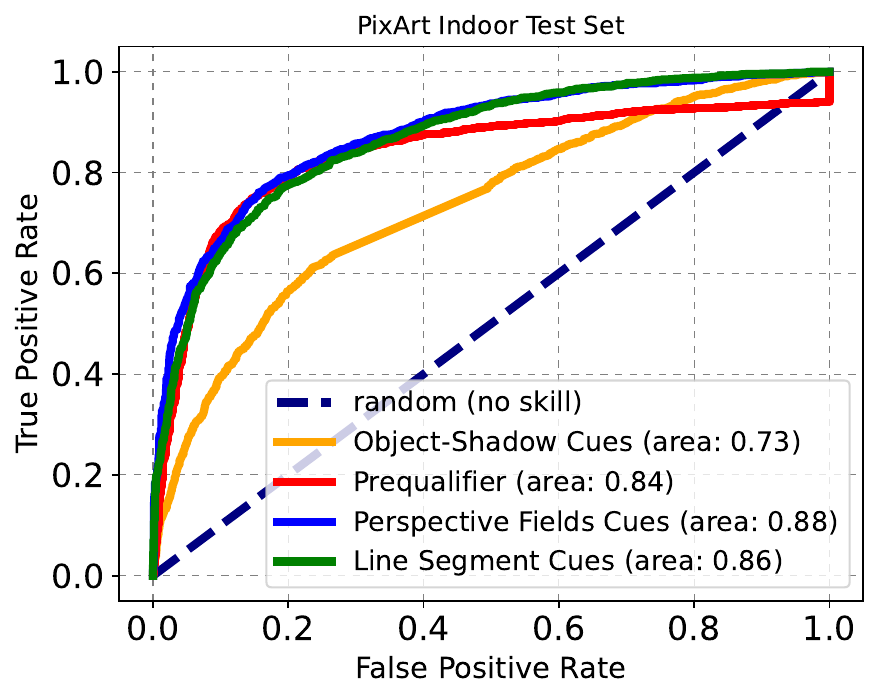}
        \caption{PixArt-$\alpha$ (Indoor)}
        \label{fig:pixart-indoor}
    \end{subfigure}
    \begin{subfigure}[b]{0.19\textwidth}
        \includegraphics[width=\textwidth]{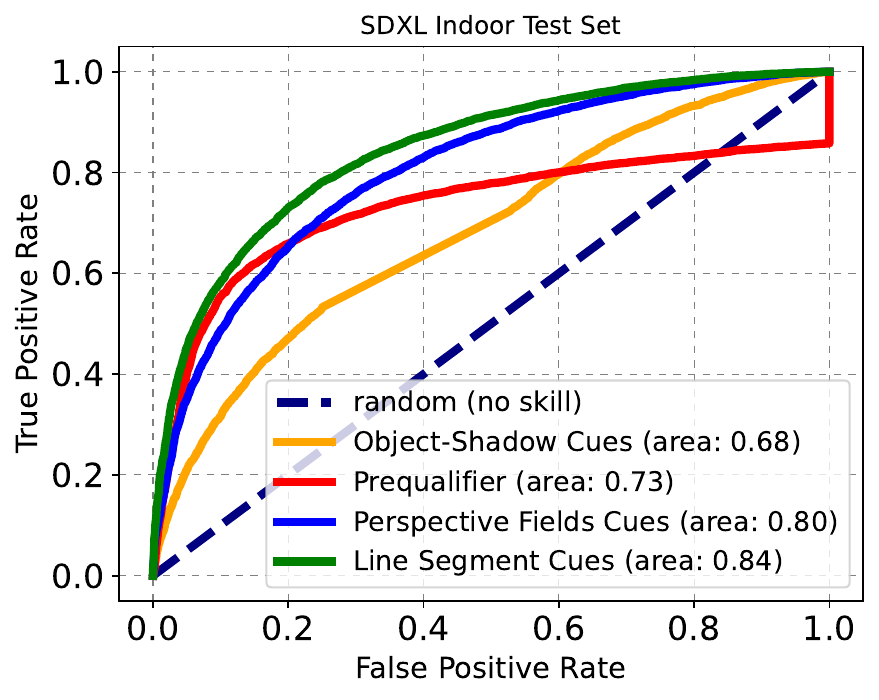}
        \caption{SDXL (Indoor)}
        \label{fig:sdxlindoor}
        \end{subfigure}
        \begin{subfigure}[b]{0.19\textwidth}
        \includegraphics[width=\textwidth]{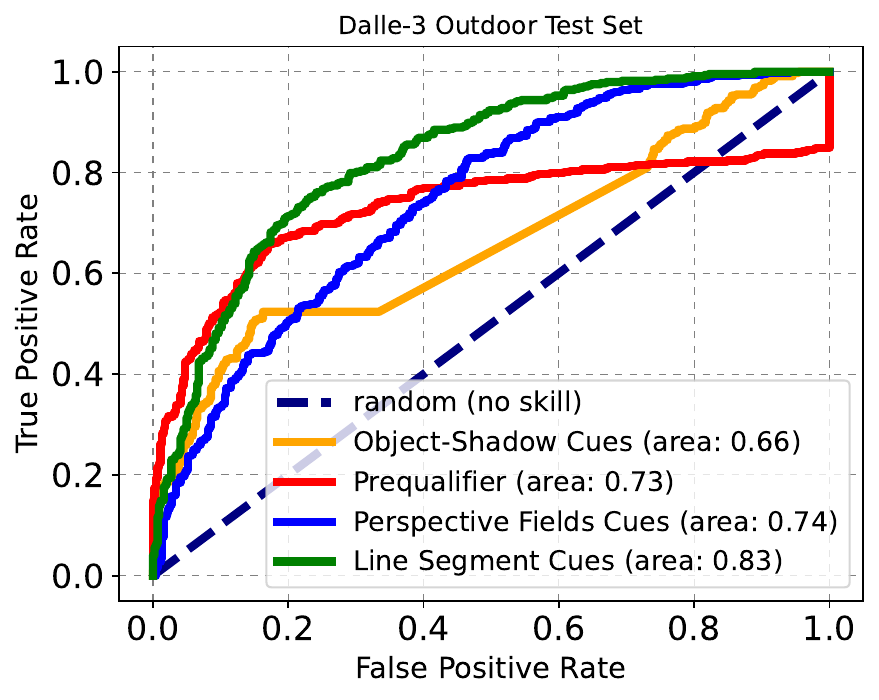}
        \caption{Dall-E 3 (Outdoor)}
        \label{fig:dalle-outdoor}
    \end{subfigure}
    \hfill
     \begin{subfigure}[b]{0.19\textwidth}
          \includegraphics[width=\textwidth]{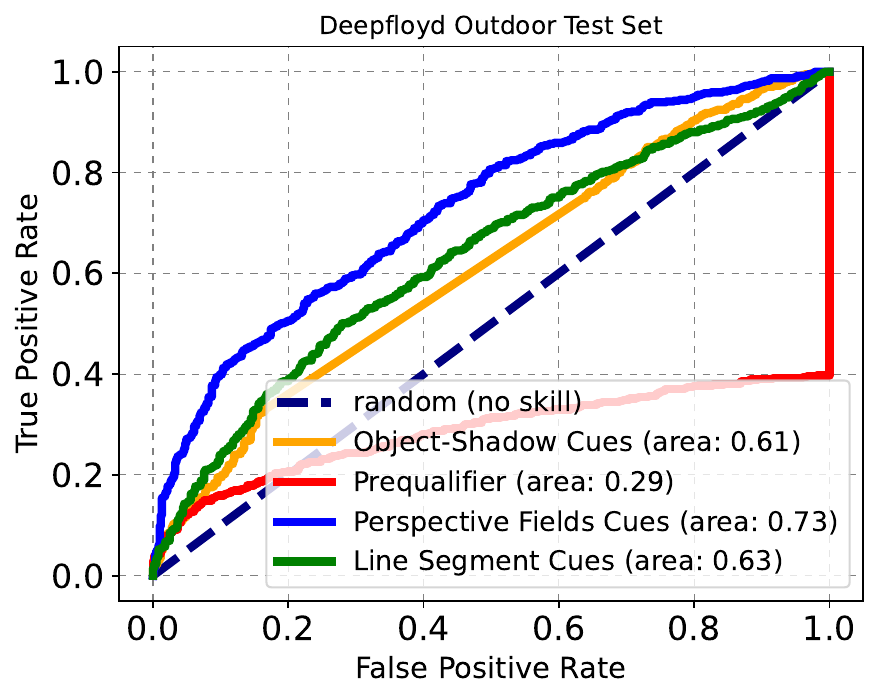}
        \caption{DeepFloyd (Outdoor)}
        \label{fig:deepfloyd-outdoor}
    \end{subfigure}
    \hfill
    \begin{subfigure}[b]{0.19\textwidth}
        \includegraphics[width=\textwidth]{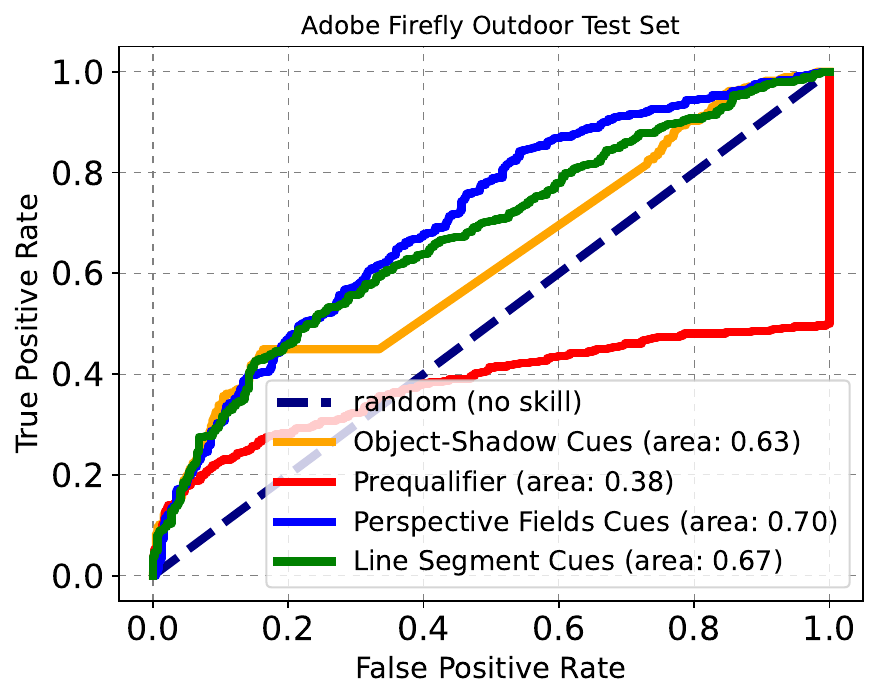}
        \caption{Adobe Firefly (Outdoor)}
        \label{fig:firefly-outdoor}
    \end{subfigure}
    \hfill
    \begin{subfigure}[b]{0.19\textwidth}
        \includegraphics[width=\textwidth]{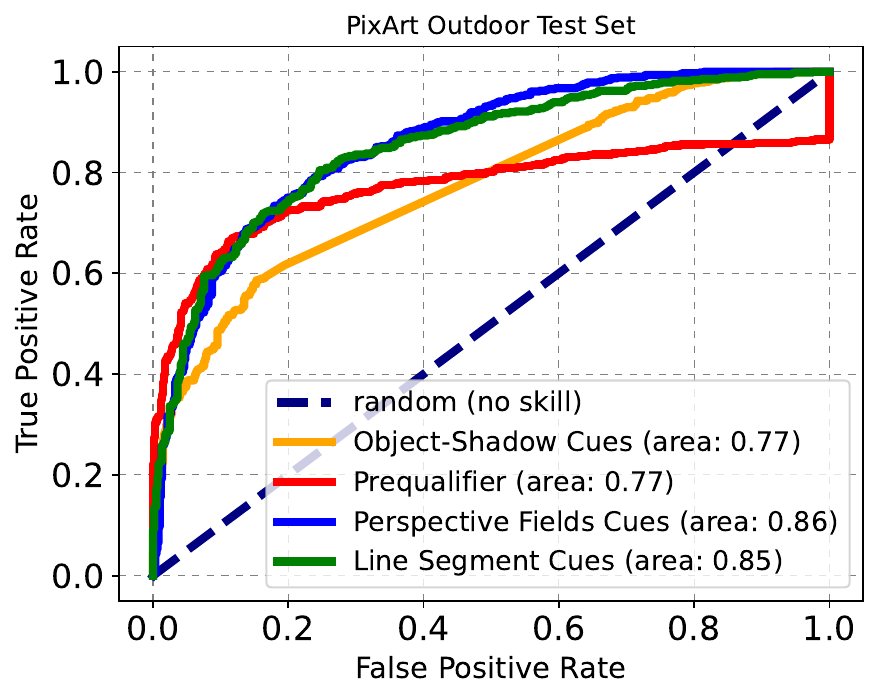}
        \caption{PixArt-$\alpha$ (Outdoor)}
        \label{fig:pixart-outdoor}
    \end{subfigure}
    \hfill
    \begin{subfigure}[b]{0.19\textwidth}
        \includegraphics[width=\textwidth]{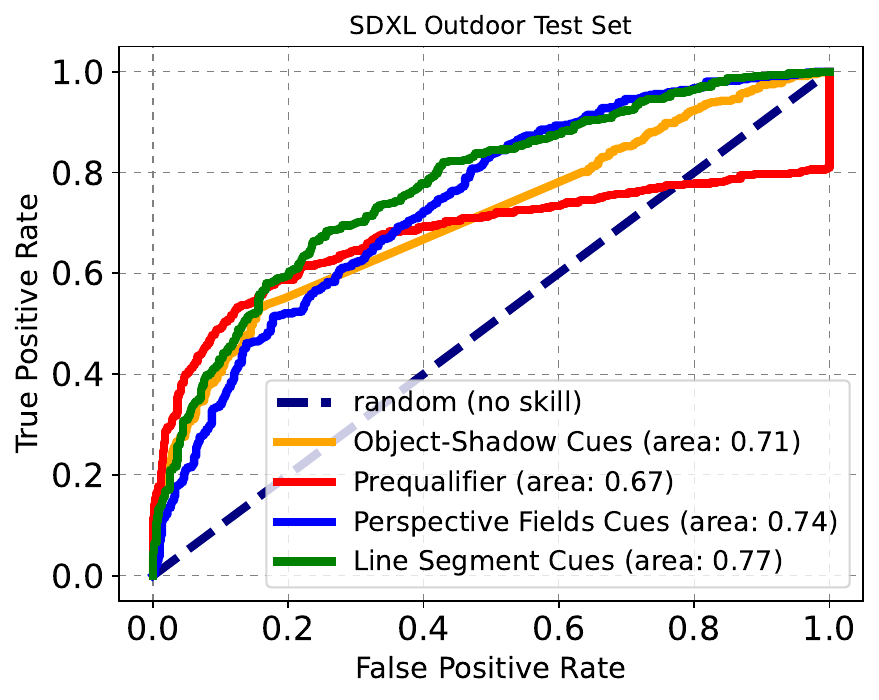}
        \caption{SDXL (Outdoor)}
        \label{fig:sdxl-outdoor}
    \end{subfigure}
    \vspace{-5pt}
    \caption{Our model trained on Kandinsky-v3 generalizes in detecting projective geometry errors from images generated by various unseen models. We evaluate the model's performance on test sets generated by Dall-E 3 (a), DeepFloyd (b), Adobe Firefly (c), PixArt-$\alpha$ (d), and Stable Diffusion XL or SDXL (e) using the same text prompts from the `misclassified' Kandinsky-v3 generated test set. The top row shows indoor scenes and the last row shows outdoor scenes. Our model generalizes to all evaluated generators except DeepFloyd. However, DeepFloyd-generated images can be reliably detected when the model is trained on a DeepFloyd-generated training set, but it shows poor generalization capabilities to other generators compared to Kandinsky. 
}
    \label{fig:roc_curves_other_generators}
    \vspace{-8pt}
\end{figure*}

\section{Evaluation}

Based on our prequalification analysis, we use Kandinsky as the primary source for our generated data to train our classifiers. We evaluated our three geometry-derived cues classifiers and analyzed their ability to generalize to other generated images that were not used during training. In addition, we performed a GradCam analysis~\cite{Selvaraju_2019} to identify any potential geometric discrepancies in images.

\subsection{Classifiers Results}

\cref{fig:roc_curves} shows ROC curves for each of our methods on indoor, outdoor and
combined (indoor+outdoor) scenes.  In each case, classifiers are trained on images that
are {\em not} prequalified and tested on prequalified scenes, meaning that performance
estimates are biased {\em low} --- likely training on prequalified data would lead to even
more accurate classification.  Each classifier is effective, with AUCs ranging from 0.72
to 0.97.  Recall these classifiers see {\em only} derived geometric features and do not see the image itself.

Qualitative examples using Grad-CAM appear in \cref{fig:various_generator_outdoor}
and \cref{fig:various_generator_indoor}.   Notice how images that might be acceptable to a line
analysis often fail a shadow analysis.  \cref{fig:grad_cam_complement} shows examples to
emphasize this point.

\section{Other Generators Evaluated}
Our classifiers do not see pixels, but derived geometric features.  This means that one
could expect a form of generalization across generators.  We illustrate that this generalization occurs - 
ROC curves in \cref{fig:roc_curves_other_generators} demonstrate that classifiers trained
to distinguish Kandinsky images from real images can also reliably distinguish Stable DIffusion XL~\cite{arkhipkin2023kandinsky}, DeepFloyd~\cite{deepfloyd2023} and PixArt-$\alpha$\cite{chen2023pixartalpha} from the open-source domain. Additionally, we assess the efficacy of our models against images from proprietary generators such as OpenAI's Dalle-3\cite{betker2023improving} and Adobe's Firefly~\cite{adobefirefly2023}, representing some of the most advanced tools in image generation. 
Finally, we show we can detect composite made by a recent SOTA method~\cite{michel2023object} by looking at Object-Shadow cues in the supplement.

%% file: sec/6_Discussion.tex
\section{Discussion}
We have shown that generated images can be reliably distinguished from real images by looking only at derived geometric cues. This is likely because image generators do not fully implement the geometry one observes in real images. Producing accurate perspective geometry or accurate shadow geometry requires very tight coordination of detailed information over very long spatial scales. Our results, together with the notorious tendency of face image generators to award subjects' left and right earlobes of different shapes, suggest that doing so is beyond the capacity of current generators. Our findings have significant implications for the development of image generation models, as the inability to accurately replicate projective geometry extends across various state-of-the-art models, indicating a widespread issue rather than a problem specific to a particular generator.

We speculate that fixing this difficulty requires structural innovation in the generator, rather than simply exposing the generator to more data. The complex interplay of light, shadows, and perspective in real-world scenes may necessitate novel approaches to modeling and encoding spatial relationships within the generator's architecture. Potential avenues for improvement could include incorporating explicit geometric reasoning or developing new loss functions that prioritize the consistency of projective geometry. Furthermore, our work highlights the importance of developing robust evaluation metrics for image generation models that assess the geometric coherence of generated images. By tackling these challenges, we could create image-generation models that more faithfully capture the complex geometric relationships in real-world scenes.

%% file: sec/X_suppl.tex
\clearpage
\maketitlesupplementary

\begin{figure}[htpb!]
  \centering
  \footnotesize
  \setlength\tabcolsep{0.2pt}
  \renewcommand{\arraystretch}{0.1}
  \begin{tabular}{ccccc}
    \includegraphics[width=0.2\linewidth]{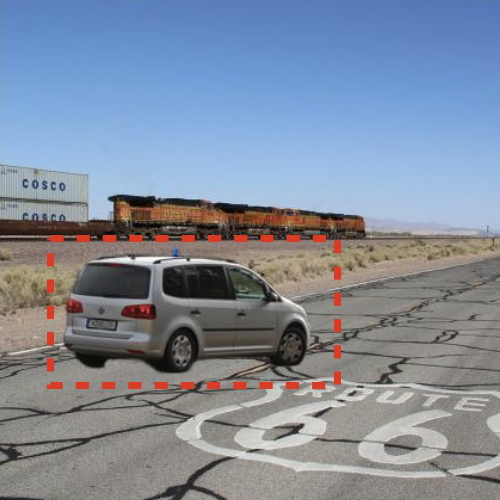} &
    \includegraphics[width=0.2\linewidth]{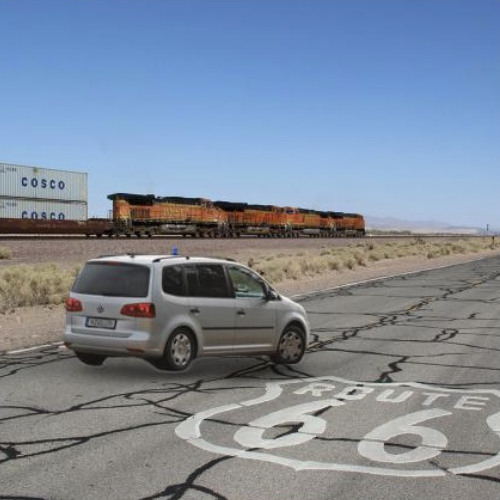} &
    \includegraphics[width=0.2\linewidth]{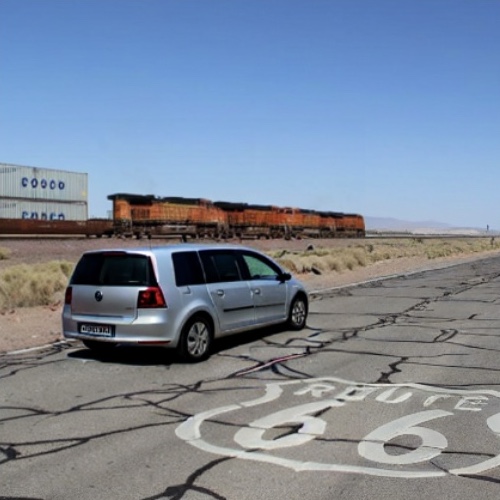} &
    \includegraphics[width=0.2\linewidth]{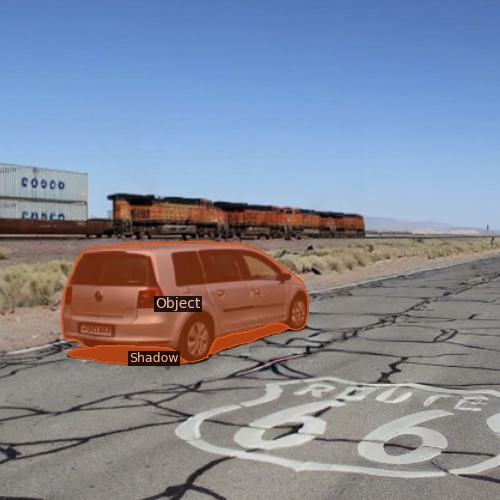} &
    \includegraphics[width=0.2\linewidth]{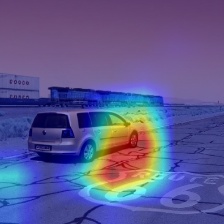} 
\\
    \includegraphics[width=0.2\linewidth]{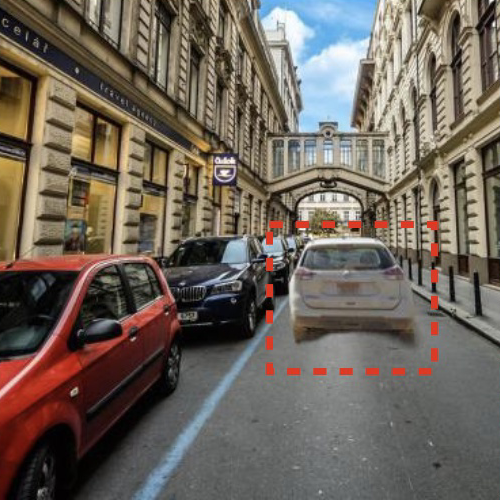} &
    \includegraphics[width=0.2\linewidth]{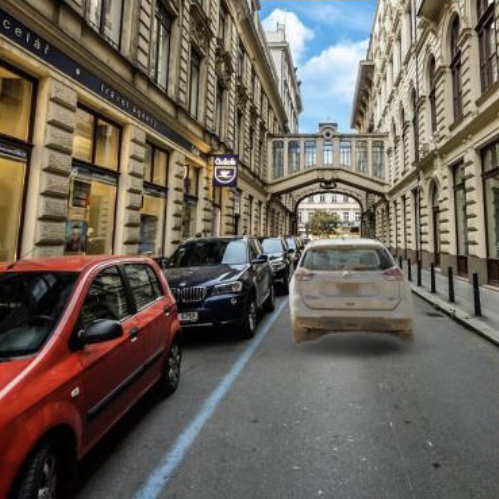} &
    \includegraphics[width=0.2\linewidth]{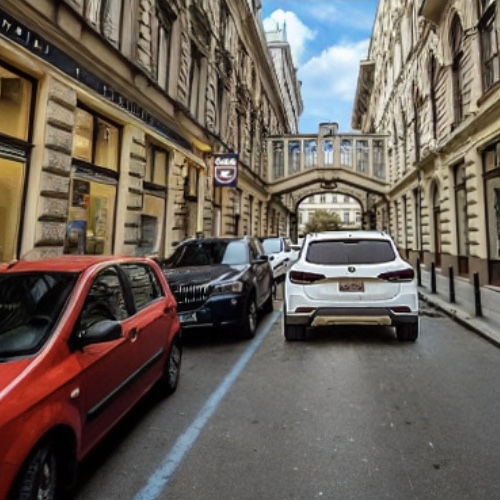} &
    \includegraphics[width=0.2\linewidth]{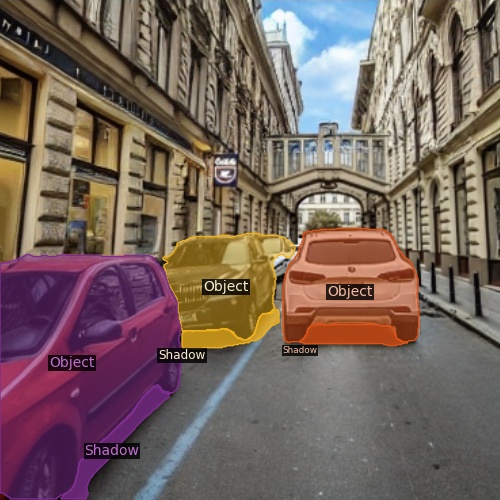} &
    \includegraphics[width=0.2\linewidth]{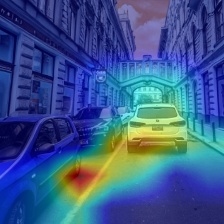} 
\\
    \includegraphics[width=0.2\linewidth]{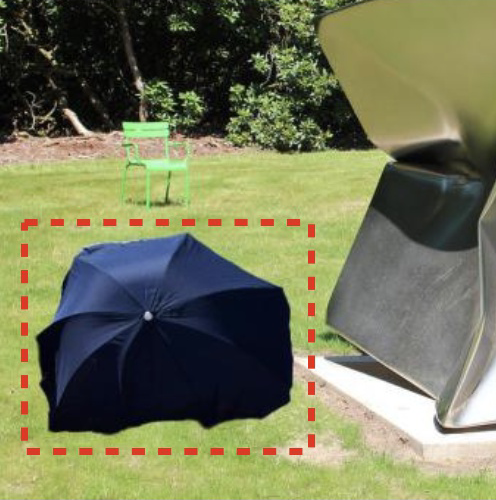} &
    \includegraphics[width=0.2\linewidth]{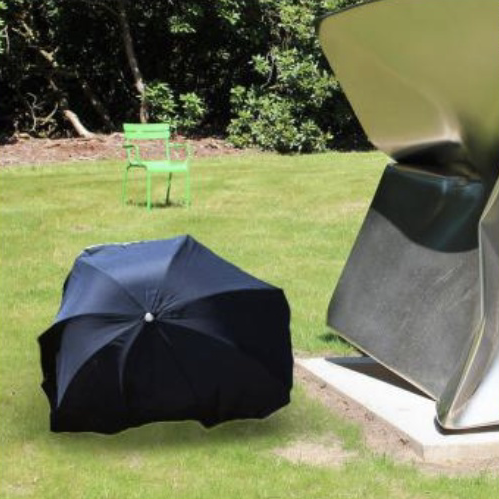} &
    \includegraphics[width=0.2\linewidth]{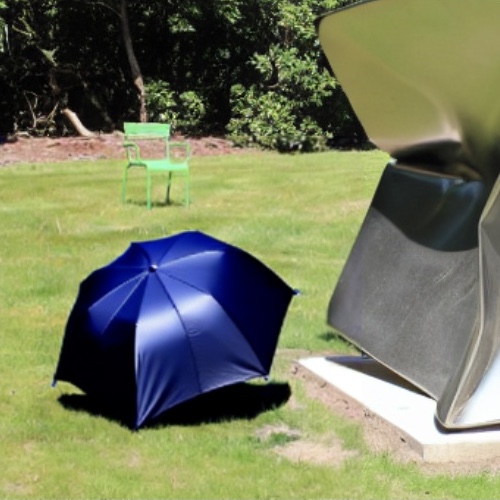} &
    \includegraphics[width=0.2\linewidth]{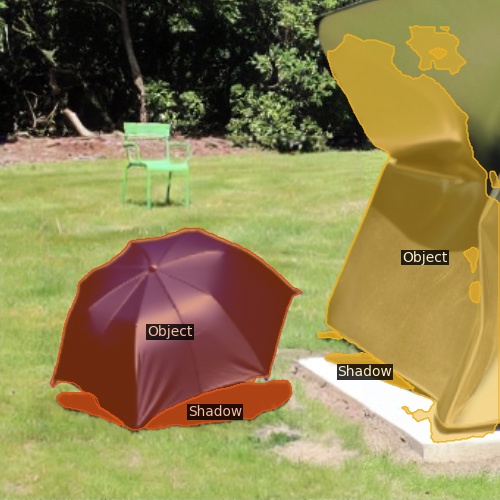} &
    \includegraphics[width=0.2\linewidth]{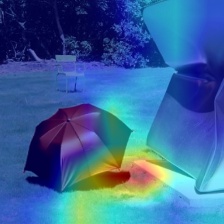} 
\\
  Copy-Paste & \thead{Traditional \\ Composites}  & \thead{ObjectStitch \\ Composites} & \thead{Object-\\ Shadow (OS)} &  OS GradCam 
        \end{tabular}
        \vspace{-10pt}
  \caption{Detecting Composite Errors with Object-Shadow (OS) Cues. We show images directly taken from Figure 1 (teaser) of ~\cite{Song_2023_CVPR} can be detected as generated using our object-shadow classifier.
  The bottom row provides a clear example where, despite the sun being positioned behind the camera, the shadows are mistakenly cast to the right. Also see the shadow of the adjacent object (marked in yellow), which is pointing upward in the opposite direction. Similarly, in the top row, shadows are cast in an implausible direction. The OS GradCam visualizations on the right successfully highlight these misdirected shadows.
  }
        \label{fig:composite_gradcam}
        \vspace{-10pt}
	\end{figure}
\begin{figure*}[t!]
\scriptsize
    \centering
       \begin{subfigure}[b]{0.24\textwidth}
        \includegraphics[width=\textwidth]{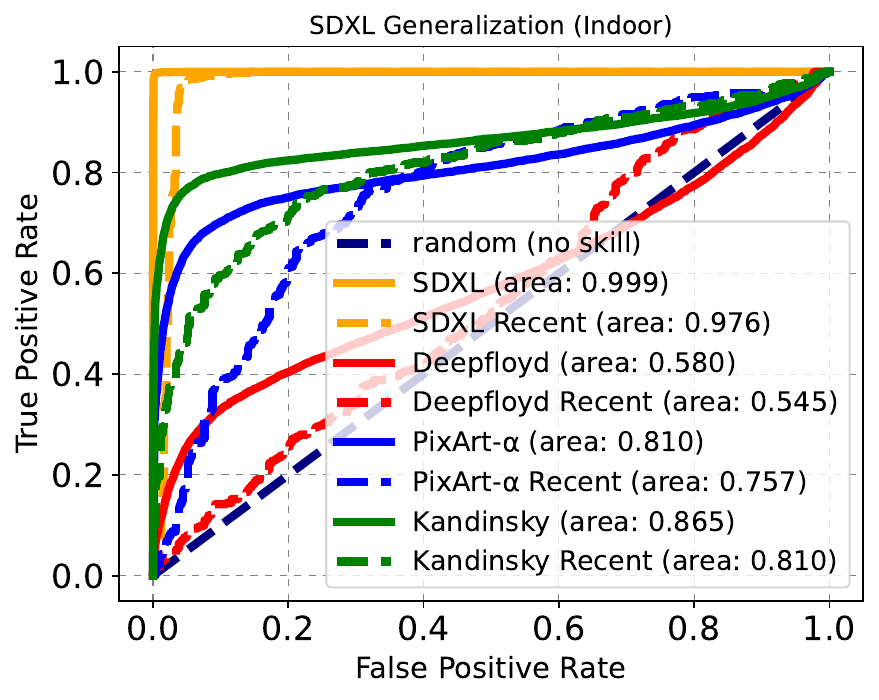}
        \caption{SDXL (Indoor)}
        \label{fig:dalle-indoor-recent}
    \end{subfigure}
    \hfill
    \begin{subfigure}[b]{0.24\textwidth}
        \includegraphics[width=\textwidth]{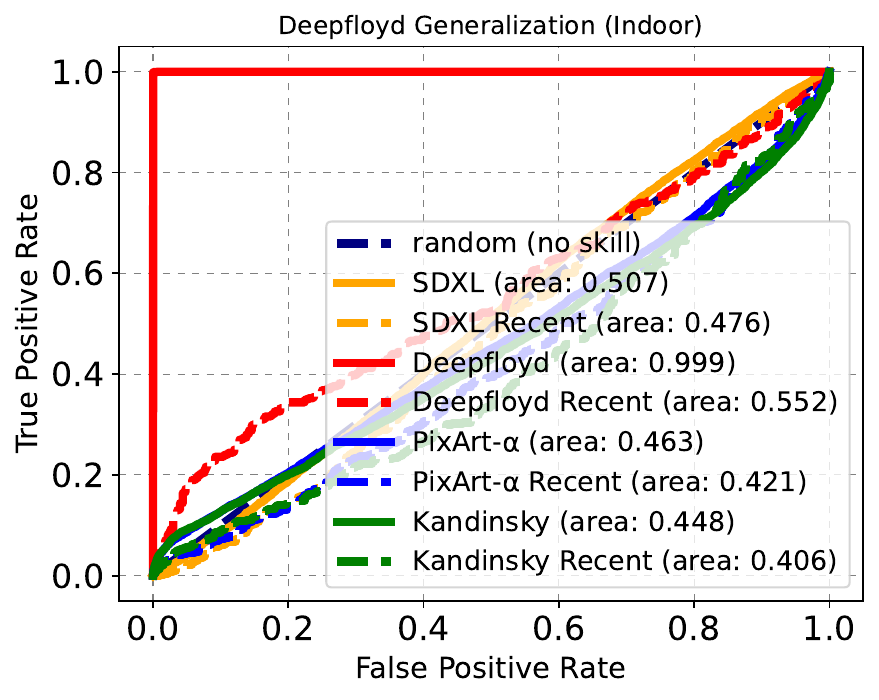}
        \caption{Deepfloyd (Indoor)}
        \label{fig:deepfloyd-indoor-recent}
    \end{subfigure}
    \hfill
     \begin{subfigure}[b]{0.24\textwidth}
        \includegraphics[width=\textwidth]{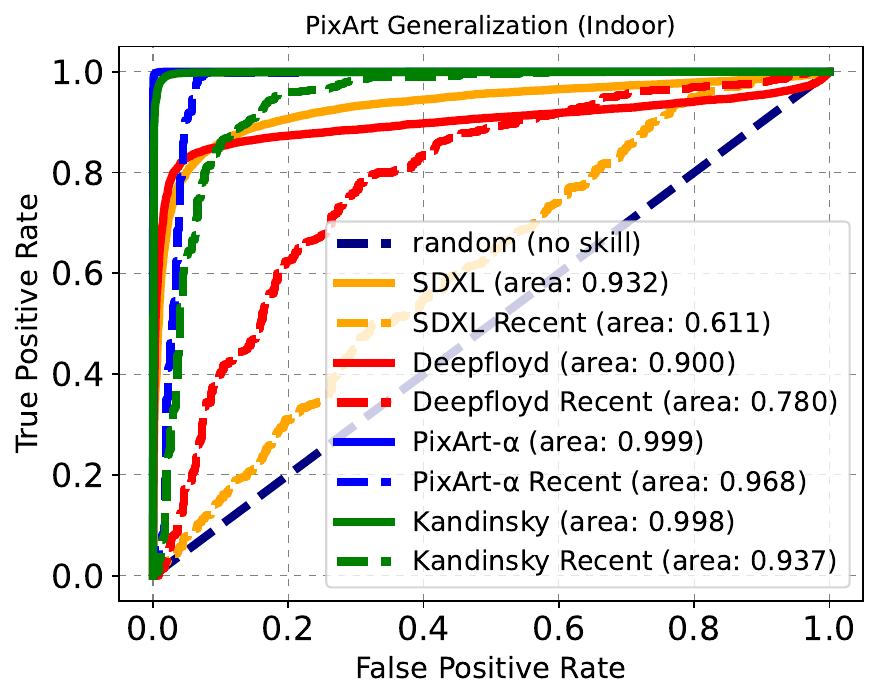}
        \caption{PixArt-$\alpha$ (Indoor)}
        \label{fig:pixart-indoor-recent}
    \end{subfigure}
    \begin{subfigure}[b]{0.24\textwidth}
        \includegraphics[width=\textwidth]{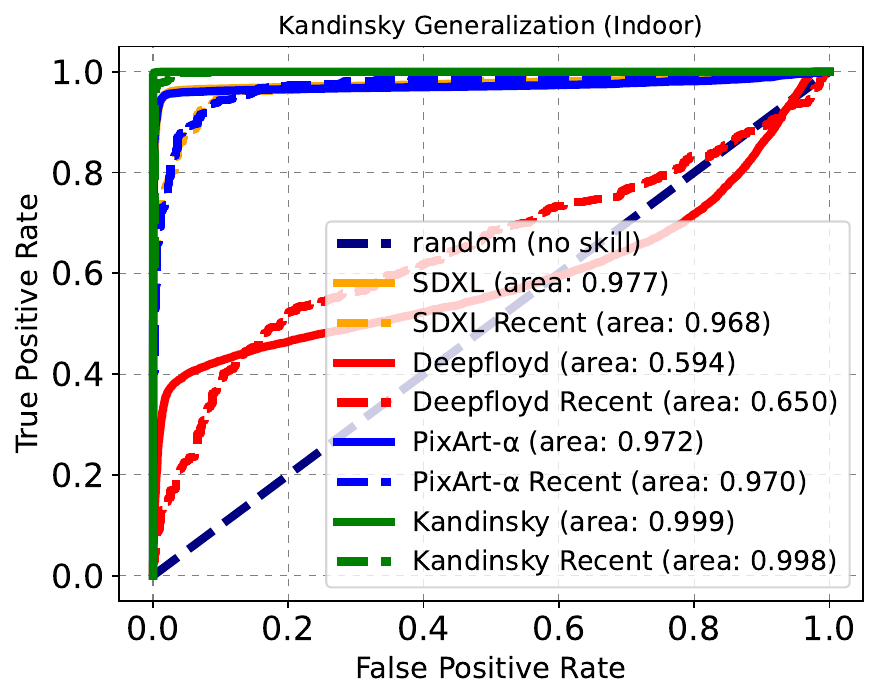}
        \caption{Kandinsky-v3 (Indoor)}
        \label{fig:firefly-indoor-recent}
    \end{subfigure}
    \hfill
     \begin{subfigure}[b]{0.24\textwidth}
          \includegraphics[width=\textwidth]{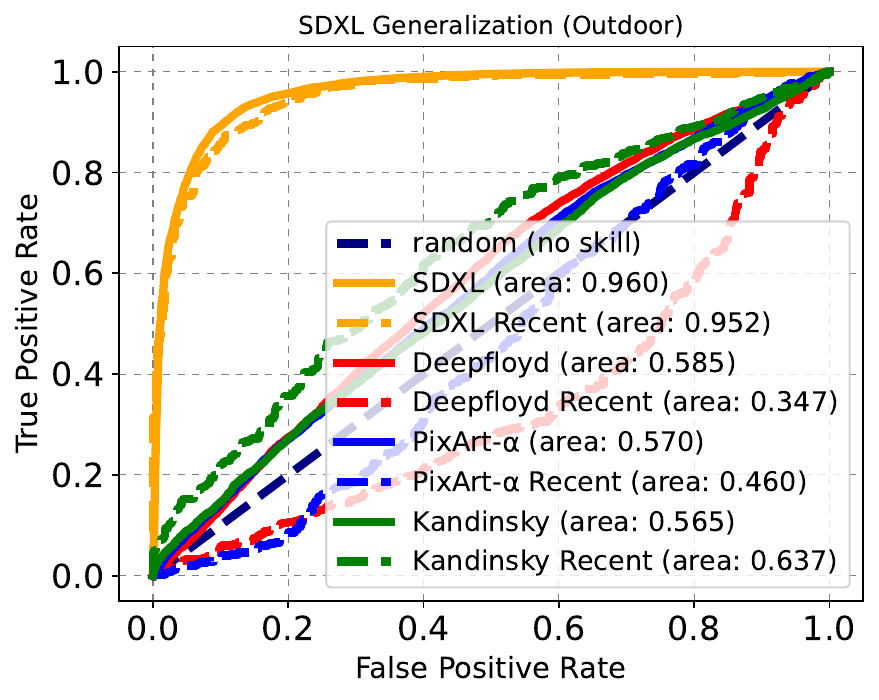}
        \caption{SDXL (Outdoor)}
        \label{fig:DeepFloyd outdoor}
    \end{subfigure}
    \hfill
    \begin{subfigure}[b]{0.24\textwidth}
        \includegraphics[width=\textwidth]{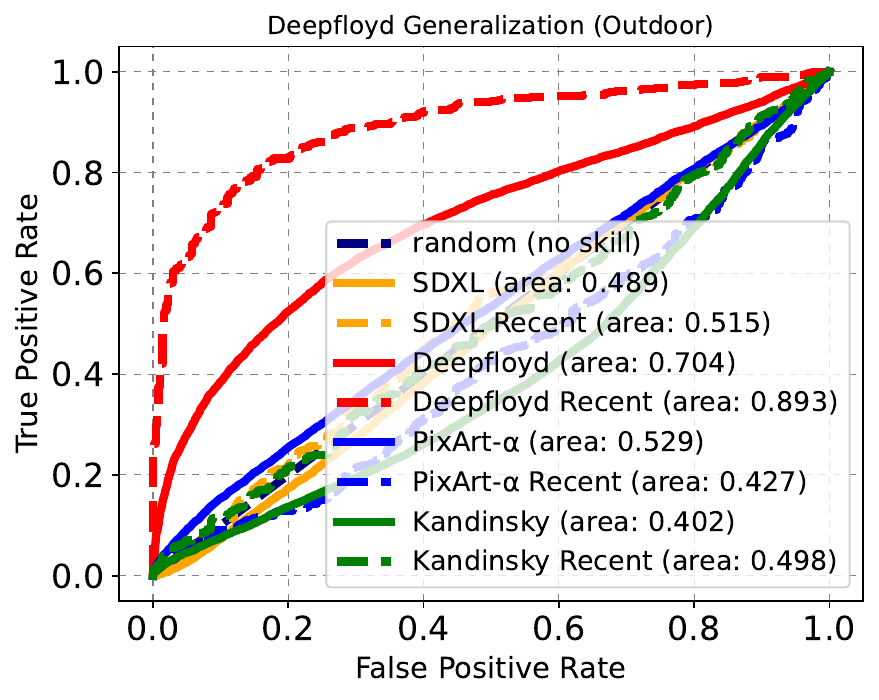}
        \caption{Deepfloyd (Outdoor)}
        \label{fig:firefly-outdoor-recent}
    \end{subfigure}
    \hfill
    \begin{subfigure}[b]{0.24\textwidth}
        \includegraphics[width=\textwidth]{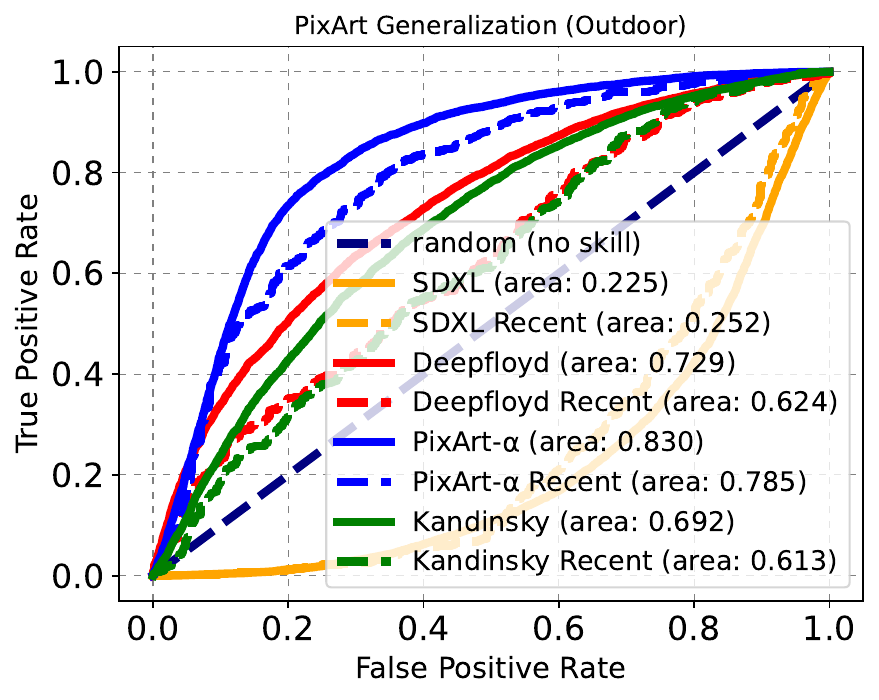}
        \caption{PixArt-$\alpha$ (Outdoor)}
        \label{fig:pixart-outdoor-recent}
    \end{subfigure}
    \hfill
    \begin{subfigure}[b]{0.24\textwidth}
        \includegraphics[width=\textwidth]{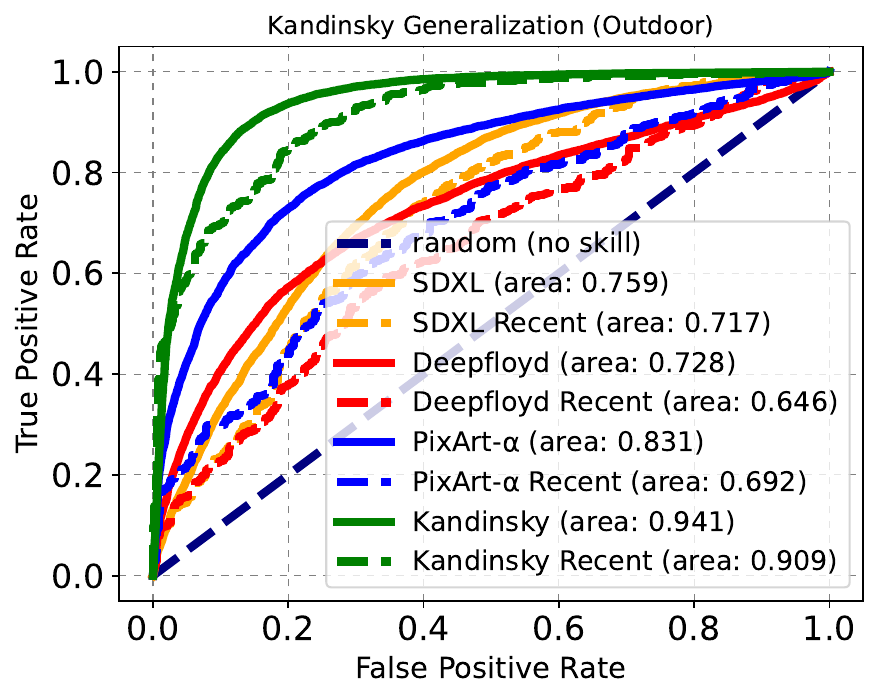}
        \caption{Kandinsky-v3 (Outdoor)}
        \label{fig:kandinsky-outdoor-recent}
    \end{subfigure}
    \vspace{-10pt}
    \caption{
    We trained classifiers on indoor and outdoor scenes using training sets consisting of images from different generators, following the indoor and outdoor splits depicted in Table \ref{tab:data_stats}. The models with the best validation accuracy over 30 epochs were selected. Data augmentation was performed using a protocol similar to that of ~\cite{wang2019cnngenerated}, without blurring and with a JPEG compression probability of 5 percent, to improve generalizability towards both images generated from unseen generators and out-of-distribution real images with recent timestamps.
The classifiers were evaluated on test sets containing 10,000 real images and 10,000 generated images from a target generator, paired caption-wise. Our results show that Kandinsky-v3 demonstrates the strongest generalization performance for both indoor and outdoor scenes. However, despite having the best validation accuracy, this Kandinsky-v3 classifier does not necessarily exhibit the absolute best generalization performance across all generators, possibly due to learning dataset-specific patterns. To address this, we selected a robust prequalifier that maintains comparable accuracy while generalizing more effectively to generators such as DeepFloyd. Also see \cref{fig:choosing_generators}.}
    \label{fig:roc_curves_other_generators_suppl}
\end{figure*}

\begin{figure*}[t!]
\scriptsize
    \centering
    \begin{subfigure}[b]{0.33\textwidth}
        \includegraphics[width=\textwidth]{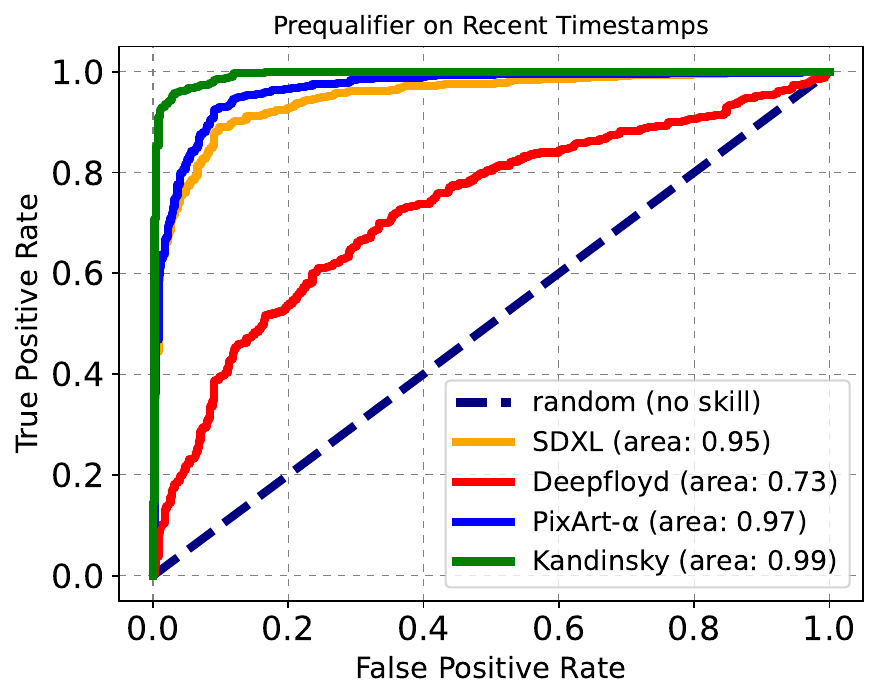}
        \caption{Indoor Prequalifer on Recent Timestamps}
        \label{fig:indoor-preq-recent}
    \end{subfigure}
    \hfill
    \begin{subfigure}[b]{0.33\textwidth}
        \includegraphics[width=\textwidth]{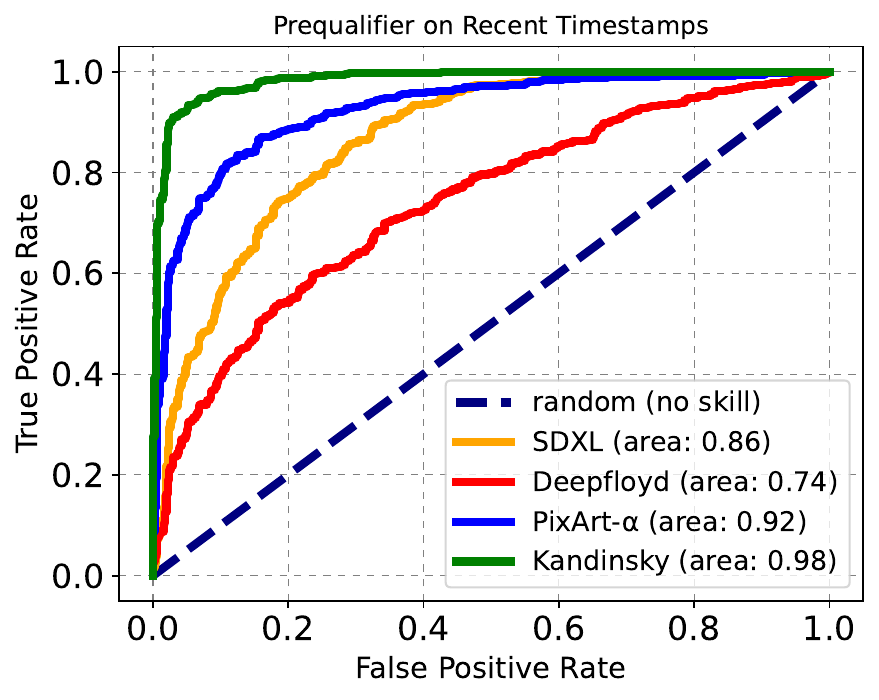}
        \caption{Outdoor Prequalifer on Recent Timestamps}
        \label{fig:outdoor-preq-recent}
    \end{subfigure}
    \hfill
     \begin{subfigure}[b]{0.33\textwidth}
        \includegraphics[width=\textwidth]{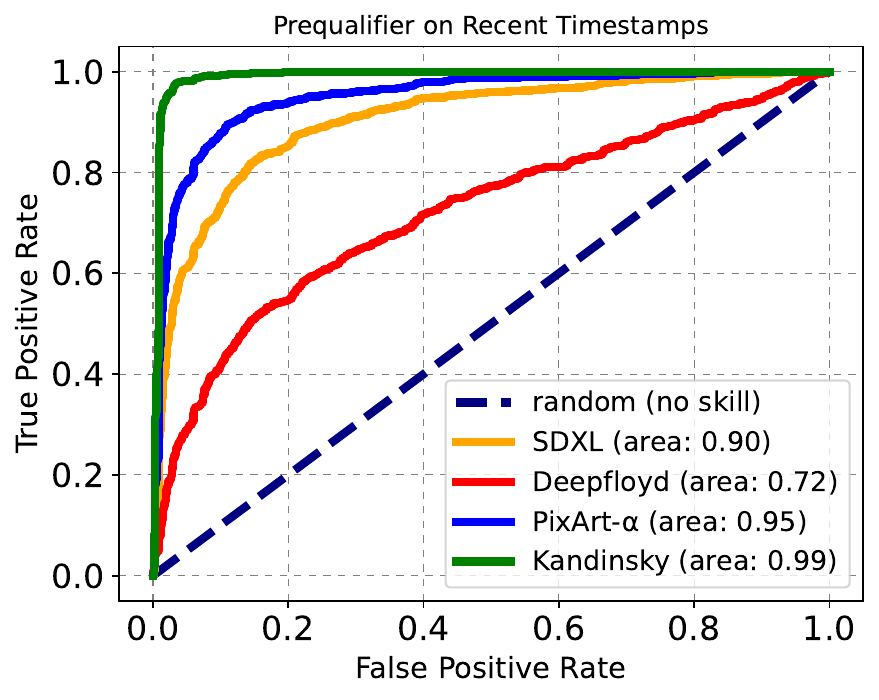}
        \caption{Combined Prequalifer on Recent Timestamps}
        \label{fig:combined-preq-recent}
    \end{subfigure}
    \hfill
    \vspace{-10pt}
    \caption{
    {Based on the transferability experiments in \cref{fig:roc_curves_other_generators_suppl}, we chose to train our prequalifiers on Kandinsky-generated images. Instead of solely focusing on the highest validation accuracy, we selected prequalifiers that performed comparably well while demonstrating the best generalization towards other generators. This approach ensures the robustness of our prequalifiers. The resulting prequalifiers for indoor, outdoor, and combined settings all exhibit strong generalization performance on recent timestamp images from various generators, with particularly significant improved results on DeepFloyd-generated images.}
}
    \label{fig:choosing_generators}
\end{figure*}

\begin{figure*}[t!]
    \centering
    \hfill
    \begin{subfigure}[b]{0.33\textwidth}
        \includegraphics[width=\textwidth]{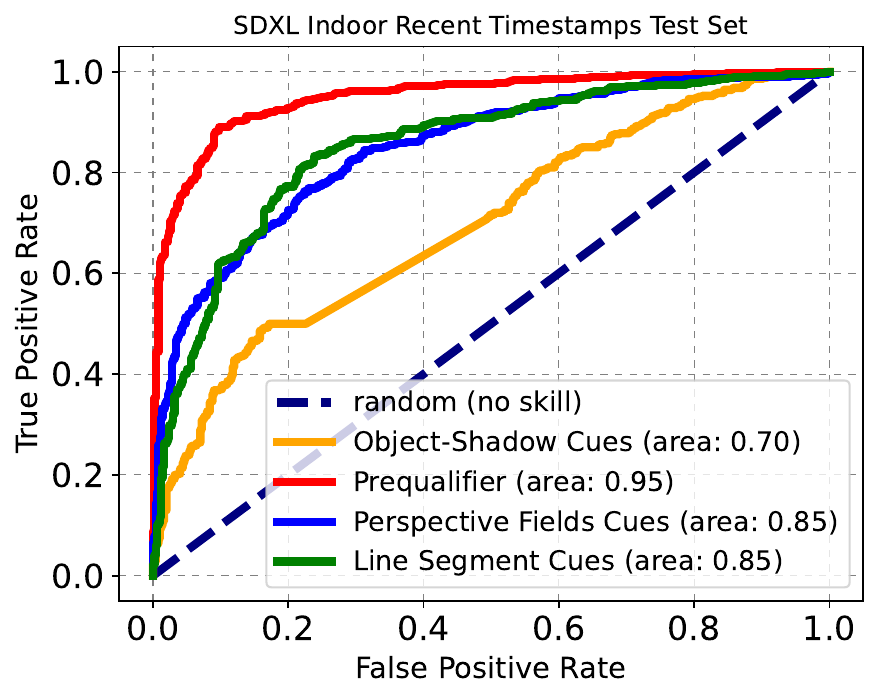}
        \caption{SDXL Recent Timestamp Indoor Test Set}
        \label{fig:sdxl-indoor-recent-test}
    \end{subfigure}
    \hfill
     \begin{subfigure}[b]{0.33\textwidth}
          \includegraphics[width=\textwidth]{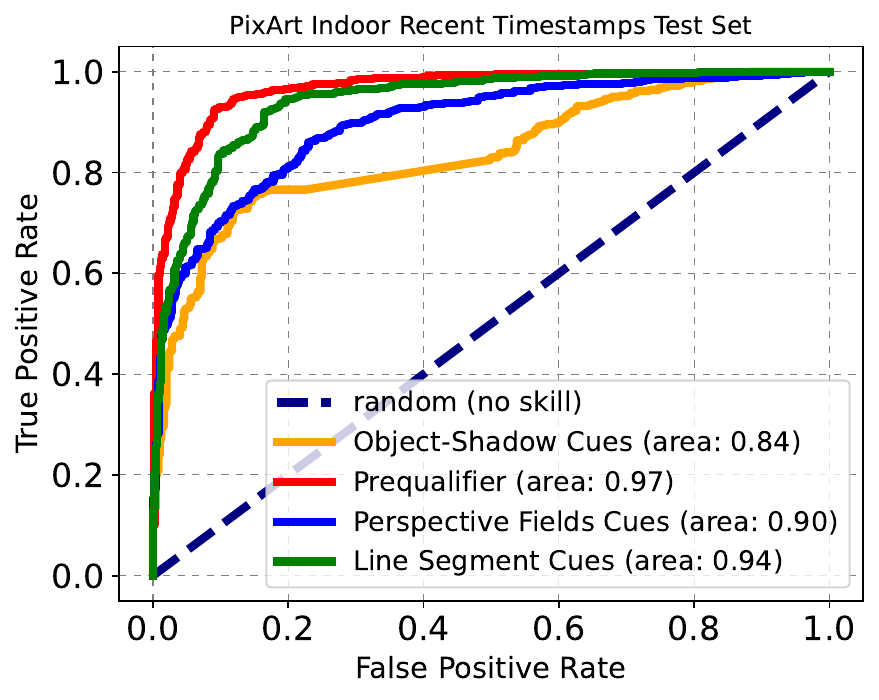}
        \caption{Pixart-$\alpha$ Recent Timestamp Indoor Test Set}
        \label{fig:pixart-indoor-recent-test}
    \end{subfigure}
    \hfill
    \begin{subfigure}[b]{0.33\textwidth}
        \includegraphics[width=\textwidth]{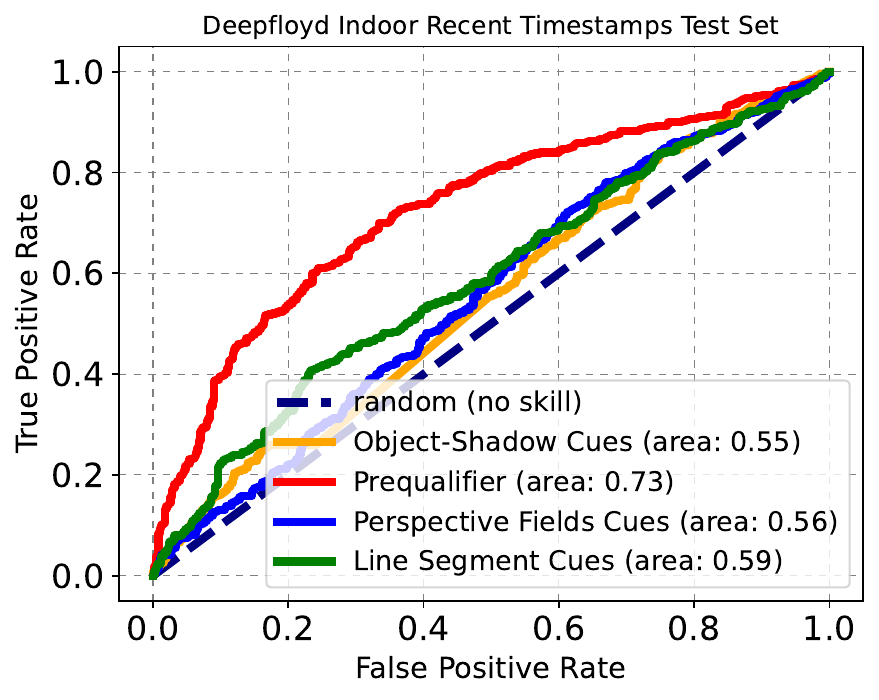}
        \caption{Deepfloyd Recent Timestamp Indoor Test Set}
        \label{fig:deepfloyd-indoor-recent-test}
    \end{subfigure}
    \hfill
    \begin{subfigure}[b]{0.33\textwidth}
        \includegraphics[width=\textwidth]
        {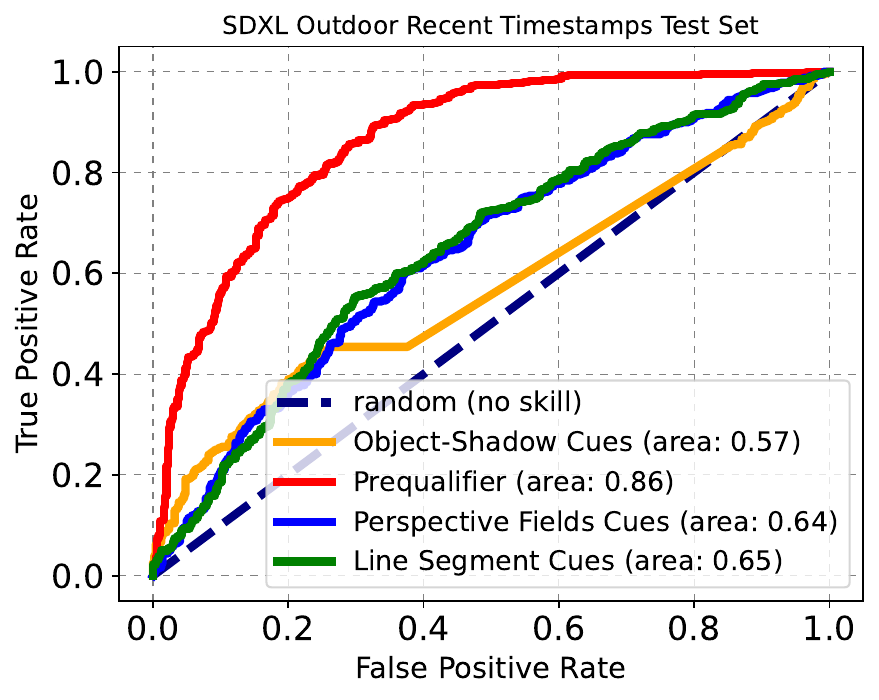}
        \caption{SDXL Recent Timestamp Outdoor Test Set}
        \label{fig:sdxl-outdoor-recent-test}
    \end{subfigure}
    \begin{subfigure}[b]{0.33\textwidth}
        \includegraphics[width=\textwidth]
        {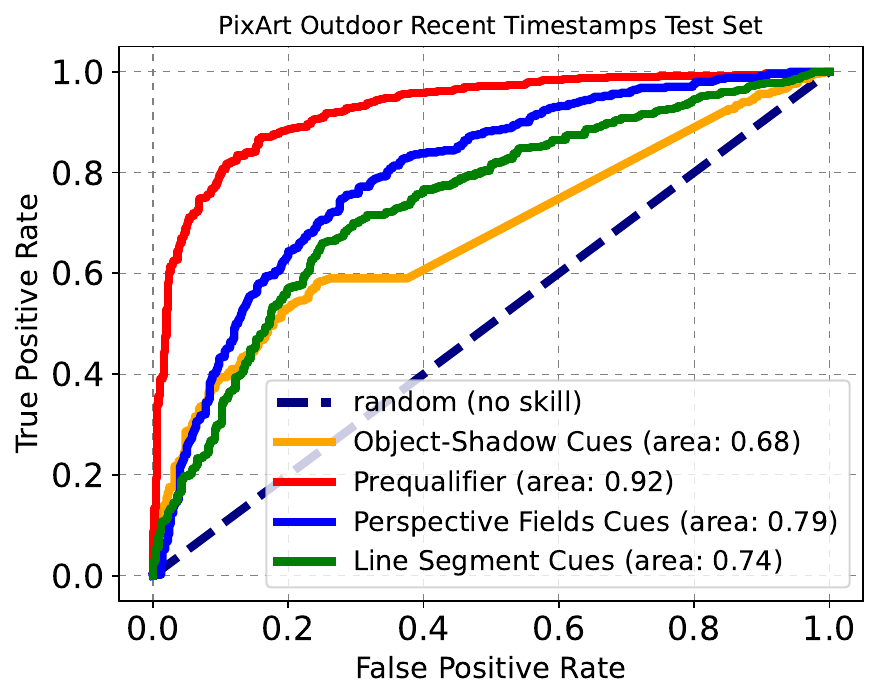}
        \caption{Pixart-$\alpha$ Recent Timestamp Outdoor Test Set}
        \label{fig:pixart-outdoor-recent-test}
    \end{subfigure}
    \begin{subfigure}[b]{0.33\textwidth}
        \includegraphics[width=\textwidth]
        {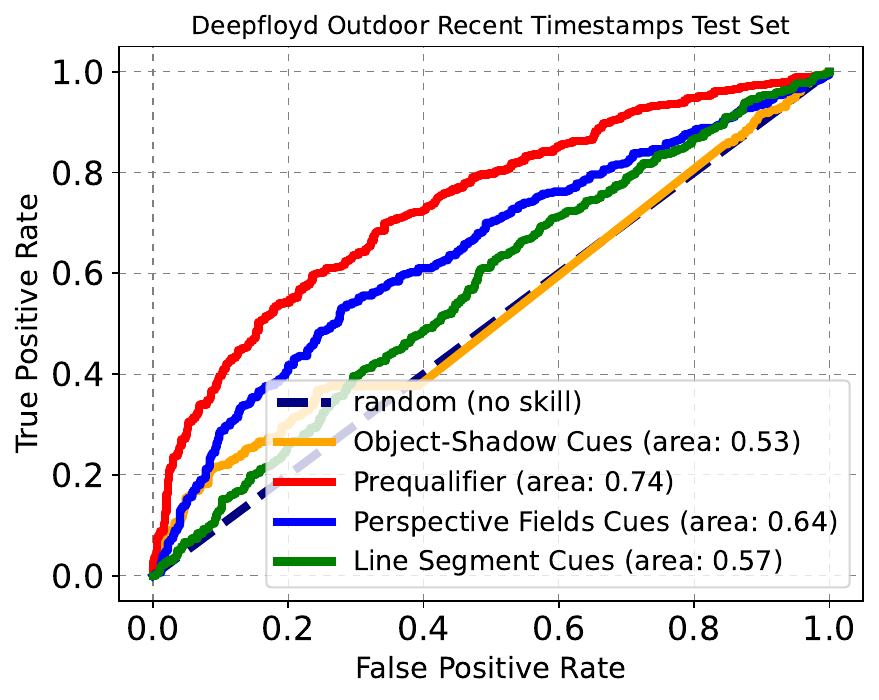}
        \caption{Deepfloyd Recent Timestamp Outdoor Test Set}
        \label{fig:deepfloyd-outdoor-recent-test}
    \end{subfigure}
    \begin{subfigure}[b]{0.33\textwidth}
        \includegraphics[width=\textwidth]
        {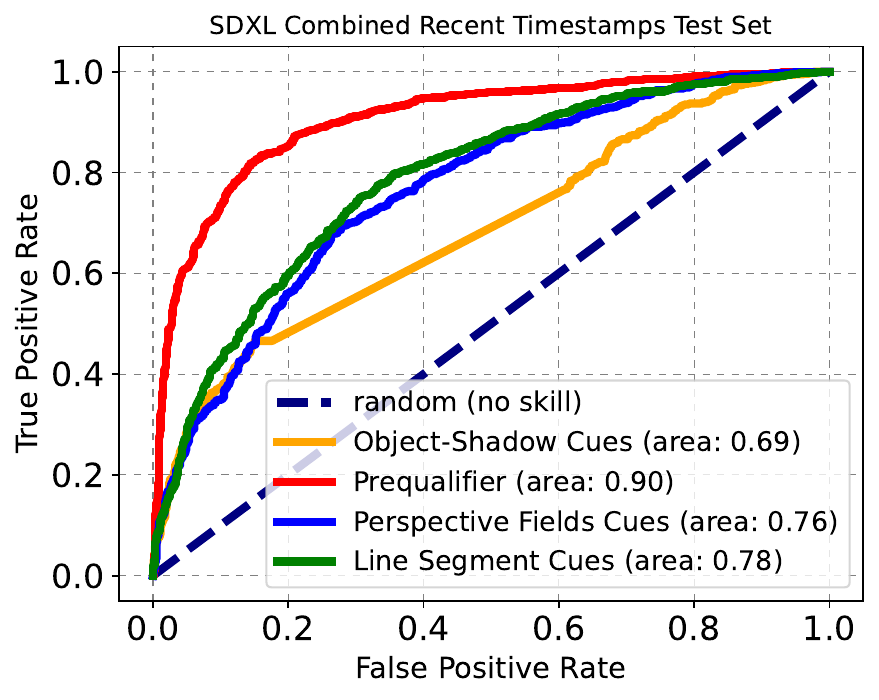}
        \caption{SDXL Recent Timestamp Combined Test Set}
        \label{fig:sdxl-combined-recent-test}
    \end{subfigure}
    \begin{subfigure}[b]{0.33\textwidth}
        \includegraphics[width=\textwidth]
        {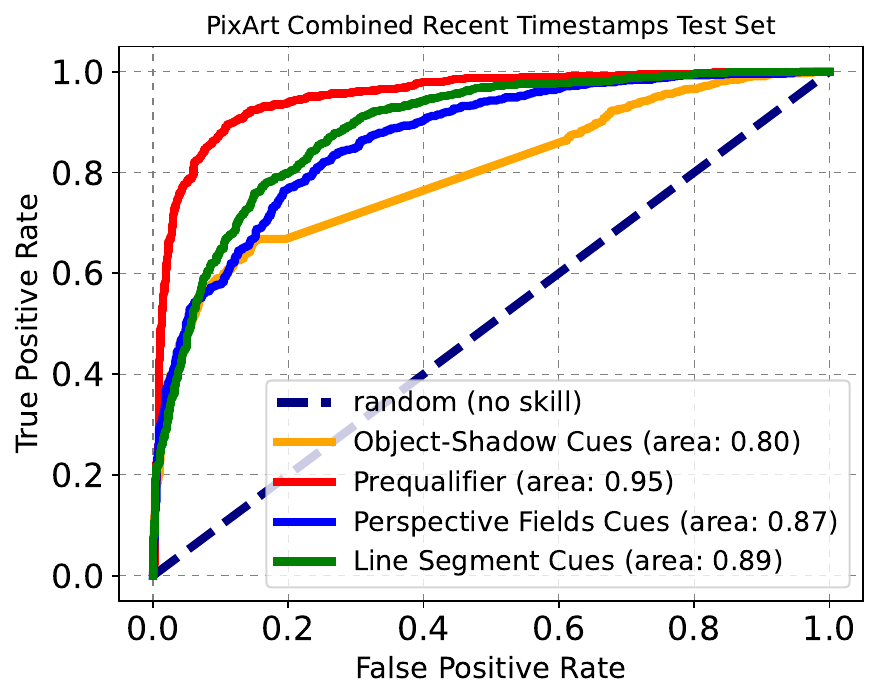}
        \caption{Pixart-$\alpha$ Recent Timestamp Combined Test Set}
        \label{fig:pixart-combined-recent-test}
    \end{subfigure}
    \begin{subfigure}[b]{0.33\textwidth}
        \includegraphics[width=\textwidth]
        {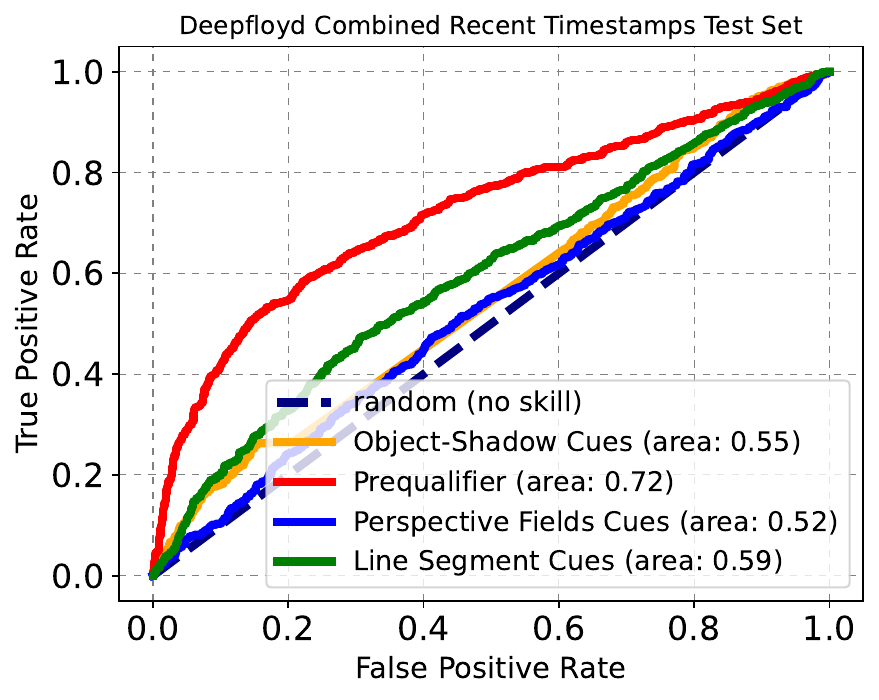}
        \caption{Deepfloyd Recent Timestamp Combined Test Set}
        \label{fig:deepfloyd-combined-recent-test}
    \end{subfigure}
\caption{Our geometric classifiers, trained on derived geometric features from Kandinsky-v3 generated images, demonstrate strong generalization in detecting projective geometry errors within images generated by various unseen generators, using captions from real images with recent timestamps. We evaluate the classifiers on sets involving indoor scenes (top row), outdoor scenes (middle row), and a combination of indoor and outdoor scenes (last row). For indoor scenes, our perspective fields and line segment classifiers maintain strong AUCs greater than 0.84 for both SDXL and Pixart-$\alpha$, while our object-shadow classifiers also exhibit comparable performance with AUCs of 0.70 and 0.84, respectively. Although the outdoor and combined settings pose a greater challenge compared to indoor scenes alone, our models, despite relying solely on geometric cues, remain robust towards SDXL and Pixart-$\alpha$. However, generalization towards DeepFloyd proves to be quite challenging overall.}
\end{figure*}

\section{Additional Analysis}

In Table \ref{tab:disagreement}, we provide quantitative analysis that our Line Segment cues and Perspective Field cues are correlated and look at similar geometric cues while Object-Shadow cues look for different geometric cues to identify if an image is generated or real.

We also provide statistical distributions of geometry cues leveraged for detecting projective geometry distortion. These include Object-Shadow pairs, Perspective Fields, and line segment distributions obtained from DeepLSD. The distributions are in Figures ~\ref{fig:object_shadow_distribution}, ~\ref{fig:perspective_field_distribution_indoor}, ~\ref{fig:perspective_field_distribution_outdoor}, ~\ref{fig:line_segment_distribution_indoor}, and ~\ref{fig:line_segment_distribution_outdoor}.

\begin{figure*}[t!]
    \centering
    \hfill
    \begin{subfigure}[b]{0.33\textwidth}
        \includegraphics[width=\textwidth]{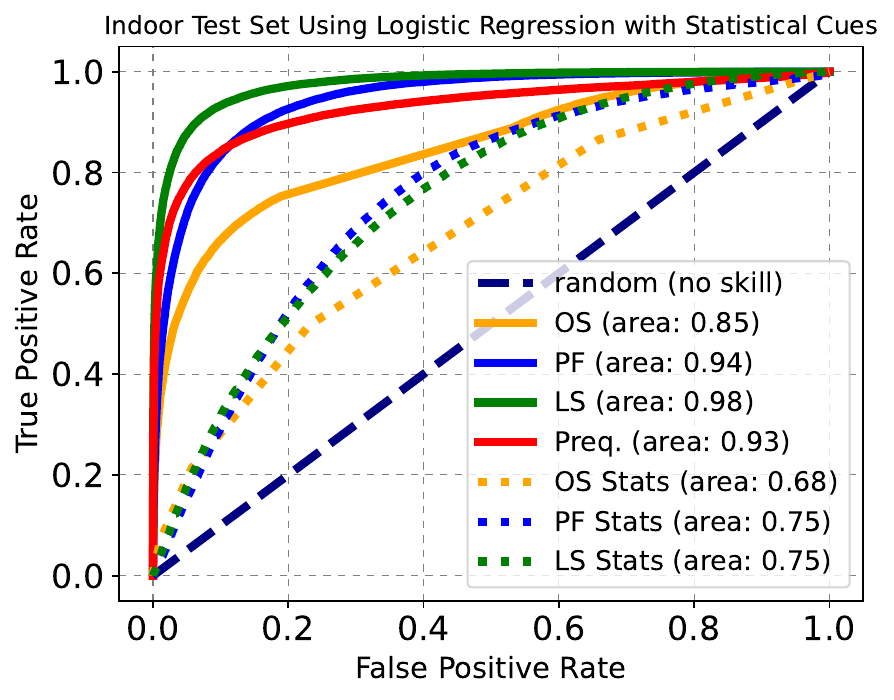}
    \end{subfigure}
    \hfill
     \begin{subfigure}[b]{0.33\textwidth}
          \includegraphics[width=\textwidth]{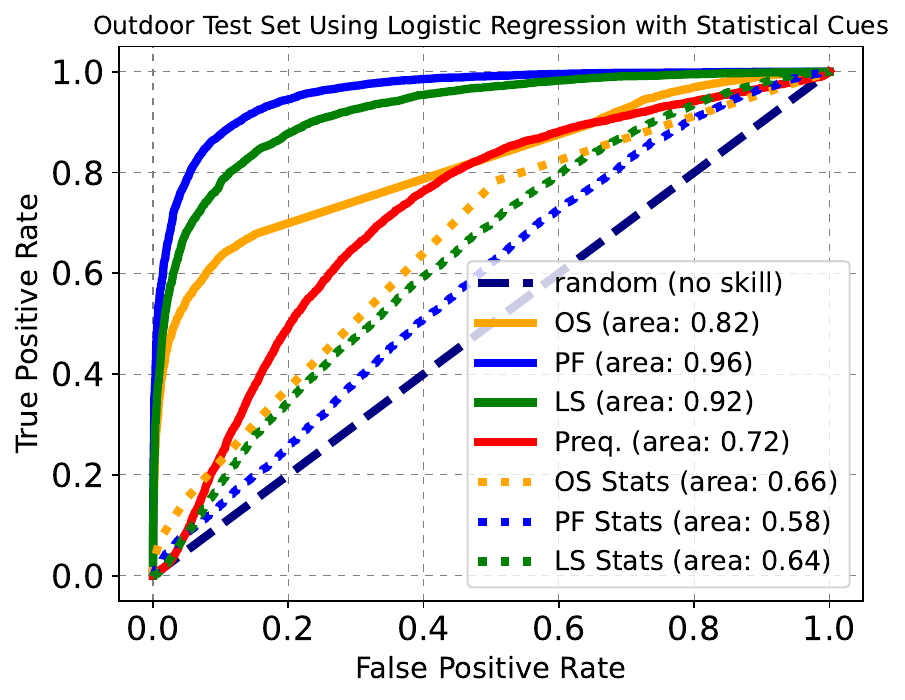}
    \end{subfigure}
    \hfill
    \begin{subfigure}[b]{0.33\textwidth}
        \includegraphics[width=\textwidth]{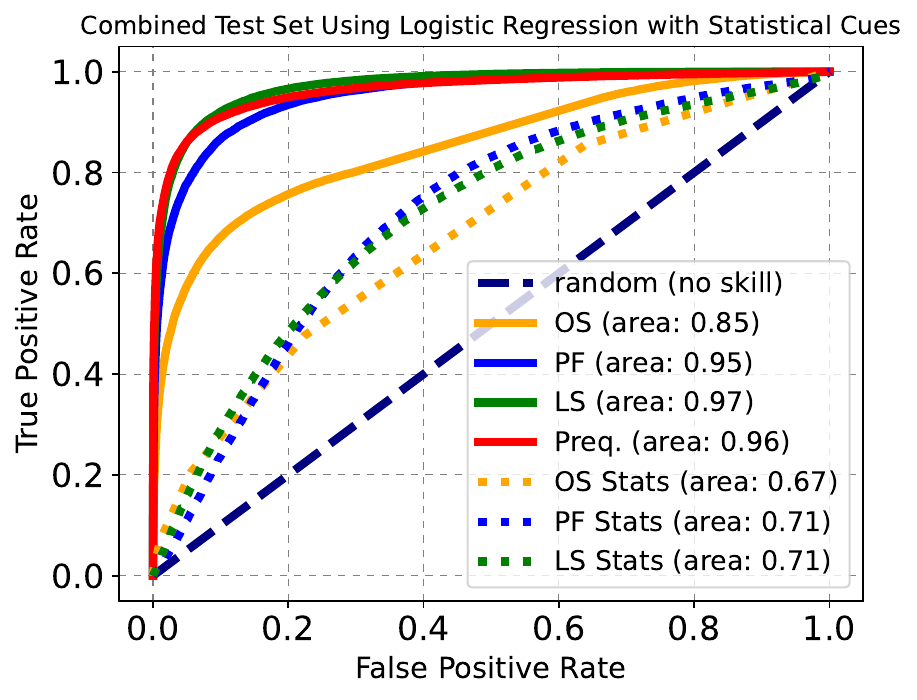}
    \end{subfigure}
    \vspace{-20pt}
    \caption{ROC analysis comparing our classifiers against basic statistical cues on our full test set. We compare the performance of our sophisticated classifiers -- Object-Shadow (OS), Perspective Fields (PF) ResNet, and Line Segments (LS) PointNet classifiers -- with basic statistical measures applied via logistic regression (LR) on indoor, outdoor, and combined test sets shown in dotted lines. While basic statistical cues like the count and mean lengths of line segments, the number of object shadows, and gravity changes per pixel indicate better-than-chance performance (AUCs ranging from 0.58 to 0.75), they are eclipsed by the more robust classifiers we developed and also the ResNet prequalifier. Our classifiers excel in identifying generated images with incorrect projective geometry by focusing on incorrect regions, not just statistical cues, as demonstrated by GradCAM visualizations. 
}
    \label{fig:LR_statistical_cues}
\end{figure*}

\begin{figure*}[t!]
    \centering
    \hfill
    \begin{subfigure}[b]{0.33\textwidth}
        \includegraphics[width=\textwidth]{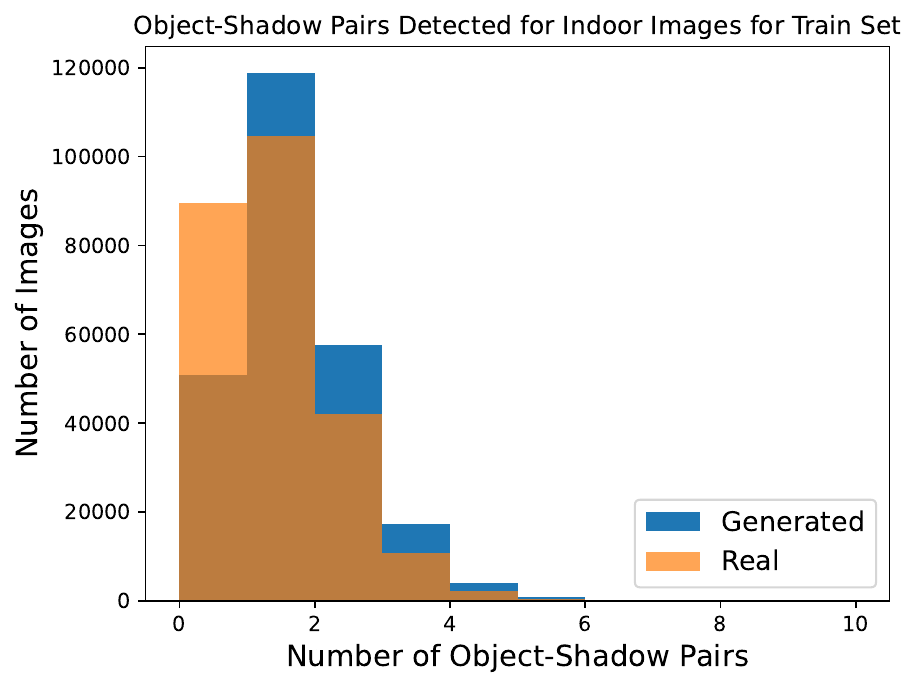}
    \end{subfigure}
    \hfill
     \begin{subfigure}[b]{0.33\textwidth}
          \includegraphics[width=\textwidth]{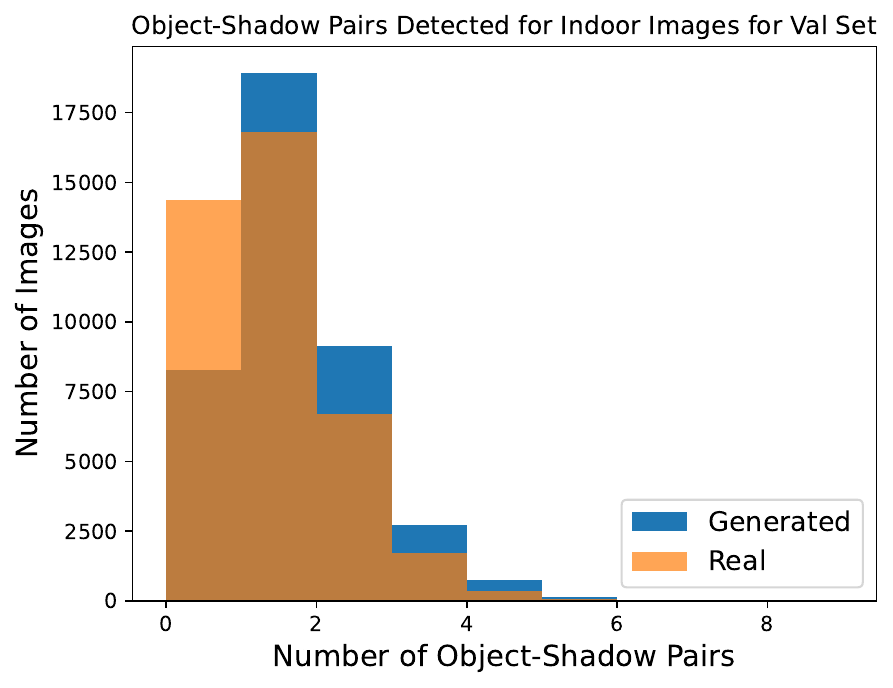}
    \end{subfigure}
    \hfill
    \begin{subfigure}[b]{0.33\textwidth}
        \includegraphics[width=\textwidth]{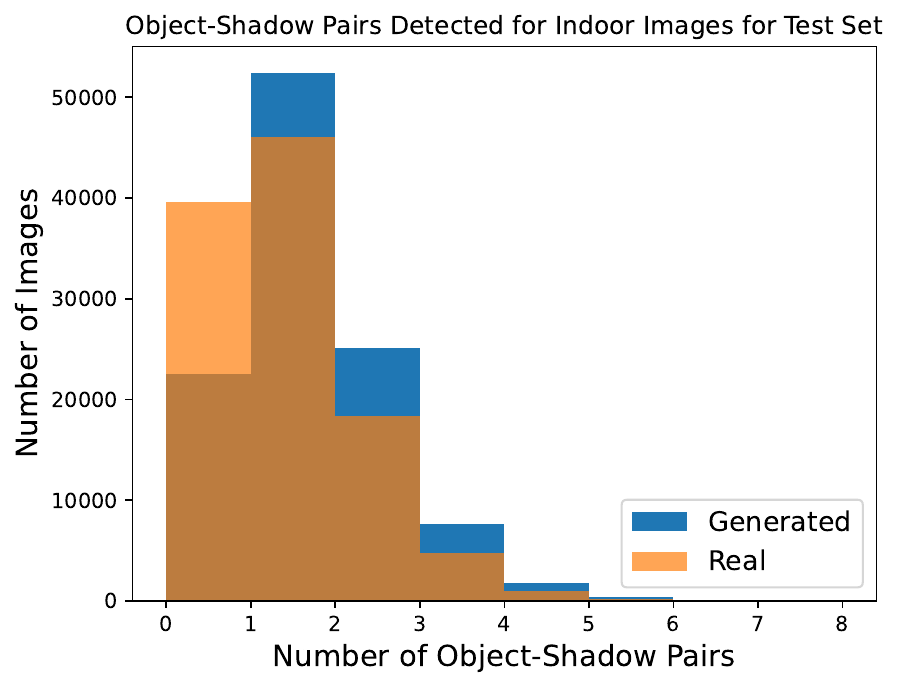}
    \end{subfigure}
    \hfill
    \begin{subfigure}[b]{0.33\textwidth}
        \includegraphics[width=\textwidth]{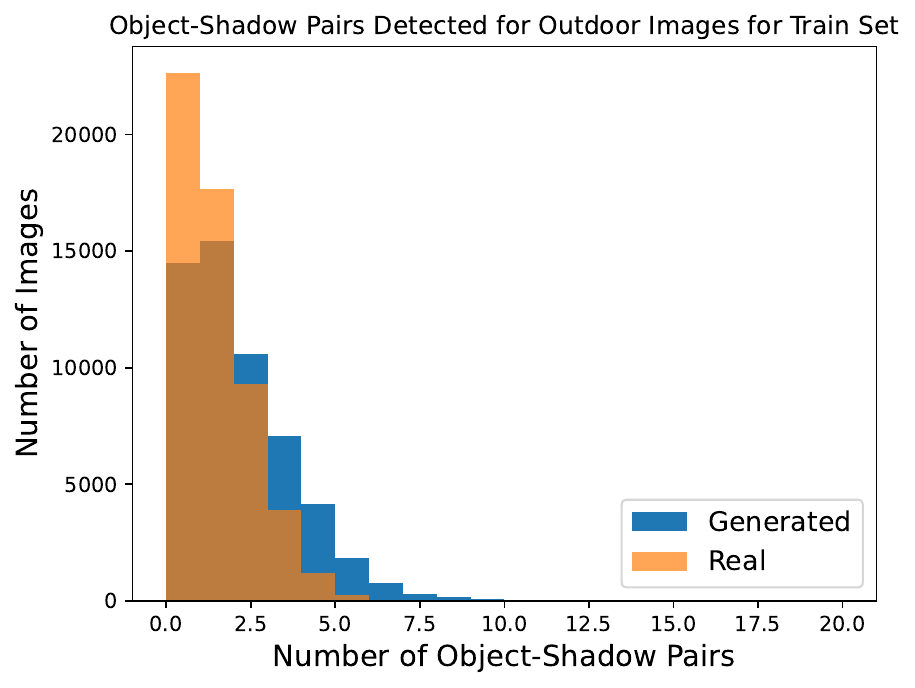}
    \end{subfigure}
    \hfill
    \begin{subfigure}[b]{0.33\textwidth}
        \includegraphics[width=\textwidth]{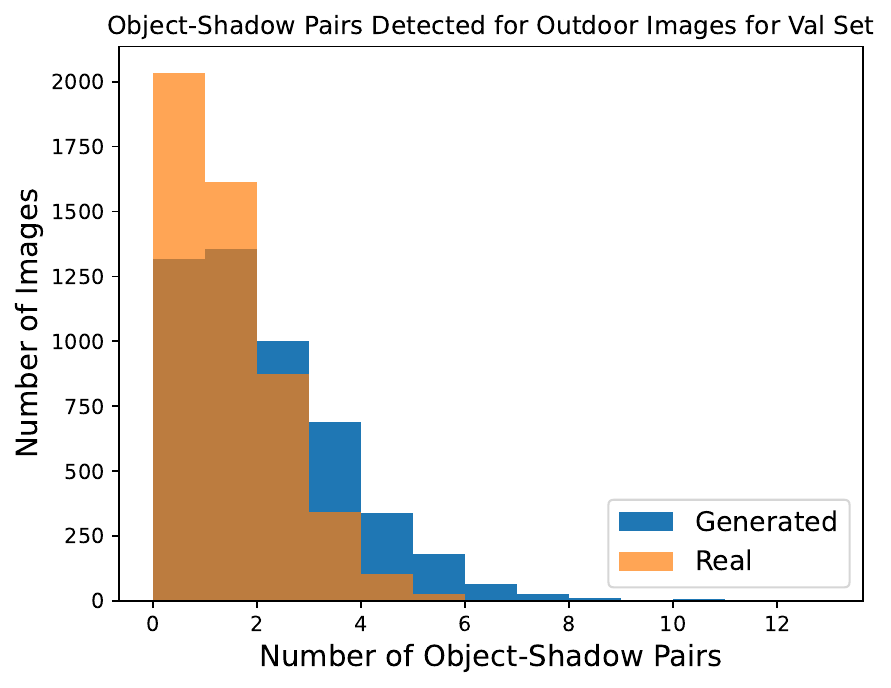}
    \end{subfigure}
    \hfill
    \begin{subfigure}[b]{0.33\textwidth}
        \includegraphics[width=\textwidth]{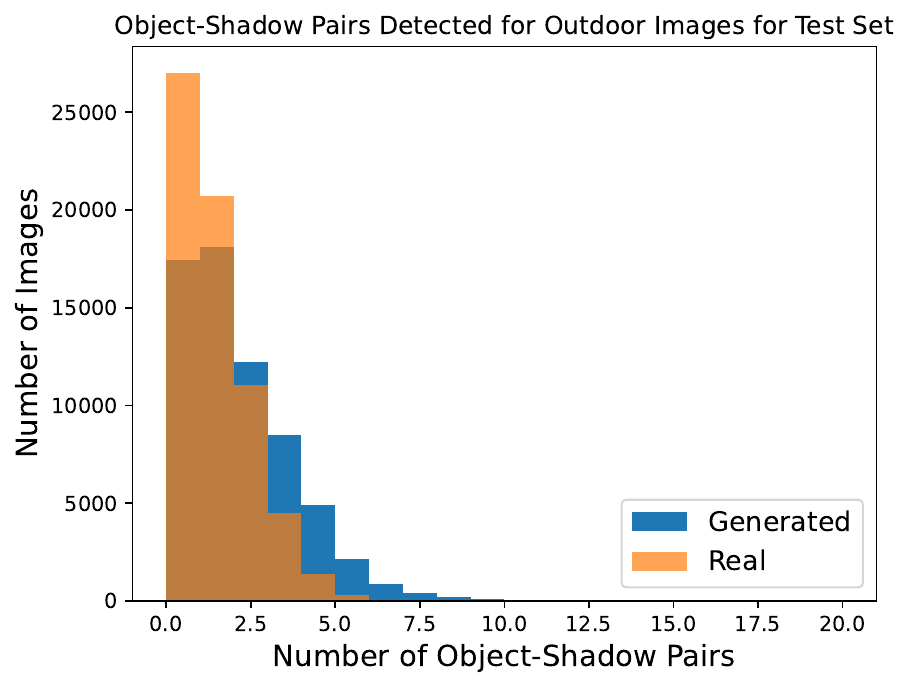}
    \end{subfigure}
    \vspace{-20pt}
    \caption{A statistical distribution analysis of a number of object-shadow pairs for both indoor and outdoor datasets.  A classifier could exploit some of the statistical biases to distinguish between generated and real images. However, we found that our derived geometry cues perform much better than a classifier trained to look at such statistical signals, as shown in Figure~\ref{fig:LR_statistical_cues}. Furthermore, the GradCam analysis indicates that these derived object-shadow cues correctly identify erroneous regions.}
    \label{fig:object_shadow_distribution}
\end{figure*}

\begin{figure*}[t!]
    \centering
    \begin{subfigure}[b]{0.3\textwidth}
        \includegraphics[width=\textwidth]{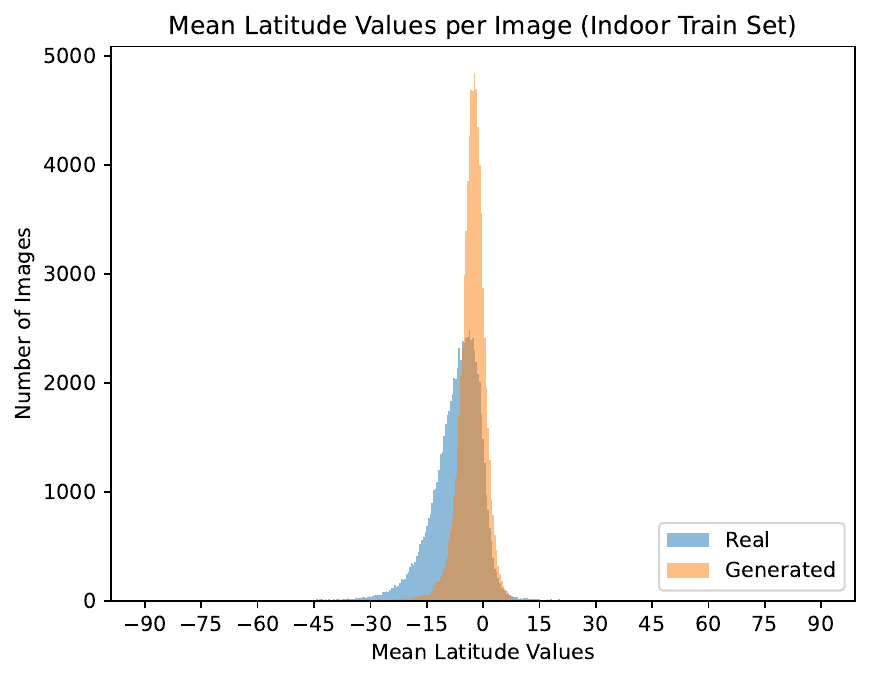}
    \end{subfigure}
    \hfill
     \begin{subfigure}[b]{0.3\textwidth}
          \includegraphics[width=\textwidth]{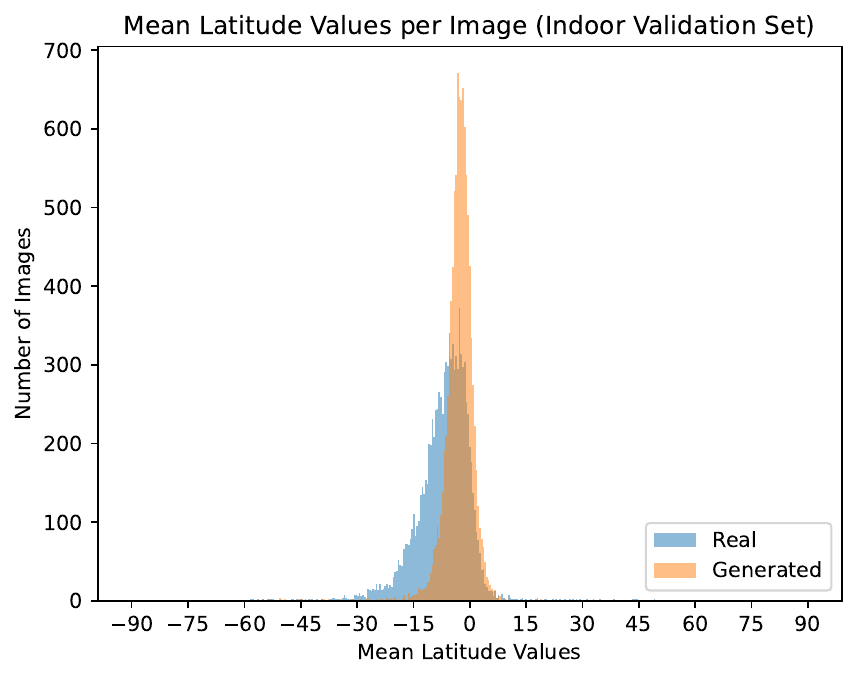}
    \end{subfigure}
    \hfill
    \begin{subfigure}[b]{0.3\textwidth}
        \includegraphics[width=\textwidth]{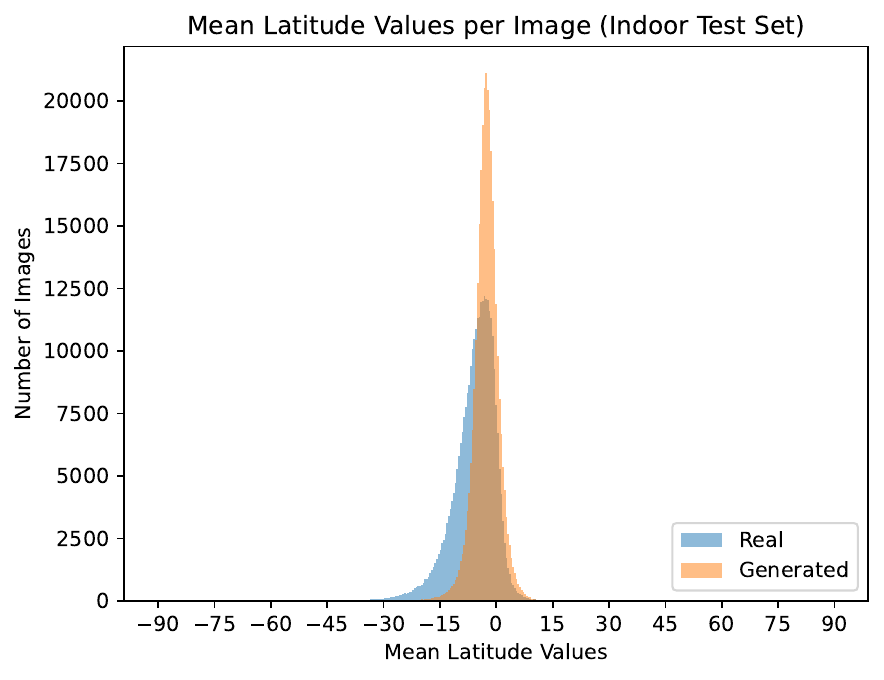}
    \end{subfigure}
    \hfill
    \begin{subfigure}[b]{0.3\textwidth}
        \includegraphics[width=\textwidth]{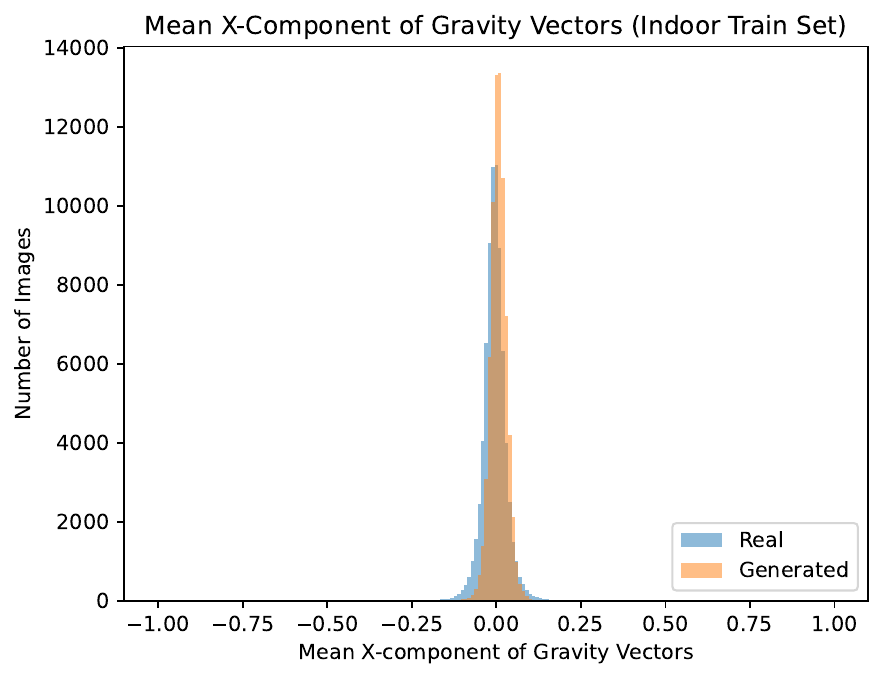}
    \end{subfigure}
    \hfill
     \begin{subfigure}[b]{0.3\textwidth}
          \includegraphics[width=\textwidth]{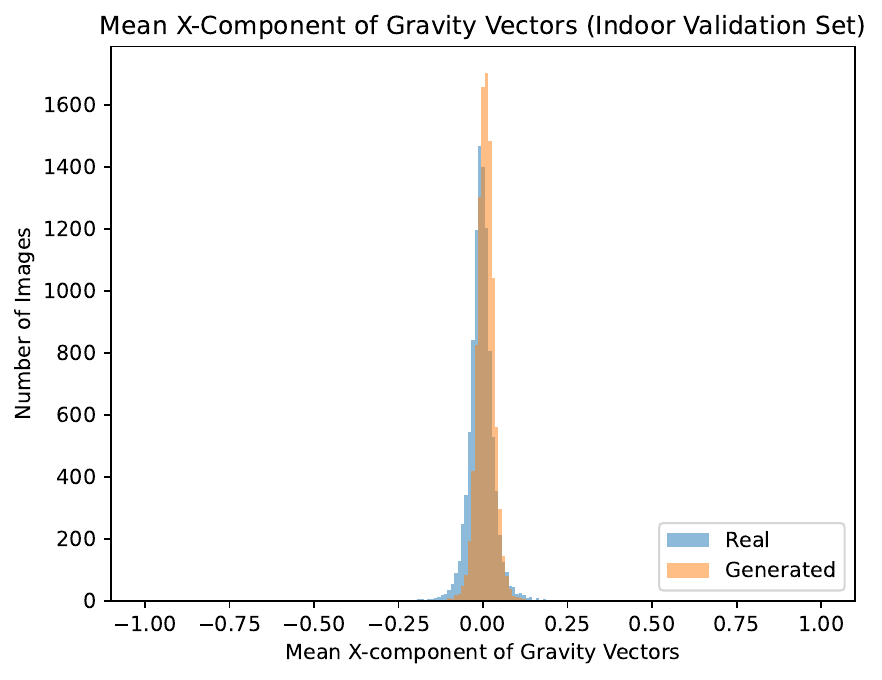}
    \end{subfigure}
    \hfill
    \begin{subfigure}[b]{0.3\textwidth}
        \includegraphics[width=\textwidth]{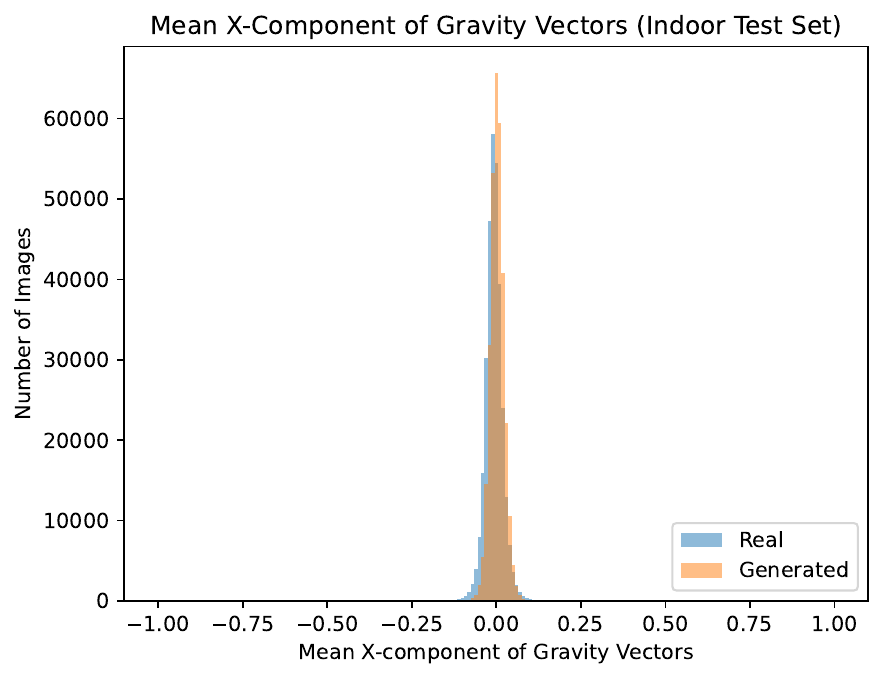}
    \end{subfigure}
    \hfill
    \begin{subfigure}[b]{0.3\textwidth}
        \includegraphics[width=\textwidth]{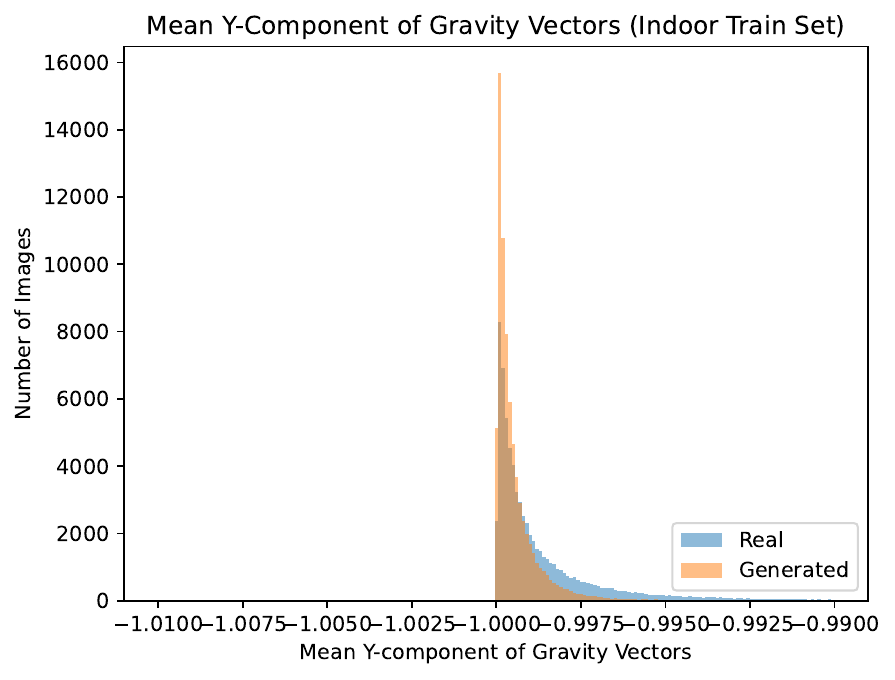}
    \end{subfigure}
    \hfill
     \begin{subfigure}[b]{0.3\textwidth}
          \includegraphics[width=\textwidth]{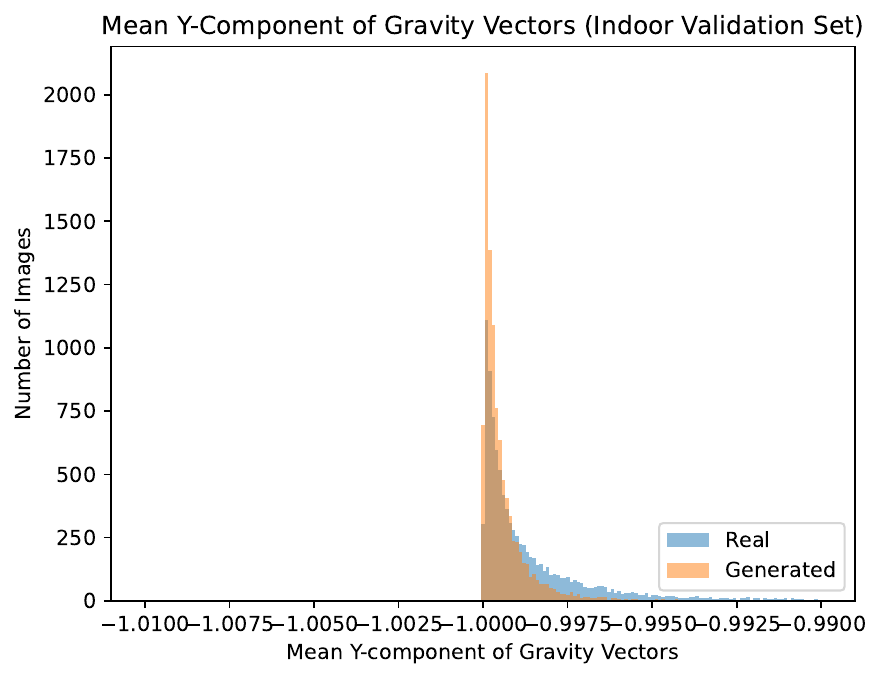}
    \end{subfigure}
    \hfill
    \begin{subfigure}[b]{0.3\textwidth}
        \includegraphics[width=\textwidth]{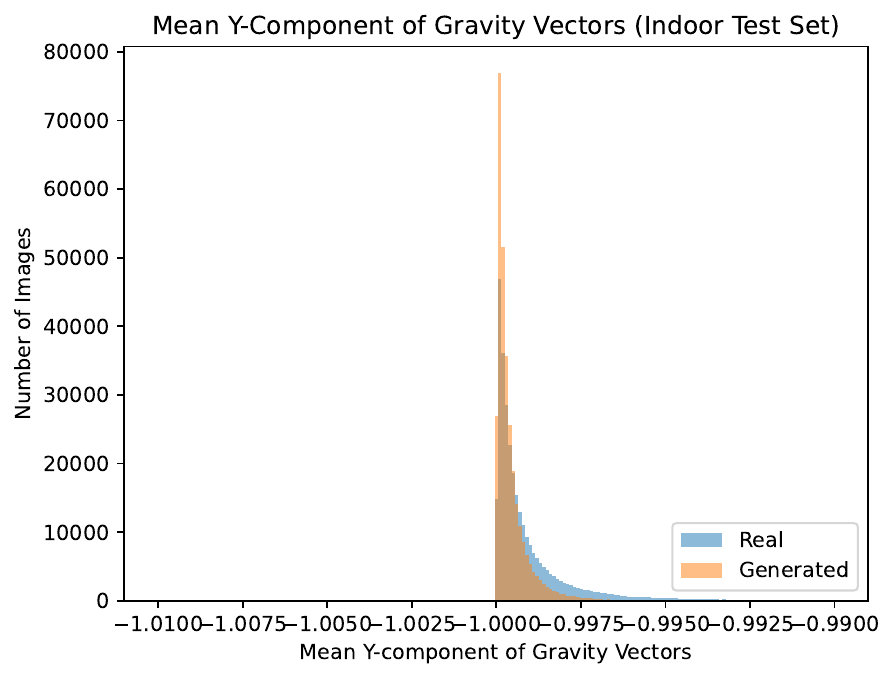}
    \end{subfigure}
    \caption{This set of histogram plots displays the statistical distribution of perspective field metrics in indoor scenes, comparing the training, validation, and test sets. The top row histograms reveal a significant difference in the distribution of latitude angles between real and generated images. The middle and bottom row plots illustrate the mean X and Y components of gravity vectors in the images, showing a clear separation between the real and generated images. These metrics indicate minor spatial inconsistencies between the real and generated images. Although these basic statistical differences provide some discriminative power, they are less effective than our ResNet classifier trained on Perspective Fields, which efficiently detects and focuses on critical geometric inconsistencies. This is validated by our comprehensive ROC analysis in Figure~\ref{fig:LR_statistical_cues}.}
    \label{fig:perspective_field_distribution_indoor}
\end{figure*}

\begin{figure*}[t!]
    \centering
    \begin{subfigure}[b]{0.33\textwidth}
        \includegraphics[width=\textwidth]{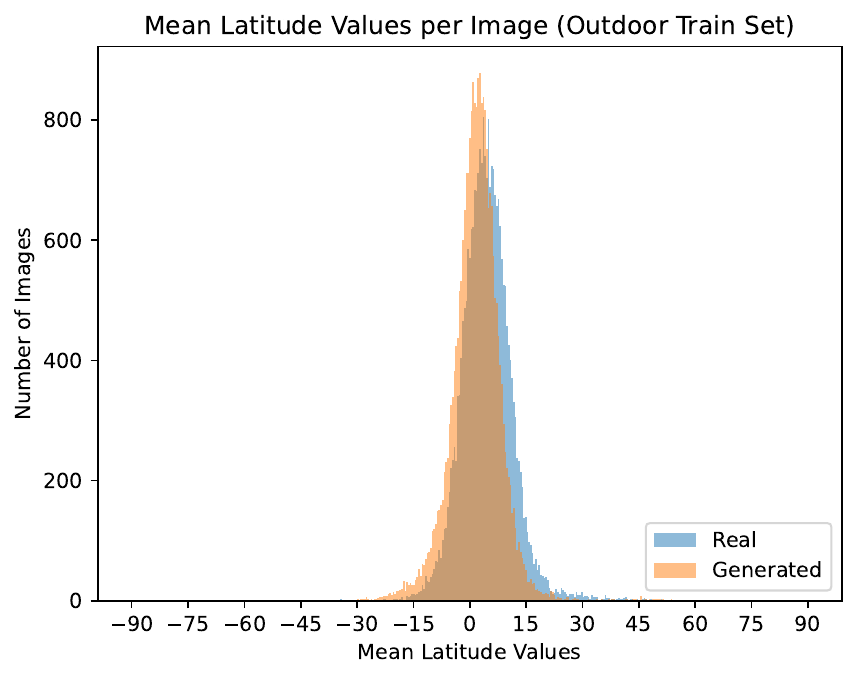}
    \end{subfigure}
    \hfill
     \begin{subfigure}[b]{0.33\textwidth}
          \includegraphics[width=\textwidth]{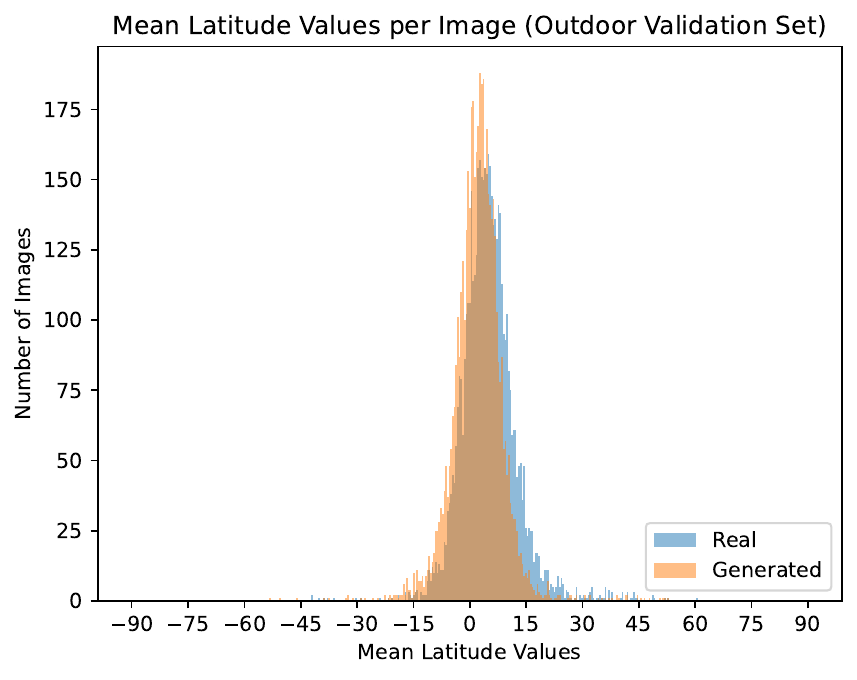}
    \end{subfigure}
    \hfill
    \begin{subfigure}[b]{0.33\textwidth}
        \includegraphics[width=\textwidth]{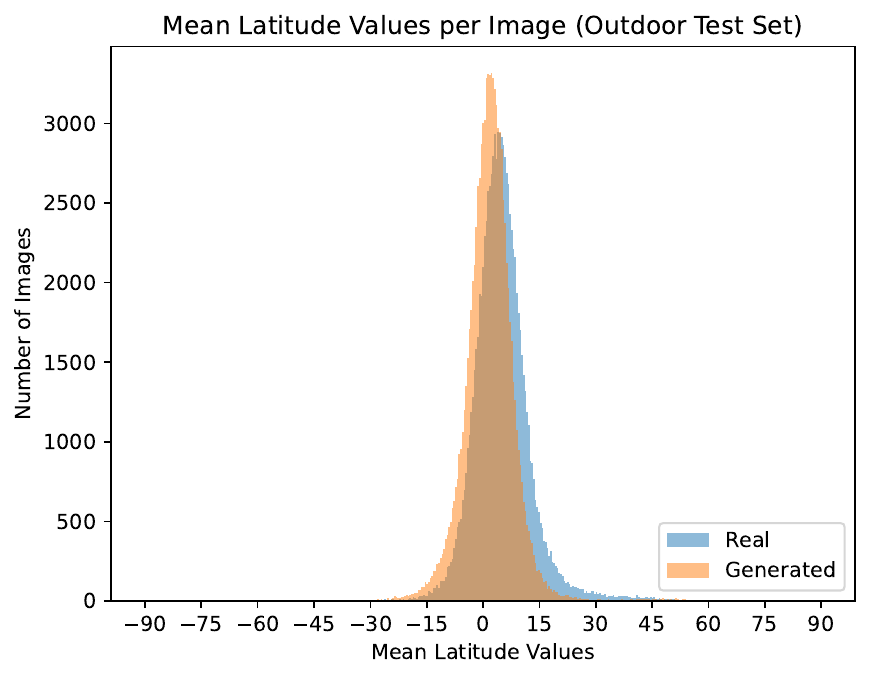}
    \end{subfigure}
    \hfill
    \begin{subfigure}[b]{0.33\textwidth}
        \includegraphics[width=\textwidth]{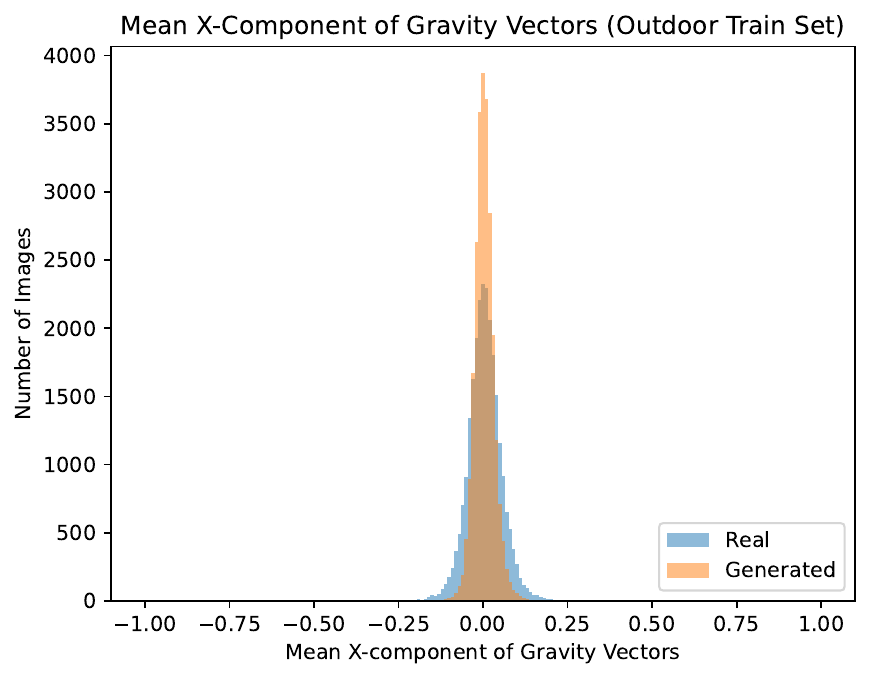}
    \end{subfigure}
    \hfill
     \begin{subfigure}[b]{0.33\textwidth}
          \includegraphics[width=\textwidth]{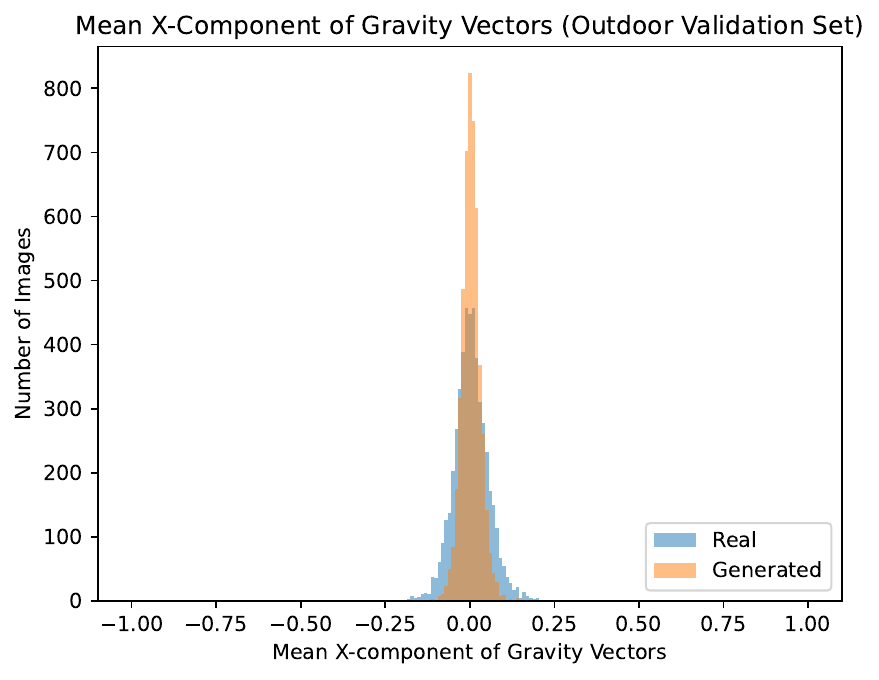}
    \end{subfigure}
    \hfill
    \begin{subfigure}[b]{0.33\textwidth}
        \includegraphics[width=\textwidth]{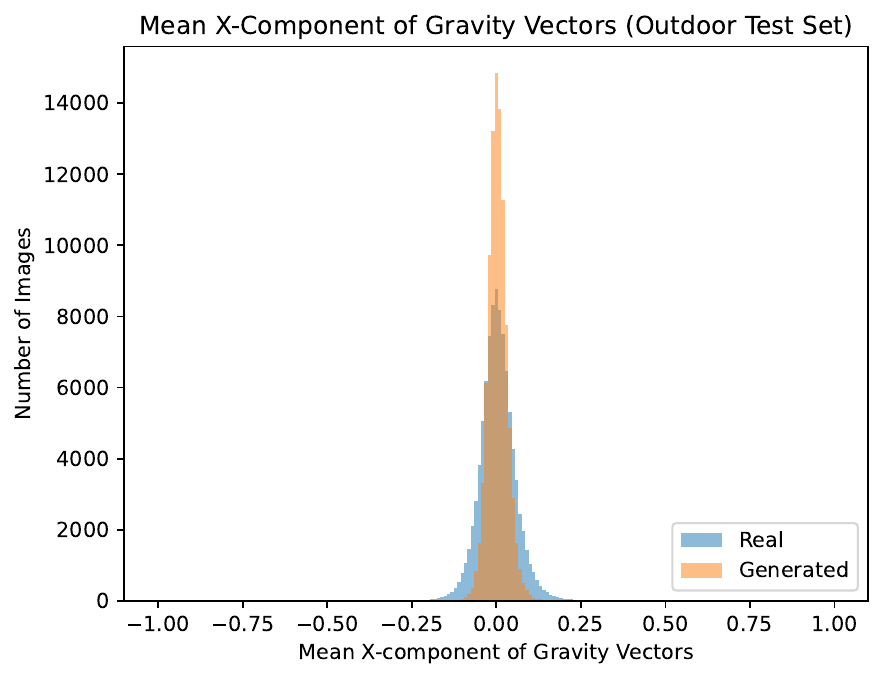}
    \end{subfigure}
    \hfill
    \begin{subfigure}[b]{0.33\textwidth}
        \includegraphics[width=\textwidth]{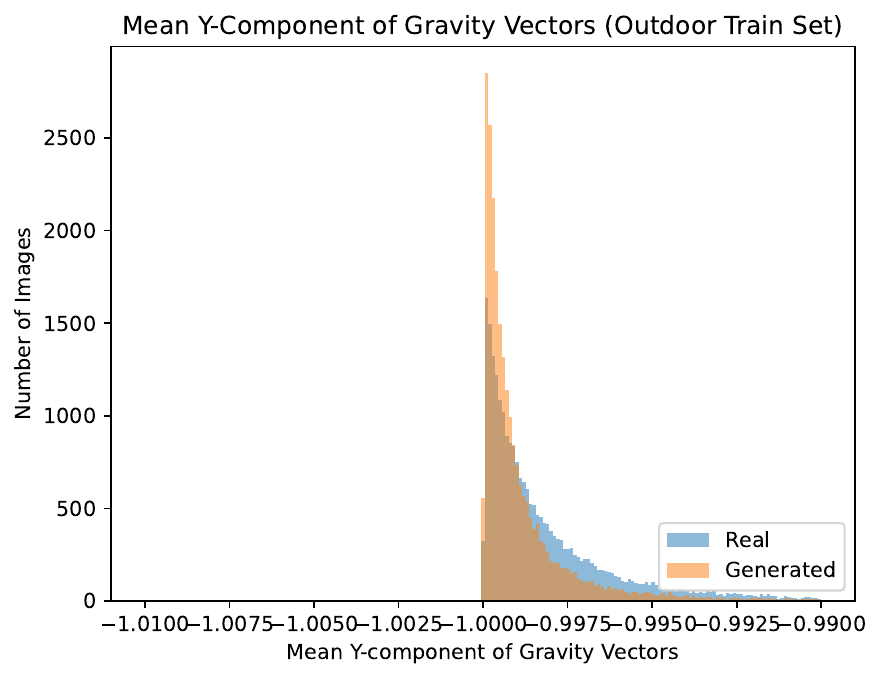}
    \end{subfigure}
    \hfill
     \begin{subfigure}[b]{0.33\textwidth}
          \includegraphics[width=\textwidth]{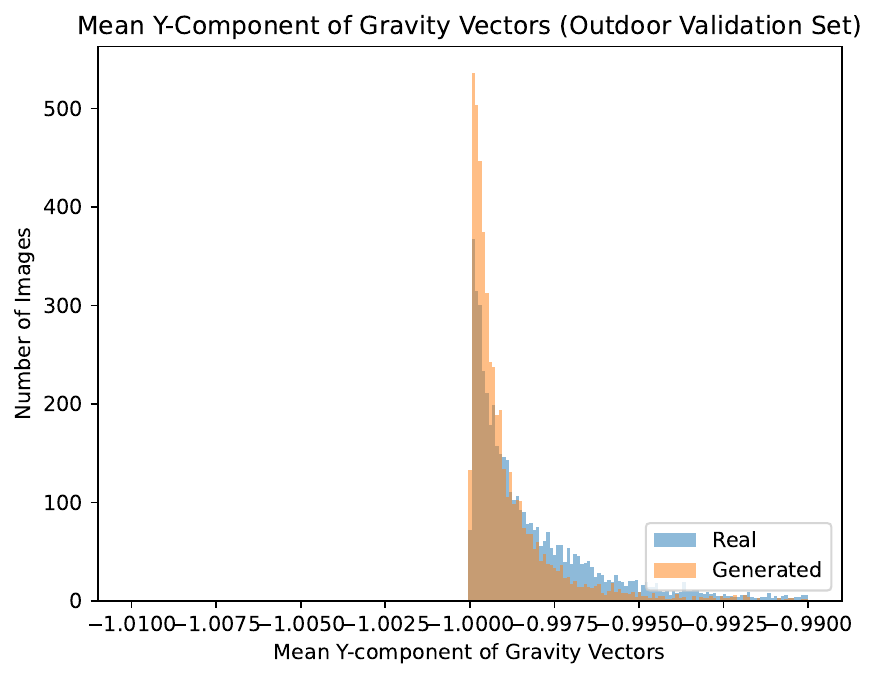}
    \end{subfigure}
    \hfill
    \begin{subfigure}[b]{0.33\textwidth}
        \includegraphics[width=\textwidth]{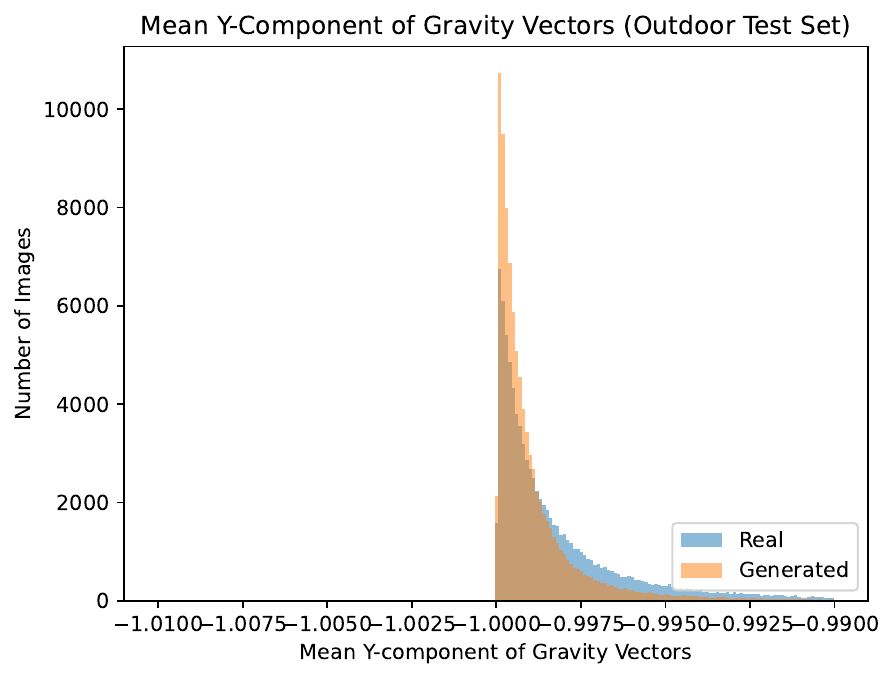}
    \end{subfigure}
    \caption{This set of histogram plots displays the statistical distribution of perspective field metrics in outdoor scenes, comparing the training, validation, and test sets. The top row histograms reveal a significant difference in the distribution of latitude angles between real and generated images. The middle and bottom row plots illustrate the mean X and Y components of gravity vectors in the images, showing a clear separation between the real and generated images. These metrics indicate minor spatial inconsistencies between the real and generated images. Although these basic statistical differences provide some discriminative power, they are less effective than our ResNet classifier trained on Perspective Fields, which efficiently detects and focuses on critical geometric inconsistencies. This is validated by our comprehensive ROC analysis in Figure~\ref{fig:LR_statistical_cues}.}
    \label{fig:perspective_field_distribution_outdoor}
\end{figure*}

\begin{figure*}[t!]
    \centering
    \begin{subfigure}[b]{0.33\textwidth}
        \includegraphics[width=\textwidth]{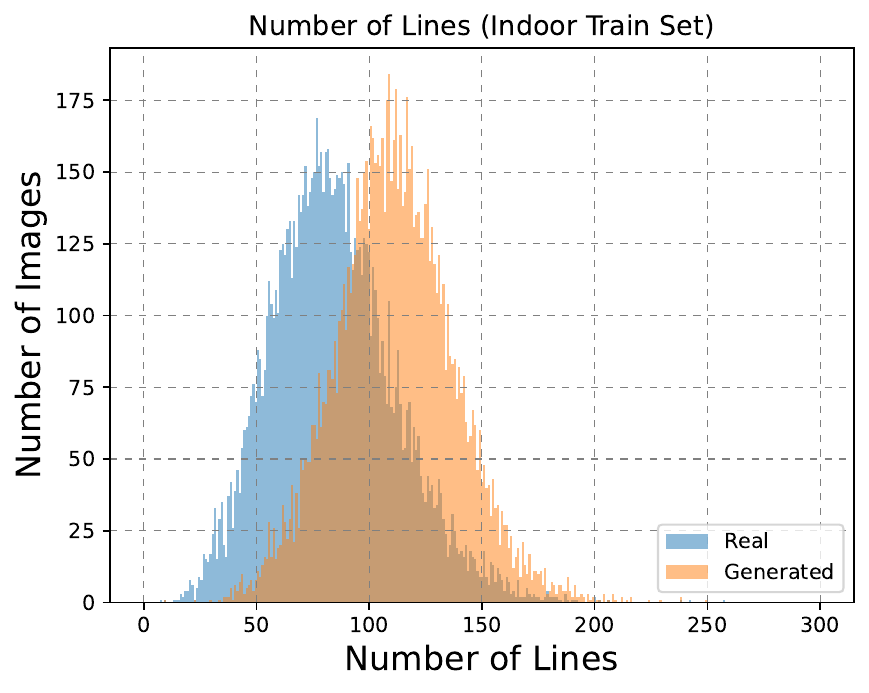}
    \end{subfigure}
    \hfill
     \begin{subfigure}[b]{0.33\textwidth}
          \includegraphics[width=\textwidth]{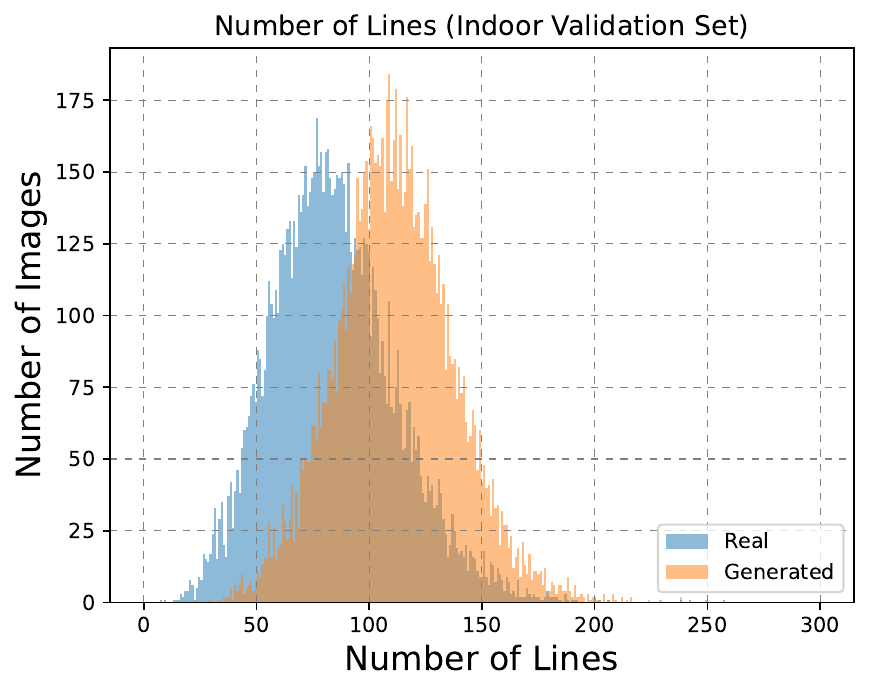}
    \end{subfigure}
    \hfill
    \begin{subfigure}[b]{0.33\textwidth}
        \includegraphics[width=\textwidth]{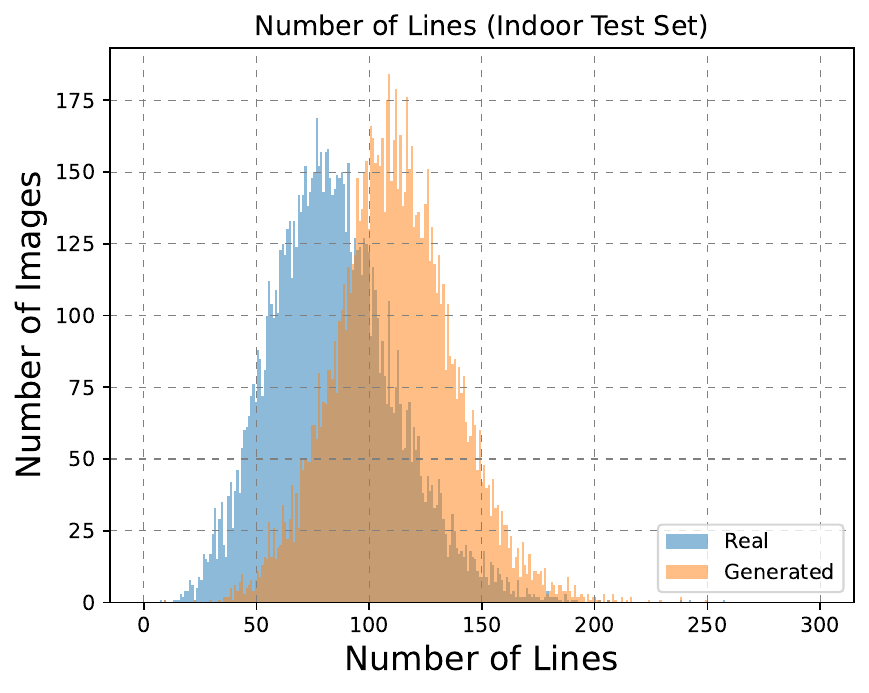}
    \end{subfigure}
    \hfill
    \begin{subfigure}[b]{0.33\textwidth}
        \includegraphics[width=\textwidth]{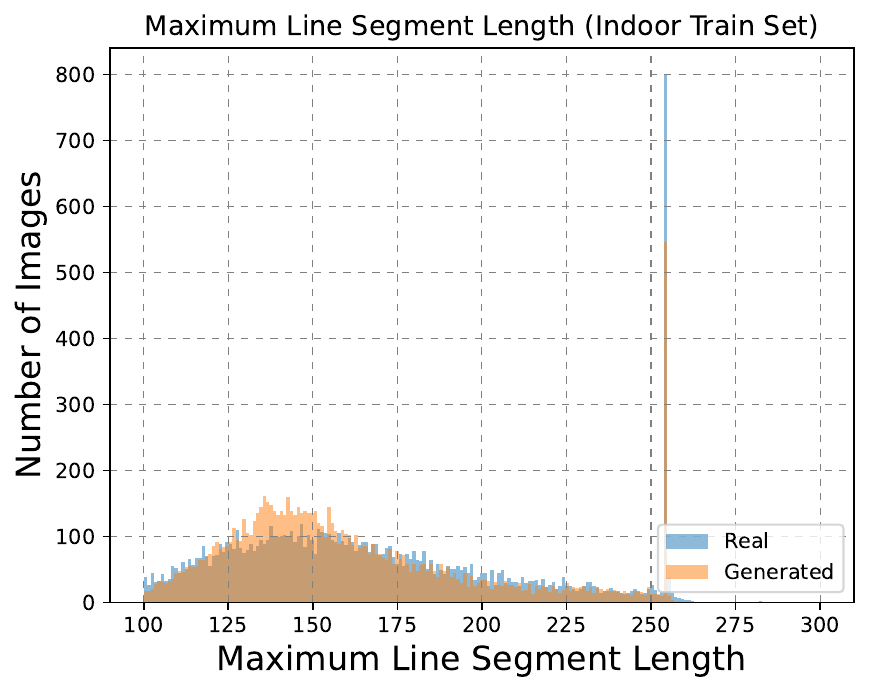}
    \end{subfigure}
    \hfill
     \begin{subfigure}[b]{0.33\textwidth}
          \includegraphics[width=\textwidth]{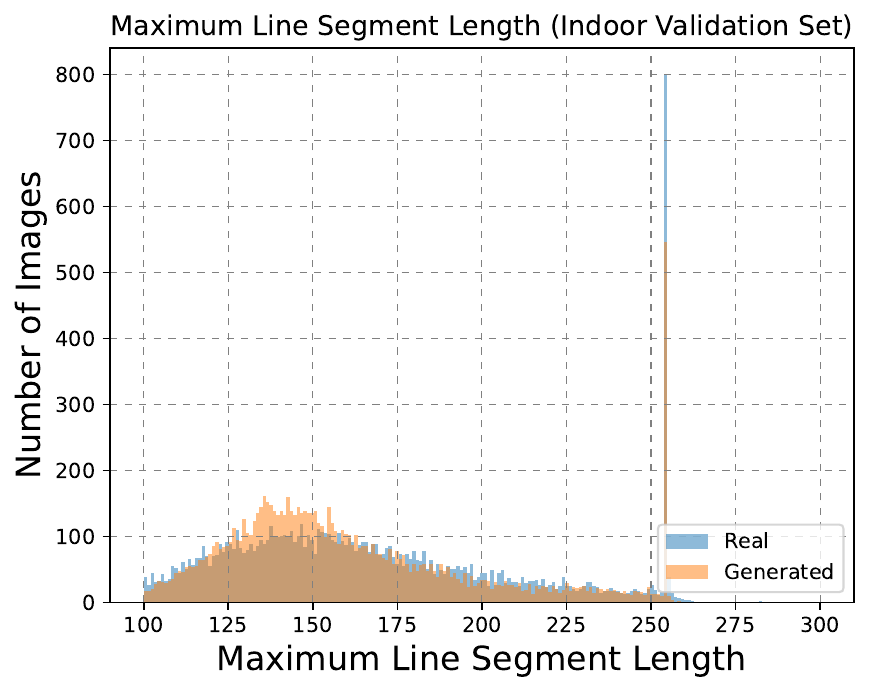}
    \end{subfigure}
    \hfill
    \begin{subfigure}[b]{0.33\textwidth}
        \includegraphics[width=\textwidth]{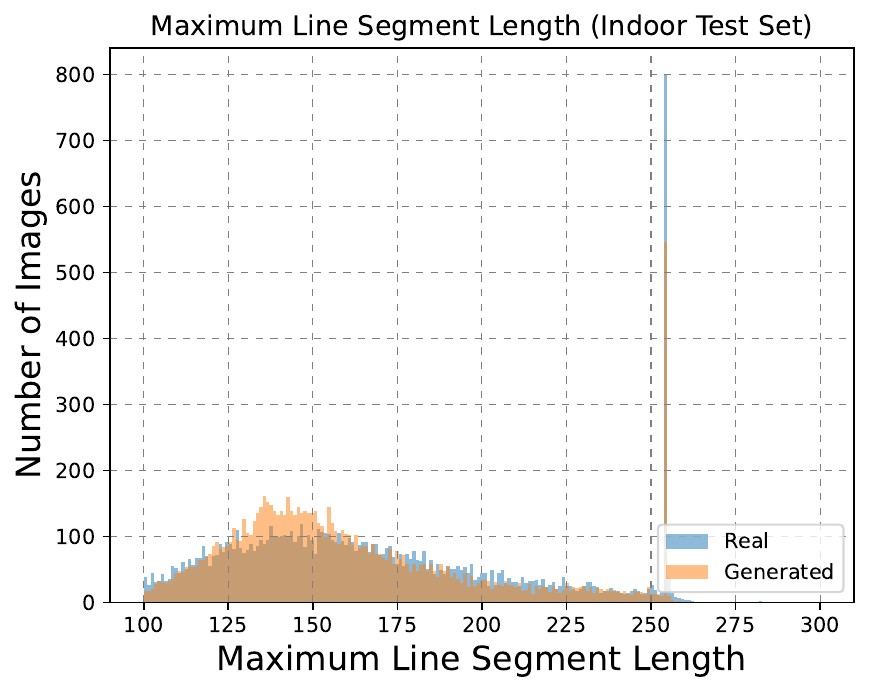}
    \end{subfigure}
    \hfill
    \begin{subfigure}[b]{0.33\textwidth}
        \includegraphics[width=\textwidth]{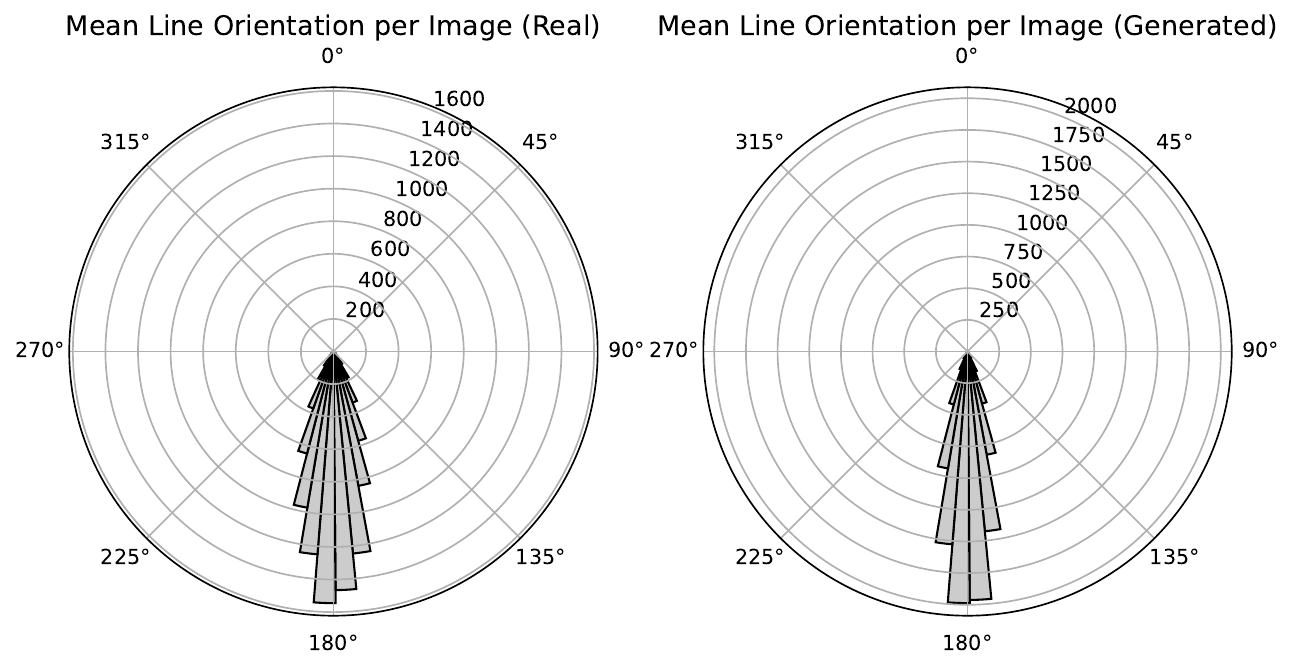}
    \end{subfigure}
    \hfill
     \begin{subfigure}[b]{0.33\textwidth}
          \includegraphics[width=\textwidth]{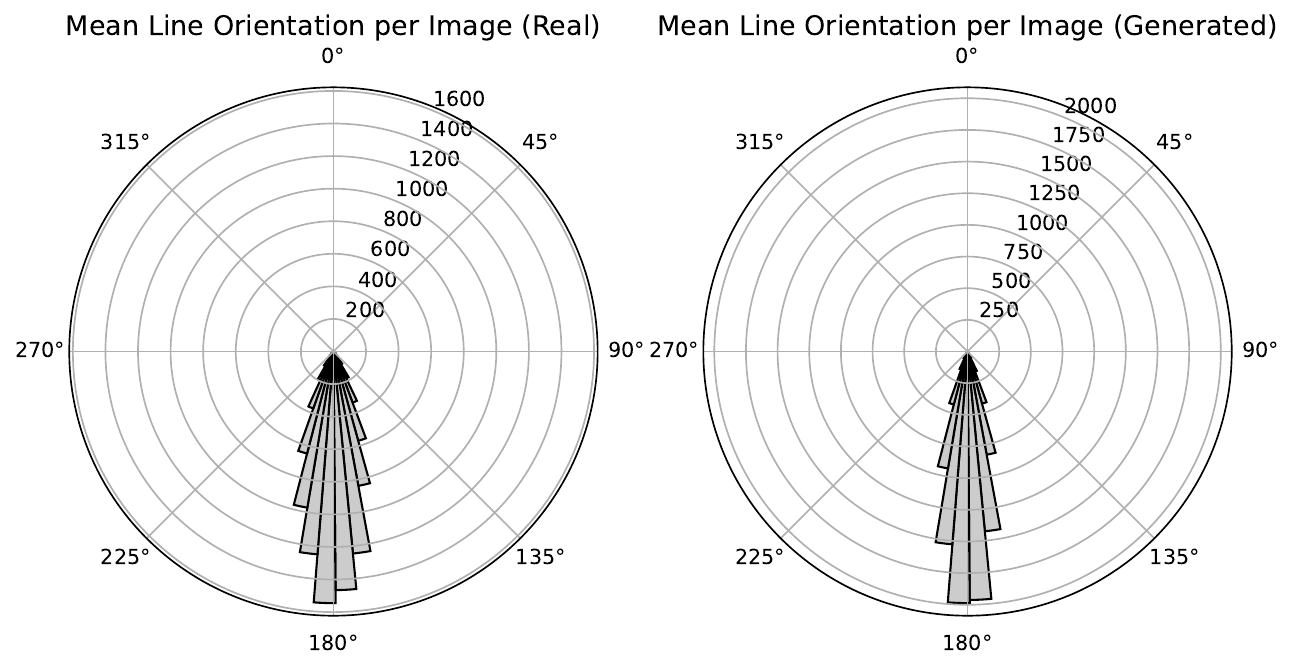}
    \end{subfigure}
    \hfill
    \begin{subfigure}[b]{0.33\textwidth}
        \includegraphics[width=\textwidth]{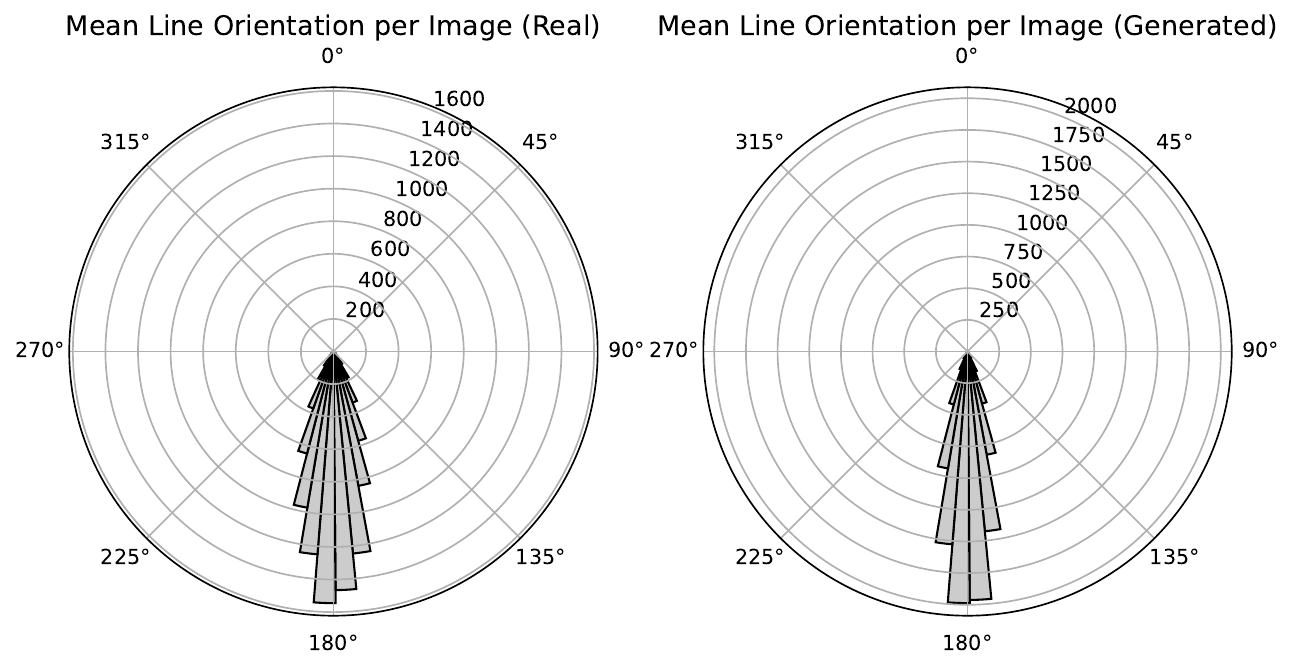}
    \end{subfigure}
    \caption{Line Segment Distribution in Indoor Scenes: We show the distribution of line segment counts and lengths in indoor scenes across training, validation, and test sets. The histograms (top row) compare the number of line segments detected in real versus generated images, with generated images generally exhibiting a different distribution, suggesting a discrepancy in line segment occurrence. The line segment length plots (middle row) show the maximum length of line segments. The polar plots (bottom row) illustrate the mean line orientation per image. While these basic statistical differences provide some discriminative power, they are notably less effective than our PointNet classifiers, which demonstrate a profound ability to detect and focus on critical geometric inconsistencies, as validated by our comprehensive ROC analysis in Figure~\ref{fig:LR_statistical_cues}.}
    \label{fig:line_segment_distribution_indoor}
\end{figure*}

\begin{figure*}[t!]
    \centering
    \begin{subfigure}[b]{0.33\textwidth}
        \includegraphics[width=\textwidth]{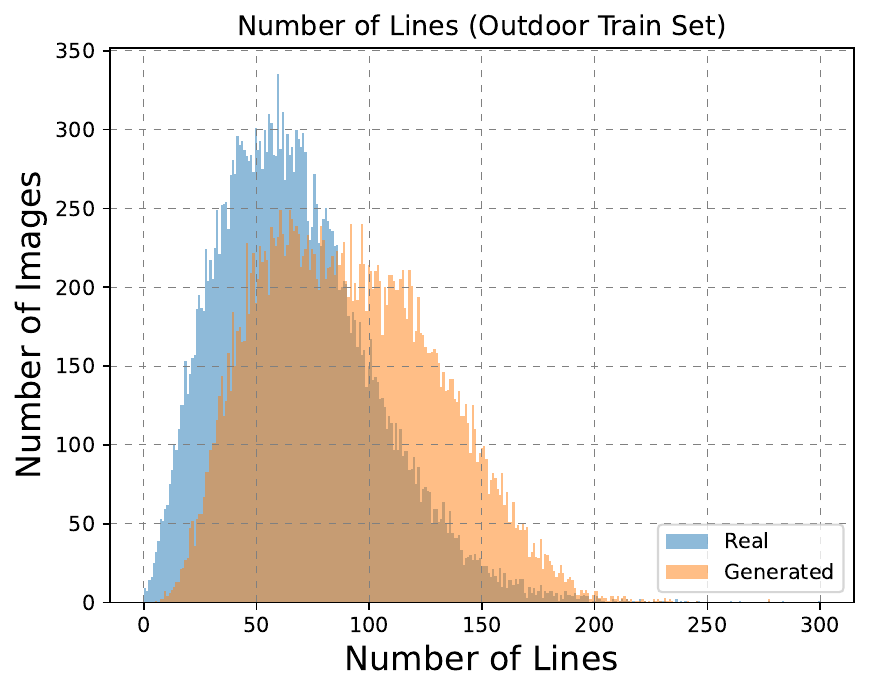}
    \end{subfigure}
    \hfill
     \begin{subfigure}[b]{0.33\textwidth}
          \includegraphics[width=\textwidth]{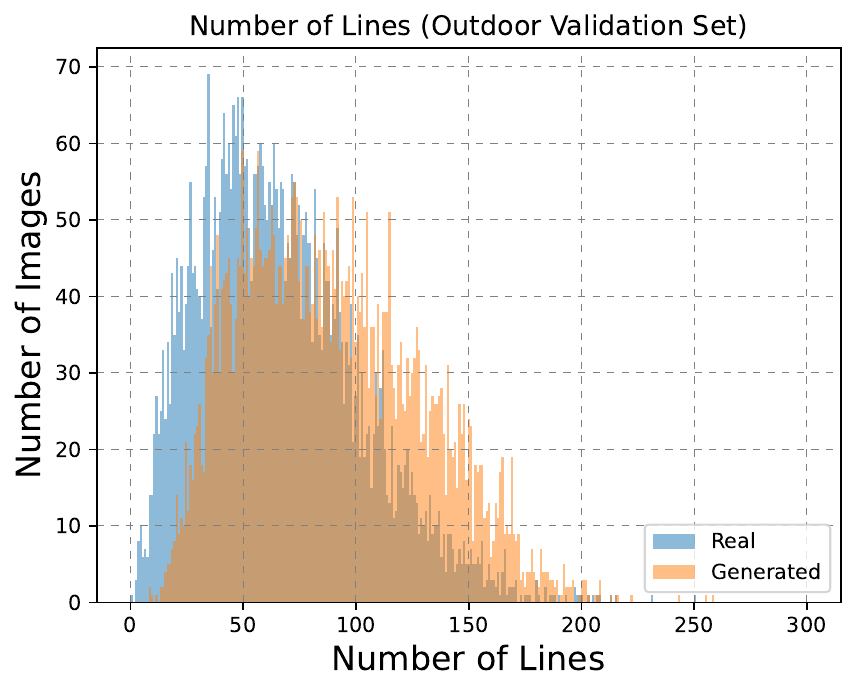}
    \end{subfigure}
    \hfill
    \begin{subfigure}[b]{0.33\textwidth}
        \includegraphics[width=\textwidth]{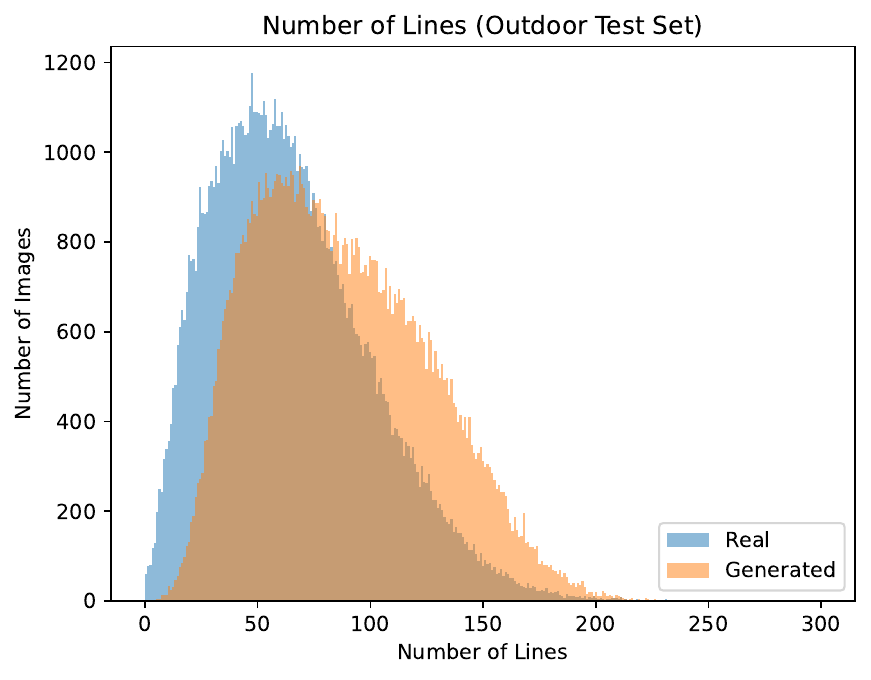}
    \end{subfigure}
    \hfill
    \begin{subfigure}[b]{0.33\textwidth}
        \includegraphics[width=\textwidth]{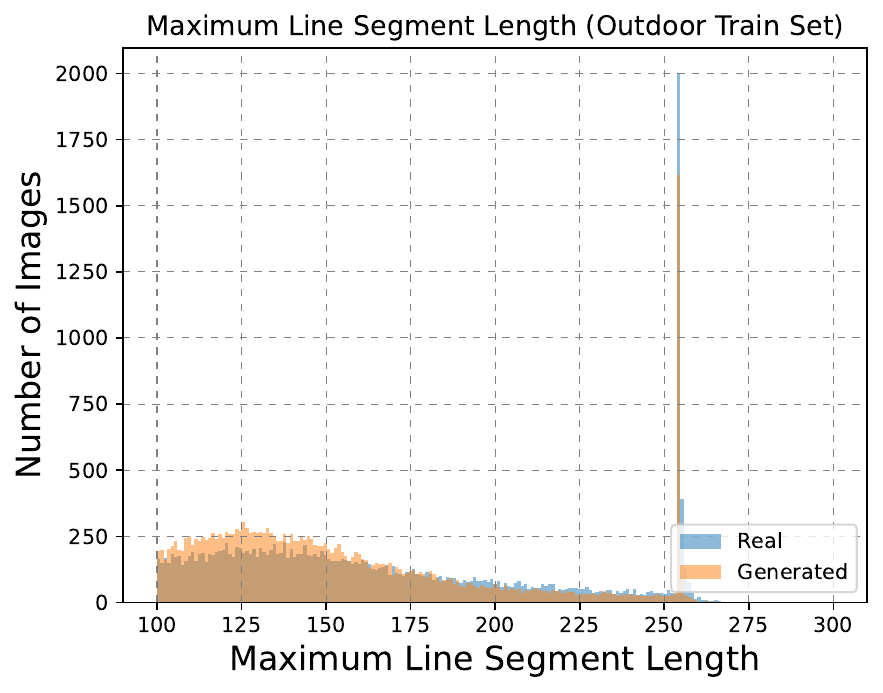}
    \end{subfigure}
    \hfill
     \begin{subfigure}[b]{0.33\textwidth}
          \includegraphics[width=\textwidth]{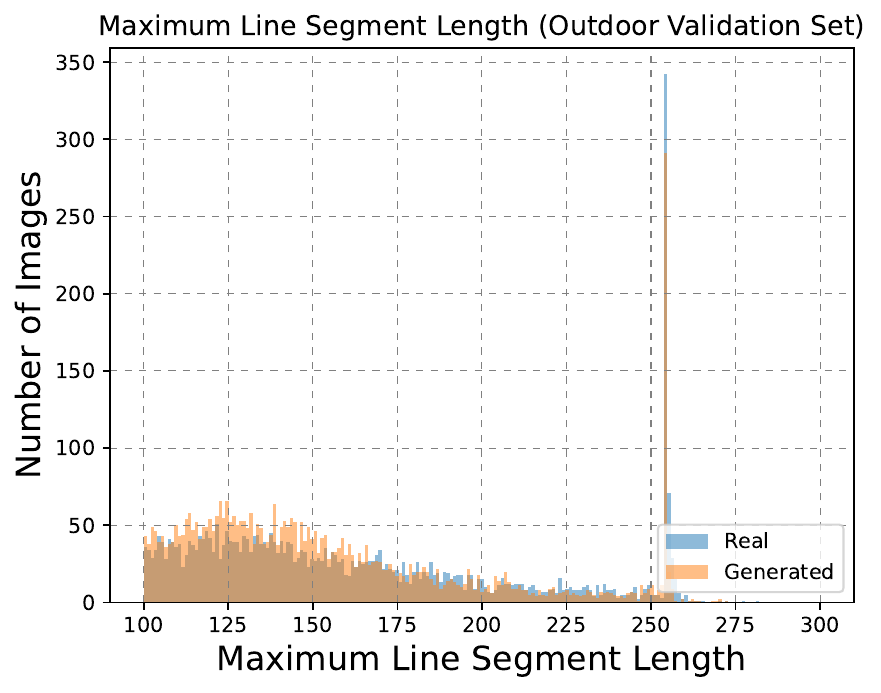}
    \end{subfigure}
    \hfill
    \begin{subfigure}[b]{0.33\textwidth}
        \includegraphics[width=\textwidth]{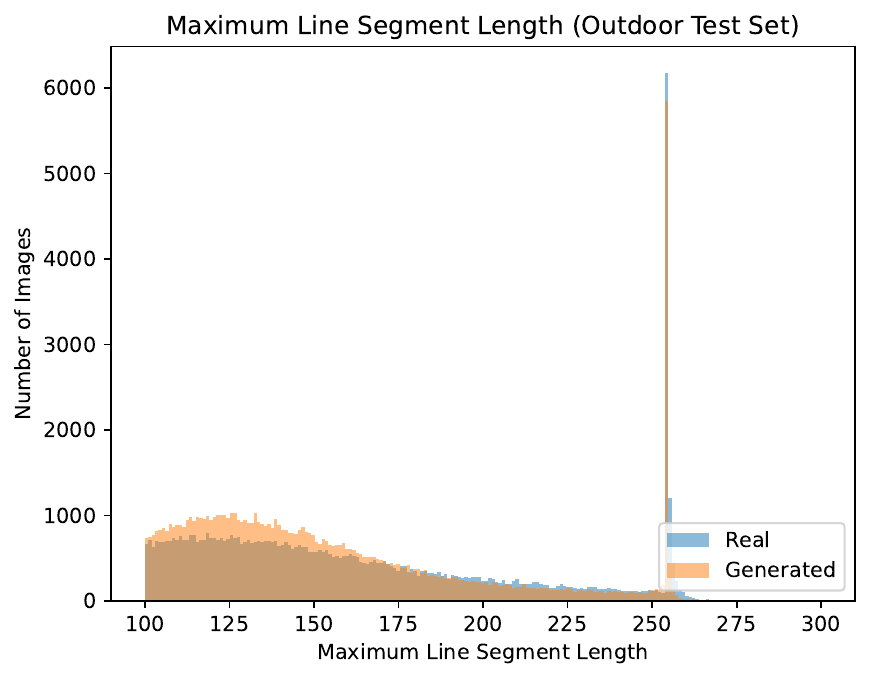}
    \end{subfigure}
    \hfill
    \begin{subfigure}[b]{0.33\textwidth}
        \includegraphics[width=\textwidth]{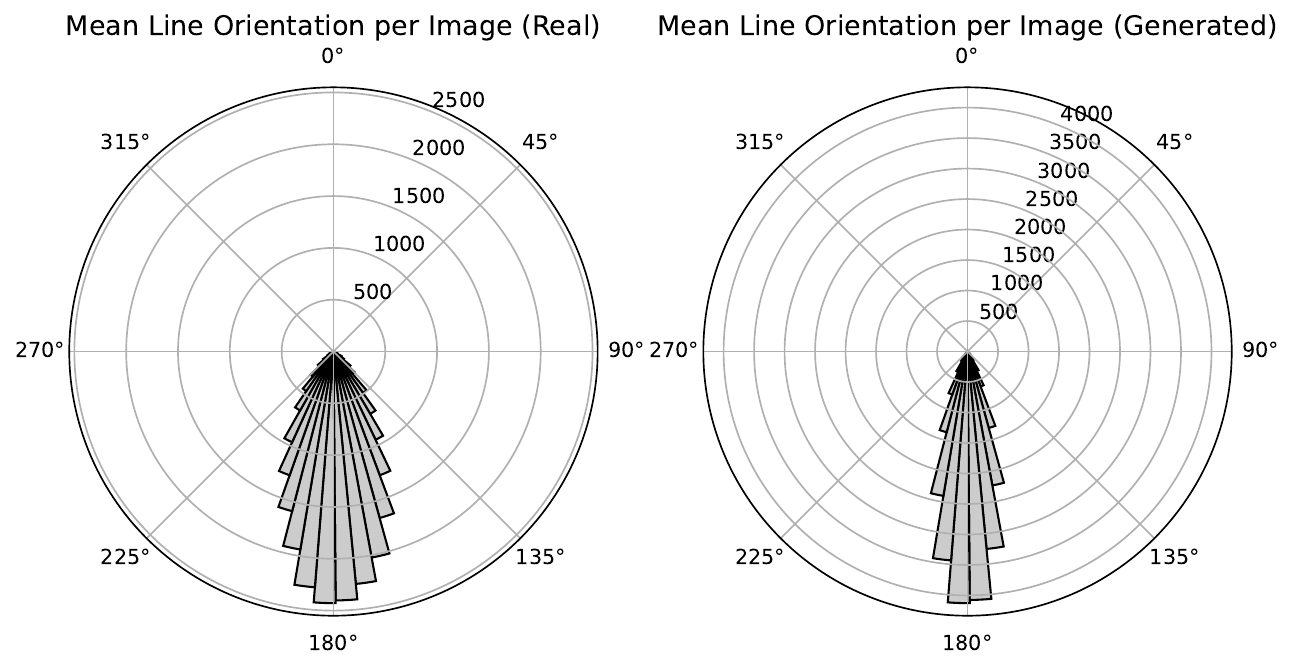}
    \end{subfigure}
    \hfill
     \begin{subfigure}[b]{0.33\textwidth}
          \includegraphics[width=\textwidth]{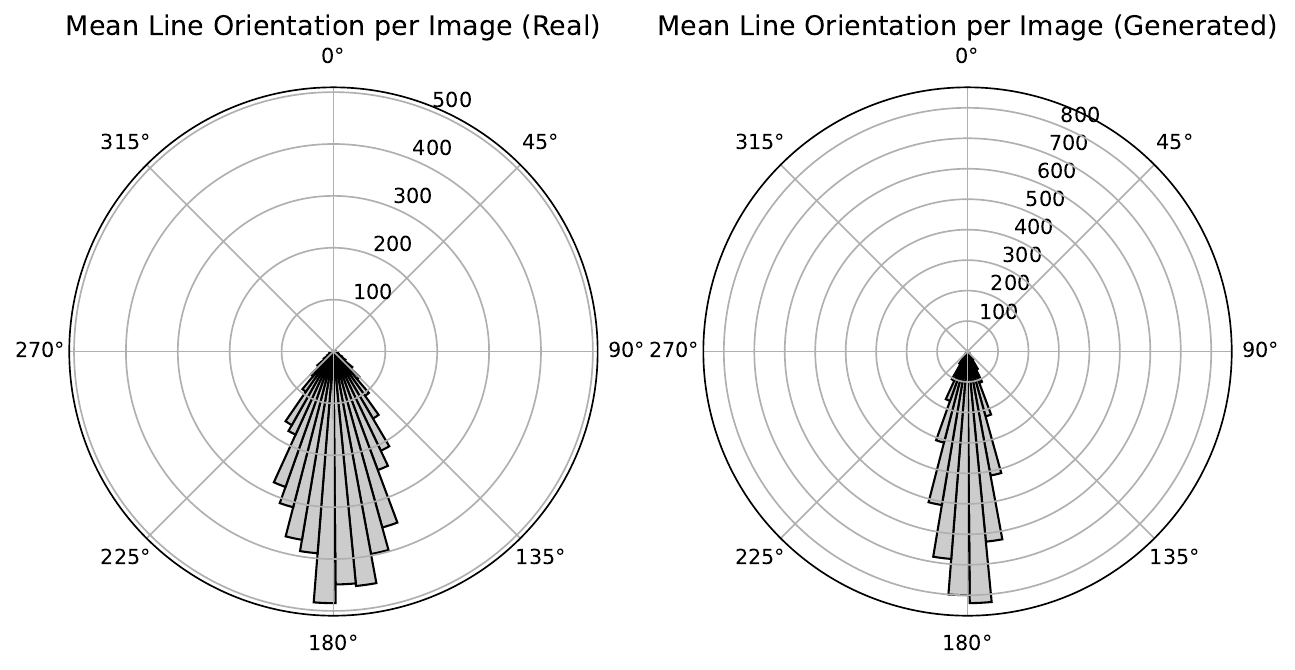}
    \end{subfigure}
    \hfill
    \begin{subfigure}[b]{0.33\textwidth}
        \includegraphics[width=\textwidth]{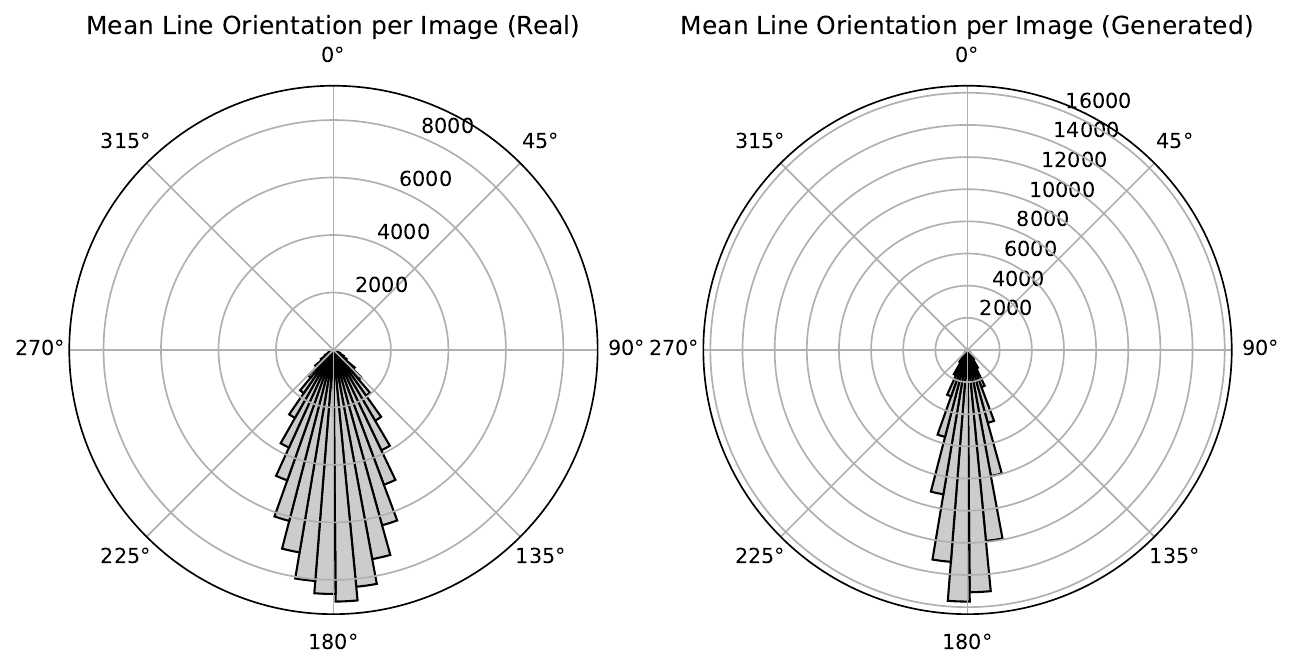}
    \end{subfigure}
    \caption{Line Segment Distribution in Outdoor Scenes: We show the distribution of line segment counts and lengths in indoor scenes across training, validation, and test sets. The histograms (top row) compare the number of line segments detected in real versus generated images, with generated images generally exhibiting a different distribution, suggesting a discrepancy in line segment occurrence. The line segment length plots (middle row) show the maximum length of line segments. The polar plots (bottom row) illustrate the mean line orientation per image.
     While these basic statistical differences provide some discriminative power, they are notably less effective than our PointNet classifiers, which demonstrate a profound ability to detect and focus on critical geometric inconsistencies, as validated by our comprehensive ROC analysis in Figure~\ref{fig:LR_statistical_cues}.}
    \label{fig:line_segment_distribution_outdoor}
\end{figure*}

An ROC plot in Figure~\ref{fig:LR_statistical_cues} shows that while using statistical biases helps detect generated images over chance, ResNet classifiers trained directly on these cues still outperform them.

\begin{table}[t!]
\centering
\caption{We quantify the distribution of detection agreement among three types of cues: Line Segment (LS), Perspective Fields (PF), and Object-Shadow (OS), for the images processed by Stable Diffusion-XL. The output indicates whether each method can accurately identify generated images as either real or generated. The ``Yes" indicates that the method has correctly detected generated images, whereas ``No" indicates that the method has identified generated images as real. We have also provided the absolute and percentage values of images for both indoor and outdoor domains' unconfident test set in the last two columns. The table reveals a statistically significant correlation between Line Segment and Perspective field cues (p-value $\approx 2e^{-16}$), suggesting they are not independent in their detection of generated images. Conversely, Object-Shadow Cues demonstrate a different pattern of detection, with the probability of identifying an image as generated being lower than that of Line Segment Cues. This shows that they are complementary and look at distinct discrepancies in the images. A qualitative figure demonstrating a complementary capability is in Figure~\ref{fig:grad_cam_complement} 
of the main text.} 
\label{tab:disagreement}
\resizebox{\linewidth}{!}{%
\begin{tabular}{ccccc}
\toprule
LS cues & PF cues & OS cues & Indoor & Outdoor\\
\midrule
Yes & Yes & Yes & 10520 (53.71\%)  & 2382 (45.38\%)\\
Yes & Yes & No & 4844 (24.73\%) &  1314 (25.03\%)\\
Yes & No & Yes & 1033 (5.27\%) &  287 (5.47\%)\\
Yes & No & No & 725 (3.70\%) &  260 (4.95\%)\\
No & Yes & Yes & 872 (4.45\%) &  322 (6.13\%)\\
No & Yes & No & 874 (4.46\%) &  423 (8.06\%)\\
No & No & Yes & 285 (1.45\%) &  102 (1.94\%)\\
No & No & No & 435 (2.22\%) &  159 (3.03\%)\\
\bottomrule
\end{tabular}
}
\end{table}

\begin{figure*}[t!]
  \centering
  \footnotesize
  \setlength\tabcolsep{0.2pt}
  \renewcommand{\arraystretch}{0.1}
  \begin{tabular}{ccccc}

    \includegraphics[width=0.195\linewidth]{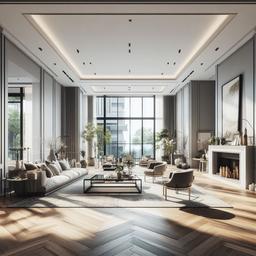} &
    \includegraphics[width=0.195\linewidth]{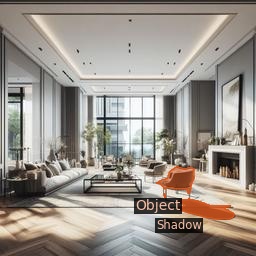} &
    \includegraphics[width=0.195\linewidth]{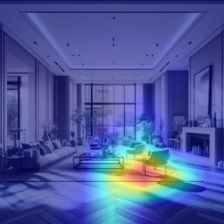} &
    \includegraphics[width=0.195\linewidth]{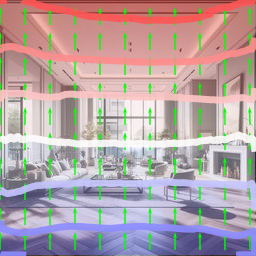} &
    \includegraphics[width=0.195\linewidth]{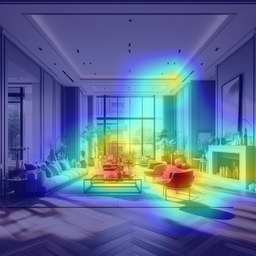} 
\\
    \includegraphics[width=0.195\linewidth]{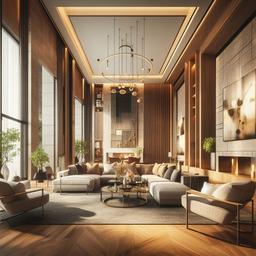} &
    \includegraphics[width=0.195\linewidth]{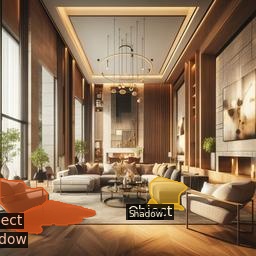} &
    \includegraphics[width=0.195\linewidth]{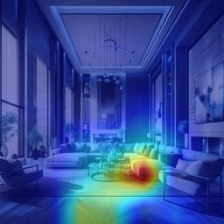} &
    \includegraphics[width=0.195\linewidth]{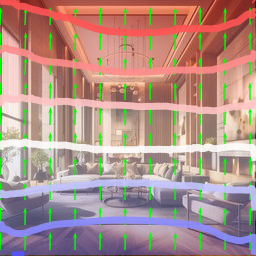} &
    \includegraphics[width=0.195\linewidth]{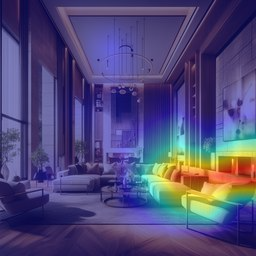} 
\\
    \includegraphics[width=0.195\linewidth]{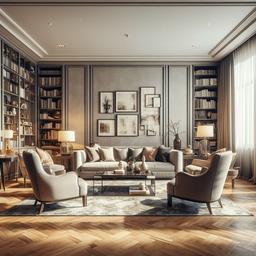} &
    \includegraphics[width=0.195\linewidth]{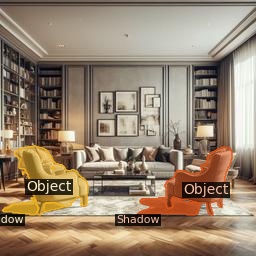} &
    \includegraphics[width=0.195\linewidth]{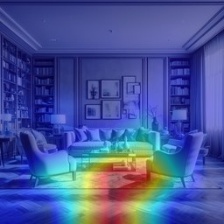} &
    \includegraphics[width=0.195\linewidth]{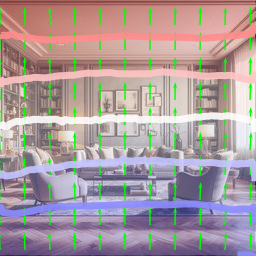} &
    \includegraphics[width=0.195\linewidth]{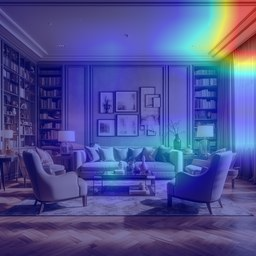} 
\\
    \includegraphics[width=0.195\linewidth]{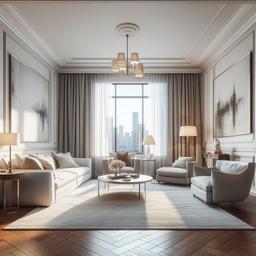} &
    \includegraphics[width=0.195\linewidth]{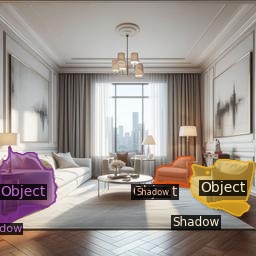} &
    \includegraphics[width=0.195\linewidth]{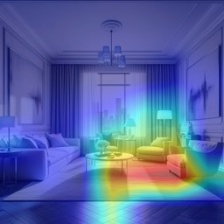} &
    \includegraphics[width=0.195\linewidth]{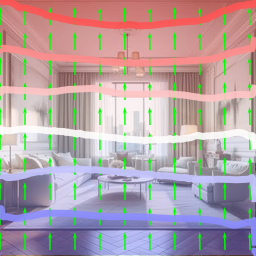} &
    \includegraphics[width=0.195\linewidth]{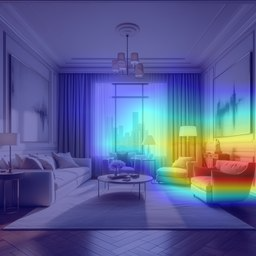}     \vspace{5pt}
\\
  Generated Image & Object-Shadow  (OS) &  OS GradCam & Perspective Fields & Perspective Fields GradCam \\
        \end{tabular}
        \vspace{-7pt}
  \caption{All interior scenes generated using Dalle-3. We analyze them using Object-Shadow (OS) cues and Perspective Fields (PF), along with their respective GradCam visualizations. The OS GradCam highlights areas where shadow directions or lengths don't appear to match the scene's lighting. For example, in the first and third rows, the shadows beneath the furniture don't seem to fit the objects casting them. The second row's OS GradCam shows an unnatural shadow on the sofa that's difficult to spot. Meanwhile, the PF analysis exposes inaccuracies in line alignment and vanishing points. In the top and third rows, the PF GradCam highlights inconsistencies along the room's ceiling lines and window frames that don't match the rest of the scene's perspective geometry. In the second and fourth rows, it detects inconsistencies on the side wall beneath the painting region.
  }
        \label{fig:dalle_gradcam}
        \vspace{-5pt}
	\end{figure*}

 \begin{figure*}[t!]
  \centering
  \footnotesize
  \setlength\tabcolsep{0.2pt}
  \renewcommand{\arraystretch}{0.1}
  \begin{tabular}{ccccccc}
    \includegraphics[width=0.141\linewidth]{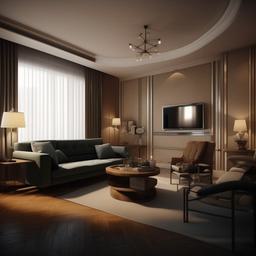} &
    \includegraphics[width=0.141\linewidth]{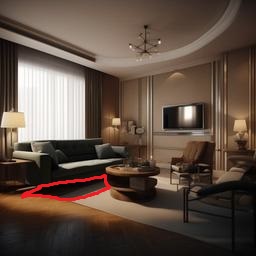} &
    \includegraphics[width=0.141\linewidth]{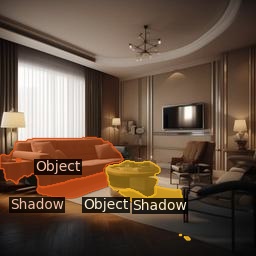} &
    \includegraphics[width=0.141\linewidth]{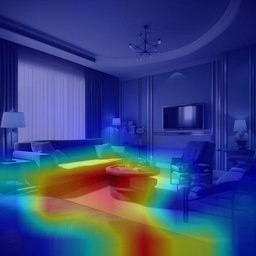} &
    \includegraphics[width=0.141\linewidth]{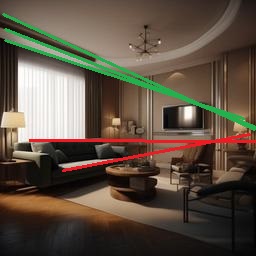} &
    \includegraphics[width=0.141\linewidth]{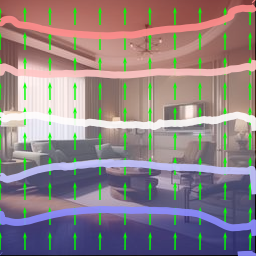} &
    \includegraphics[width=0.141\linewidth]{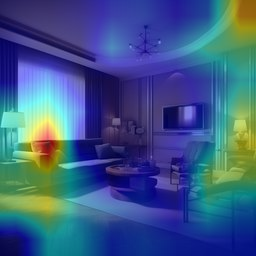} \vspace{1pt}
\\
    \includegraphics[width=0.141\linewidth]{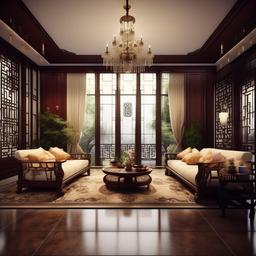} &
    \includegraphics[width=0.141\linewidth]{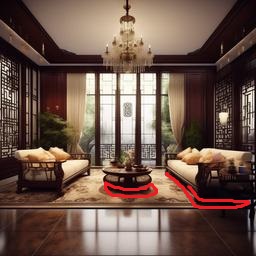} &
    \includegraphics[width=0.141\linewidth]{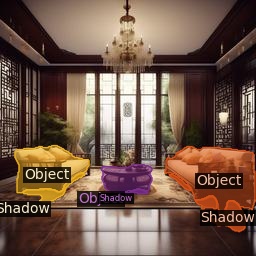} &
    \includegraphics[width=0.141\linewidth]{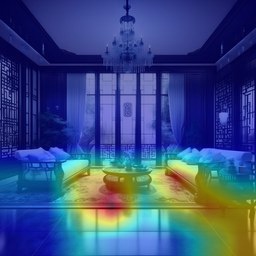} &
    \includegraphics[width=0.141\linewidth]{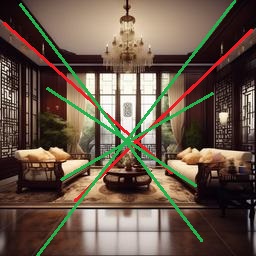} &
    \includegraphics[width=0.141\linewidth]{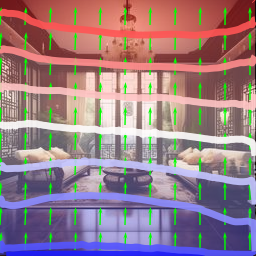} &
    \includegraphics[width=0.141\linewidth]{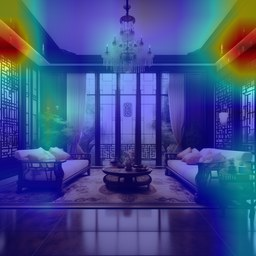} 
\\
    \includegraphics[width=0.141\linewidth]{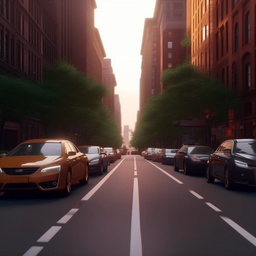} &
    \includegraphics[width=0.141\linewidth]{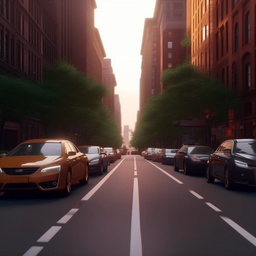} &
    \includegraphics[width=0.141\linewidth]{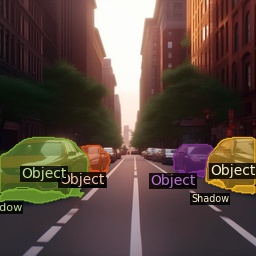} &
    \includegraphics[width=0.141\linewidth]{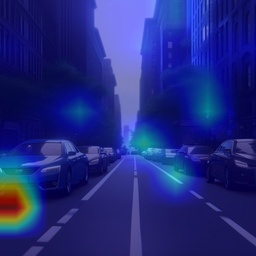} &
    \includegraphics[width=0.141\linewidth]{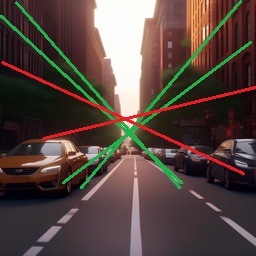} &
    \includegraphics[width=0.141\linewidth]{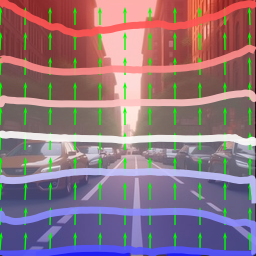} &
    \includegraphics[width=0.141\linewidth]{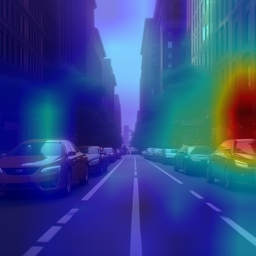} \vspace{1pt}
\\
    \includegraphics[width=0.141\linewidth]{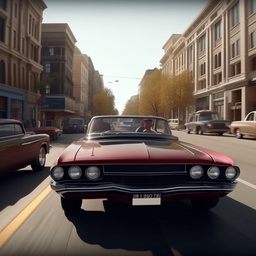} &
    \includegraphics[width=0.141\linewidth]{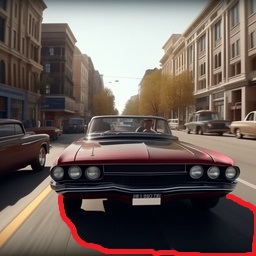} &
    \includegraphics[width=0.141\linewidth]{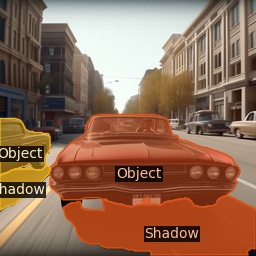} &
    \includegraphics[width=0.141\linewidth]{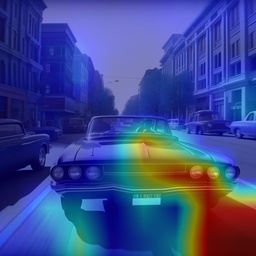} &
    \includegraphics[width=0.141\linewidth]{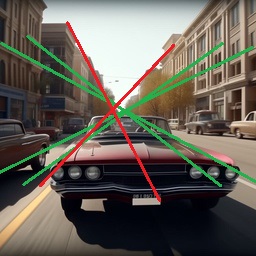} &
    \includegraphics[width=0.141\linewidth]{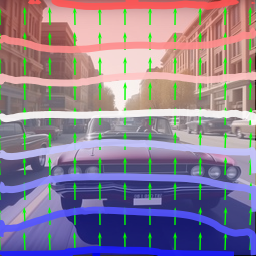} &
    \includegraphics[width=0.141\linewidth]{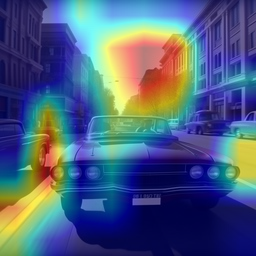}     \vspace{2pt}
\\
  Generated Image & Shadow Errors & Object-Shadow (OS) &  OS GradCam & VP Errors & {\scriptsize Perspective Fields (PF)} & PF GradCam\\
        \end{tabular}
        \vspace{-5pt}
  \caption{Grad-CAM results for indoor and outdoor scenes generated by Kandinsky. The second column shows shadow errors, while the third column overlays detected object-shadow pairs~\cite{wang2022instance}. Grad-CAM applied to our Object-Shadow classifier (fourth column) reveals mismatched shadow lengths (first row),  shorter-than-expected shadows (third row) and incorrect shadow shapes (fourth row). The sixth column shows Perspective Fields~\cite{jin2023perspective}, and Grad-CAM applied to our Perspective Fields classifier (last column) confirms large perspective distortions on building facades, also supported by the vanishing point errors in the fifth column.
  }
        \label{fig:gradcam_kandinsky}
        \vspace{-3pt}
	\end{figure*} 

\begin{figure*}[t!]
  \centering
  \footnotesize
  \setlength\tabcolsep{0.2pt}
  \renewcommand{\arraystretch}{0.1}
  \begin{tabular}{ccccc}

    \includegraphics[width=0.195\linewidth]{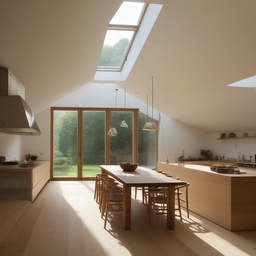} &
    \includegraphics[width=0.195\linewidth]{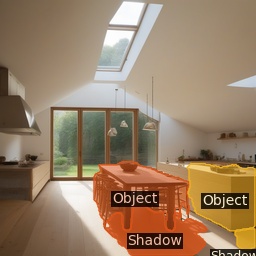} &
    \includegraphics[width=0.195\linewidth]{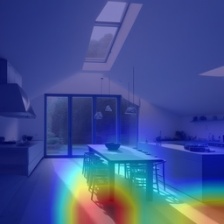} &
    \includegraphics[width=0.195\linewidth]{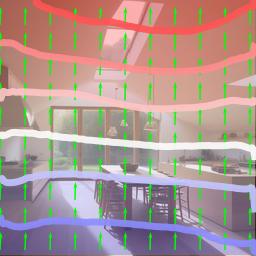} &
    \includegraphics[width=0.195\linewidth]{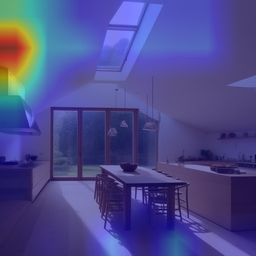}
\\
    \includegraphics[width=0.195\linewidth]{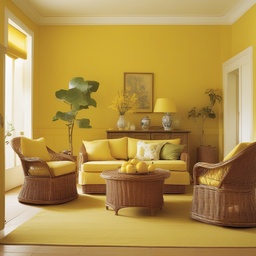} &
    \includegraphics[width=0.195\linewidth]{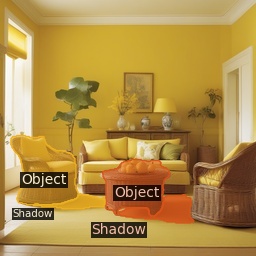} &
    \includegraphics[width=0.195\linewidth]{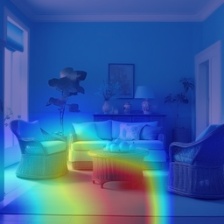} &
    \includegraphics[width=0.195\linewidth]{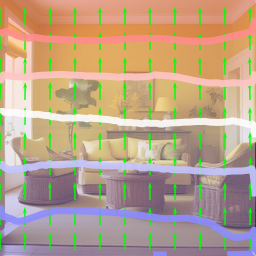} &
    \includegraphics[width=0.195\linewidth]{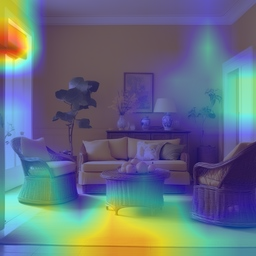}
\\
    \includegraphics[width=0.195\linewidth]{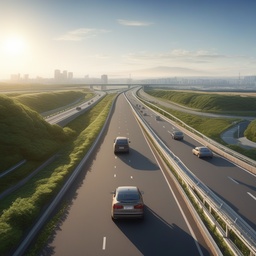} &
    \includegraphics[width=0.195\linewidth]{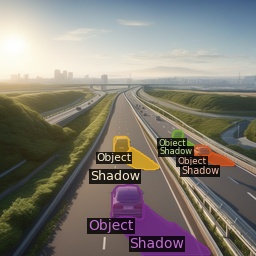} &
    \includegraphics[width=0.195\linewidth]{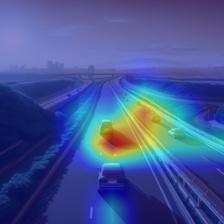} &
    \includegraphics[width=0.195\linewidth]{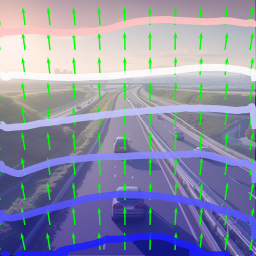} &
    \includegraphics[width=0.195\linewidth]{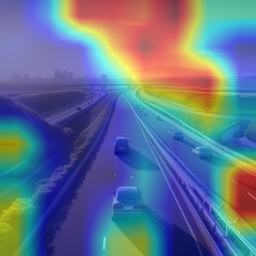} 
\\
    \includegraphics[width=0.195\linewidth]{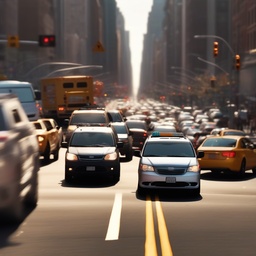} &
    \includegraphics[width=0.195\linewidth]{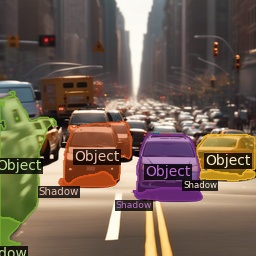} &
    \includegraphics[width=0.195\linewidth]{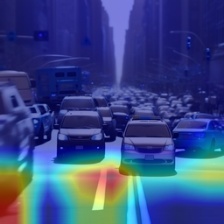} &
    \includegraphics[width=0.195\linewidth]{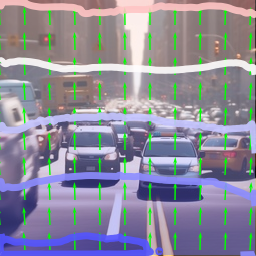} &
    \includegraphics[width=0.195\linewidth]{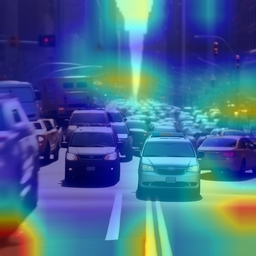} 
\\
    \includegraphics[width=0.195\linewidth]{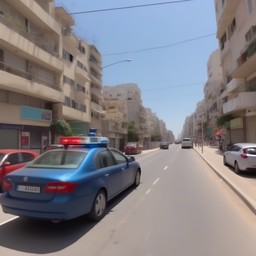} &
    \includegraphics[width=0.195\linewidth]{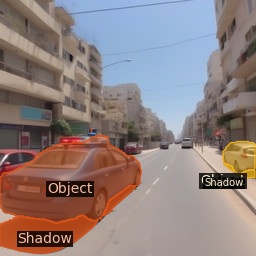} &
    \includegraphics[width=0.195\linewidth]{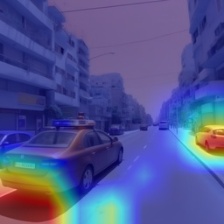} &
    \includegraphics[width=0.195\linewidth]{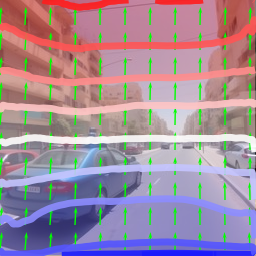} &
    \includegraphics[width=0.195\linewidth]{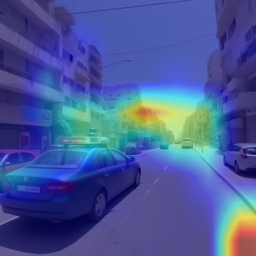}
\vspace{5pt}
\\   
  Generated Image & Object-Shadow (OS) &  OS GradCam & Perspective Fields & Perspective Fields GradCam \\
        \end{tabular}
        \vspace{-5pt}
  \caption{The first column displays images generated by Stable Diffusion-XL. The second column overlays detected object-shadow pairs from ~\cite{wang2022instance}, highlighting the model's ability to identify these features. The third column applies Grad-CAM to our Object-Shadow classifier.  This shows areas most diagnostic of
    synthetic generation.  Note: in the first row, the Grad-CAM weights suggest a shadow problem at the left side chair, which is difficult to check but plausible; in the
    second row, the shadow cast by the coffee table is in the wrong direction and Grad-CAM identifies this error as diagnostic.
    The fourth column shows the Perspective Fields of~\cite{jin2023perspective}, and the fifth column shows Grad-CAM when applied to our
    Perspective Fields classifier.  Note: in the first row,
    Grad-CAM weights identify a problem with the top of the cupboard on the left, which is difficult to confirm but plausible; in the second row, Grad-CAM
    weights identify a visible problem with the blind on the left. in
    the third row, the cars cast shadows in different directions and Grad-CAM identifies this
    error as diagnostic; in the fourth row, two cars in front cast shadows in different
    directions and Grad-CAM identifies this
    error as diagnostic; in the  fifth row, Grad-CAM identifies the (very odd) structure
    of the buildings near the vanishing point as a problem, based on perspective field distortion. Best viewed on screen. 
  }
        \label{fig:gradcam_SDXL}
	\end{figure*}

 \begin{figure*}[t!]
  \centering
  \footnotesize
  \setlength\tabcolsep{0.2pt}
  \renewcommand{\arraystretch}{0.1}
  \begin{tabular}{ccccc}

    \includegraphics[width=0.195\linewidth]{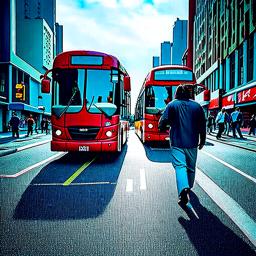} &
    \includegraphics[width=0.195\linewidth]{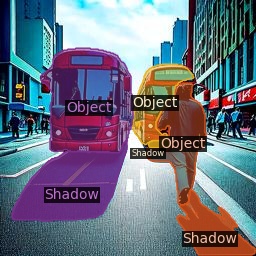} &
    \includegraphics[width=0.195\linewidth]{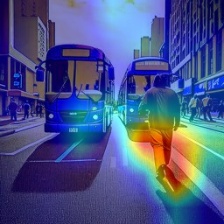} &
    \includegraphics[width=0.195\linewidth]{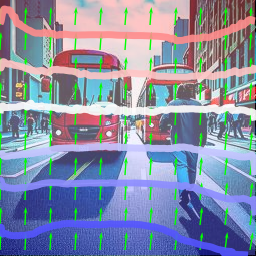} &
    \includegraphics[width=0.195\linewidth]{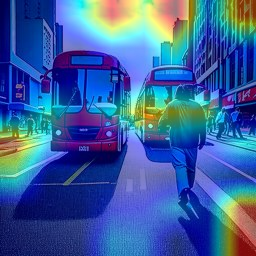} 
\\
    \includegraphics[width=0.195\linewidth]{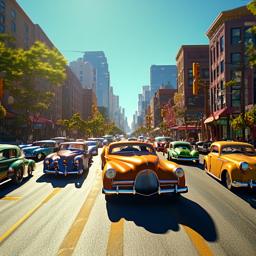} &
    \includegraphics[width=0.195\linewidth]{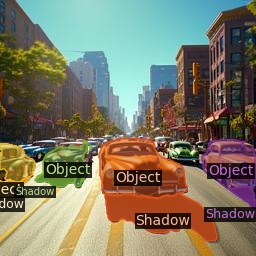} &
    \includegraphics[width=0.195\linewidth]{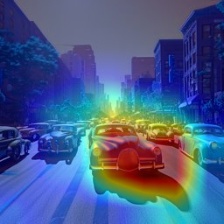} &
    \includegraphics[width=0.195\linewidth]{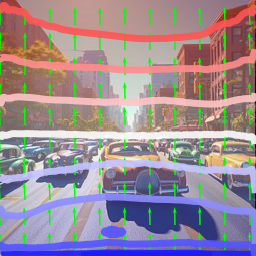} &
    \includegraphics[width=0.195\linewidth]{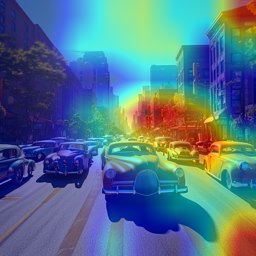} 
\\
    \includegraphics[width=0.195\linewidth]{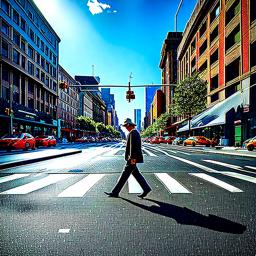} &
    \includegraphics[width=0.195\linewidth]{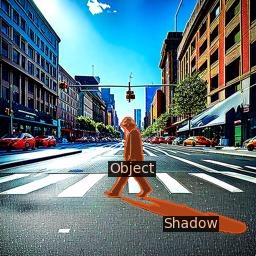} &
    \includegraphics[width=0.195\linewidth]{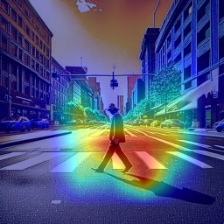} &
    \includegraphics[width=0.195\linewidth]{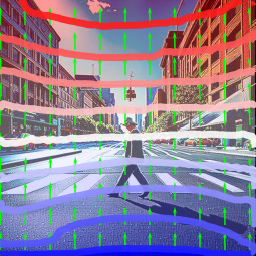} &
    \includegraphics[width=0.195\linewidth]{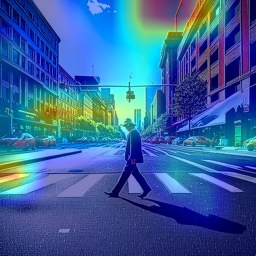}  
    \vspace{5pt}
\\
  Generated Image & Object-Shadow  (OS) &  OS GradCam & Perspective Fields & Perspective Fields GradCam \\
        \end{tabular}
        \vspace{-7pt}
  \caption{The generated street scenes in Adobe's Firefly have inconsistencies in projective geometry. We show Object-Shadow (OS) and Perspective Fields (PF) analyses and have presented each generated image alongside the results. In the first row, the shadow of the bus on the left is in one direction, while the shadow of the bus on the right and the pedestrian point is in opposite directions. The second row shows the OS GradCam pinpointing a car's shadow that is unrealistically elongated on one side. In the third, we observe pedestrians with shadows that are inconsistent with the lighting. The Perspective Fields analysis in rows two and four detects line inconsistencies deep in the scene and near vanishing points, while in the first and last rows, it captures discrepancies on the road markings and building facades.
  }
        \label{fig:firefly_gradcam}
        \vspace{-5pt}
	\end{figure*}

